\documentclass[runningheads]{llncs}

\usepackage{eccv}

\usepackage{eccvabbrv}

\usepackage{graphicx}
\usepackage{booktabs}
\usepackage{comment}
\usepackage{xcolor}
\usepackage{wrapfig}
\usepackage{algorithm}
\usepackage{algpseudocode}
\usepackage{marvosym}
\usepackage{bbold}

\usepackage[accsupp]{axessibility}  %

\usepackage{hyperref}

\usepackage{orcidlink}

\newif\ifshowteaser
\showteasertrue %

\definecolor{lightgray}{gray}{0.8}
\definecolor{lightgreen}{rgb}{0.74,0.96,0.65}
\definecolor{commentgreen}{rgb}{0,0.4,0.26}
\definecolor{commentblue}{rgb}{0.1,0.1,0.6}
\definecolor{lightred}{rgb}{0.96,0.74,0.65}
\definecolor{cvprblue}{rgb}{0.21,0.49,0.74}

\renewcommand{\eqref}[1]{Eq.~\ref{#1}}

\newcommand{\vecind}[2]{{#1}({#2})}

\newcommand{\upind}[2]{{#1}^{#2}}

\newcommand{\albedo}{\alpha}

\newcommand{\freq}{f}

\newcommand{\probe}{p}

\newcommand{\light}{l} %

\newcommand{\timesym}{t}

\newcommand{\flux}{\phi}

\newcommand{\tof}{\tau}
\newcommand{\spacevol}{v}

\newcommand{\exposure}{\timesym_{\text{exp}}}

\newcommand{\lightloc}[1]{\expandafter\ifx\expandafter\relax
\detokenize{#1}\relax\mathbf{\light}\else {\mathbf{\light}}_{#1}\fi}

\newcommand{\tofsym}[1]{\expandafter\ifx\expandafter\relax
  \detokenize{#1}\relax\tof\else \upind{\tof}{#1}\fi}

\newcommand{\albedosym}[1]{\expandafter\ifx\expandafter\relax
  \detokenize{#1}\relax\albedo\else \vecind{\albedo}{#1}\fi}

\newcommand{\fluxfunc}[1]{\flux{(#1)}}

\newcommand{\probemeas}{\mathcal{E}}
\newcommand{\probemeasnorm}{\widetilde{\probemeas}}

\newcommand{\spacetime}{\mathbf{x}}
\newcommand{\spacex}{x}
\newcommand{\spacey}{y}

\newcommand{\setspacetime}{\mathrm{\Omega}}

\newcommand{\freqx}{\freq_{\spacex}}
\newcommand{\freqy}{\freq_{\spacey}}
\newcommand{\freqt}{\freq_{\timesym}}
\newcommand{\probespacetime}{\probe(\spacex, \spacey, \timesym)}

\newcommand{\width}{w}
\newcommand{\height}{h}

\newcommand{\spacetimeflux}{\fluxfunc{\spacex, \spacey, \timesym}}

\newcommand{\freqvec}{\mathbf{\freq}}

\newcommand{\freqsall}{\mathcal{F}}
\newcommand{\freqscfar}{\freqsall_{\text{det}}}

\newcommand{\events}{\mathcal{P}}
\newcommand{\eventpol}{\sigma}

\newcommand{\cfarsigval}{\alpha}
\newcommand{\cfarsigvalvel}{\cfarsigval_{\text{vel}}}
\newcommand{\critvalue}{c_\cfarsigval}

\begin{document}

\title{Spatiotemporal Flux Probing for Single-Photon Videography
} 

\author{
Jerry Yan\inst{1}$^{*}$ \and
Matteo Forlivesi\inst{1}$^{*}$ \and
Bowen Tan\inst{1} \and
Andrew Xie\inst{2} \and
Siddharth Somasundaram\inst{3} \and
Sotiris Nousias\inst{1}
}

\authorrunning{J.~Yan et al.}

\institute{Dept. of Computer Science, Purdue University \and
Dept. of Computer Science, University of Toronto \and
MIT Media Lab, Massachusetts Institute of Technology \\
\url{https://jerukan.github.io/spt-flux-probing}
}

\maketitle
\let\svthefootnote\thefootnote
\let\thefootnote\relax\footnotetext{$^{*}$ Equal contribution: \email{\{yan569,mforlive\}@purdue.edu}}
\let\thefootnote\svthefootnote
\begin{abstract}

We address the problem of recovering high-speed videos from dynamic scenes under extreme photon sparsity. Existing methods rely on aggregating photon detections in local spatiotemporal windows to improve signal-to-noise ratio; however, this local grouping discards global structure and fails in low-light regimes where photon detections are sparse in space and time. In this work, we show that the information needed to recover both motion and illumination is encoded in correlations over the full space-time pattern of photon arrivals. Building on this insight, we develop a spatiotemporal flux probing theory and an algorithm that estimates the Fourier coefficients of the underlying intensity directly from the photon stream. We demonstrate that our approach (1) recovers fast motion and temporal illumination dynamics with substantially fewer photons than prior methods, (2) enables velocity-selective videography that automatically refocuses video onto specific detected motions, and (3) generalizes across sensing modalities including single-photon, event, and spike cameras.

\end{abstract}

\section{Introduction}

High-speed videography is fundamentally limited by photon scarcity: as exposure windows shrink to capture rapid motion or illumination changes, photon detections become extremely sparse in both space and time, making reliable intensity estimation difficult. Single-photon cameras provide a different imaging paradigm by recording individual photon arrivals with precise timestamps, allowing photon detections to be flexibly aggregated after acquisition~\cite{Morimoto2019MegapixelTS, ma2020quanta}. The key challenge is how to combine these sparse, asynchronous detections to recover a rapidly varying intensity.

Existing approaches for reconstructing high-speed video from single-photon data largely follow two strategies. Per-pixel temporal methods analyze each pixel independently, recovering time-varying intensity from photon arrival times~\cite{ingle2021passive, rapp_few_2016, wei2023passive}. Spatiotemporal pooling methods~\cite{ma2020quanta, sundar2025quanta, liu2024bit2bit, nousias2025opportunistic, lee2023caspi} improve photon efficiency by aggregating detections across local space-time neighborhoods; for example, collecting photons across motion trajectories of individual scene points~\cite{ma2020quanta}. However, both classes share the same fundamental premise: reconstruction begins by explicitly grouping photons locally in space, in time, or both. By confining estimation to local regions, existing methods discard long-range spatiotemporal correlations. Under extreme photon scarcity, these local windows may contain too few photons to reliably estimate intensity.

\ifshowteaser
\begin{figure}[t!]  %
\centering
\includegraphics[width=\textwidth]{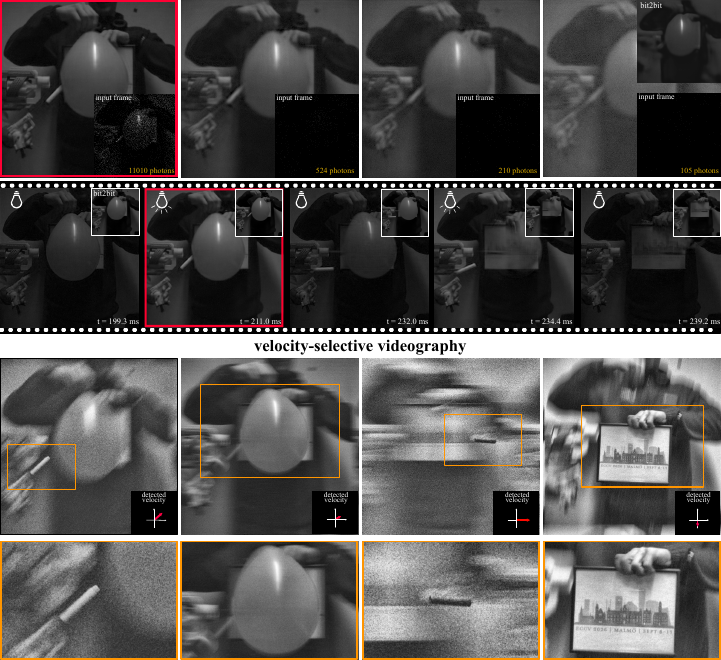}  %
\caption{\textbf{Top:} High-speed videography under extreme photon sparsity. Our method reconstructs high-quality 100~kfps video from binary photon detections or timestamp data, recovering both periodic illumination flicker (indicated by the lightbulb icons) and fast nonperiodic events (speeding bullets, balloon rupture) simultaneously.
\textbf{Bottom:} Velocity-selective videography. 
Our method detects the velocities of moving objects directly from the photon stream (arrows indicate magnitude and direction) and reconstructs video focused on a chosen velocity, blurring all other motion.}
\label{fig:teaser}
\end{figure}
\fi

In this work, we recover high-speed video from sparse photon data in regimes where photon detections arrive far apart in both space and time and local reconstruction methods start to break down. Our central insight is that, even under extreme photon scarcity, the information needed to recover motion and illumination is encoded in correlations across the full space-time pattern of photon arrivals. The key challenge is how to extract these correlations from sparse, asynchronous photon streams. To tackle this challenge, we develop a spatiotemporal flux probing theory that establishes a direct relation between the photon detections in space-time and the Fourier coefficients of the underlying intensity. Instead of aggregating photons within predefined neighborhoods, our approach estimates global spatiotemporal structure jointly from all detections.
 
Our work brings together two previously distinct regimes: per-pixel ultra-wideband temporal recovery~\cite{wei2023passive}, and spatial pooling for photon efficiency~\cite{ma2020quanta, nousias2025opportunistic}. Rather than using local space-time neighborhoods to estimate intensity, we leverage photon detections in a global spatiotemporal volume. This enables coherent aggregation of photons across space and time, increasing the effective signal-to-noise ratio (SNR). As a result, fewer photons are required to achieve a given reconstruction quality, allowing us to operate in regimes that are particularly challenging for existing methods: scenes exhibiting simultaneous fast motion and rapid illumination modulation under extreme photon sparsity in both space and time. Overall, we make the following contributions:
\begin{itemize}
	\item We develop a spatiotemporal flux probing theory that leverages photon detections across the full space-time observation volume, enabling joint recovery of fast motion and high-frequency illumination modulation directly from extremely sparse photon streams (Fig.~\ref{fig:teaser}, top).
	\item We introduce velocity-selective videography: by identifying motion-specific structure directly in the photon stream, we detect linear object velocities and reconstruct videos refocused at a chosen velocity while suppressing others (Fig.~\ref{fig:teaser}, bottom).
    \item We show that our theory extends across sensing modalities---including quanta sensors, asynchronous photon timestamp streams, spike cameras, and event cameras---suggesting a unified framework for reconstructing dynamic scenes from sparse asynchronous measurements.
\end{itemize}

\section{Single-Photon Videography}

\paragraph{Spatiotemporal flux function.}
The goal of videography is to recover the spatiotemporal flux $\spacetimeflux$~\cite{boyd1983radiometry} of light incident on the sensor.
The flux is the instantaneous photon arrival rate per unit area and time.
We assume that $\spacetimeflux$ has support over the spatiotemporal volume $(x,y,t) \in \setspacetime = [0,\width]\times[0,\height]\times[0,\exposure]$ defined by the bounded width $\width$ and height $\height$ of the sensor plane and the finite exposure time $\exposure$.

\paragraph{Stream of photon detections.}
Photon arrivals at the sensor are governed by a 3D inhomogeneous Poisson process with rate function $\spacetimeflux$~\cite{ross1995stochastic, goodman_statistical_1985}.
This process generates a stream of photon detection events $\events = \{(\spacex_i,\spacey_i,\timesym_i)\}_{i=1}^{N}$, from which we seek to recover $\spacetimeflux$. 
Due to recent advances in single-photon avalanche diode (SPAD) sensors, it is now feasible to obtain such event-based photon representations at up to picosecond temporal~\cite{Lin_2024, zappa2007principles, itzler2007single} and megapixel spatial resolution~\cite{Morimoto2019MegapixelTS}.

\paragraph{Flux constancy and local grouping.}
Conventional image sensors assume that $\spacetimeflux$ is locally constant within a bounded region of space and time---such as in a pixel area during the exposure.
Under this assumption, nonoverlapping spatiotemporal volumes can be modeled as independent homogeneous Poisson processes.
The flux in each region can then be estimated either by photon counting~\cite{ma2020quanta, sundar2023sodacam, sundar2024generalized, seets2021motion}, where the number of detections scales linearly with the flux, or by inter-photon imaging~\cite{ingle2021passive, ingle2019high, kirmani2014first, shin2016photon}, where the mean interval between photon arrivals is inversely proportional to the flux. Incorrectly assuming flux constancy in regions where the flux is changing results in motion blur or temporal aliasing. 

\paragraph{Videography by motion compensation.}
In dynamic scenes, flux is not constant within a fixed pixel because scene points move over time. Motion compensation aggregates photons along motion-aligned space-time trajectories; in this warped coordinate system, detections along each trajectory approximate a homogeneous Poisson process, enabling flux estimation using a homogeneous Poisson model. Learning-based methods adopt similar strategies (either implicitly or through explicit alignment)~\cite{sundar2025quanta, chennuri2024quanta, chennuri2025quanta, zhang2024streaming, jungerman2023panoramas, gupta2023eulerian}. However, this is still a locally confined formulation---now along trajectories rather than pixels. In our regime of extreme sparsity with simultaneous fast motion and temporal modulation, trajectories may contain too few detections for reliable estimation, and appearance changes along the path make motion recovery unreliable.

\paragraph{Videography by denoising.}
Another common approach is to treat reconstruction as a denoising problem~\cite{gap2024, liu2024bit2bit}. However, these methods still rely on local flux smoothness. In practice, 
we observe that denoising priors prioritize spatial consistency, effectively smoothing temporal variation to stabilize reconstruction (Fig.~\ref{fig:teaser}, top). We provide further analysis in supplement Section~E. 

\paragraph{Temporal modulation from ultrafast illumination.}
In the presence of ultrafast illumination, such as pulsed lasers and fluorescent light bulbs, the scene radiance can experience temporal modulation at very high frequencies ranging from 10~kHz to 60~MHz~\cite{hampf_satellite_2019, wang2021megahertz, lumentum2024qseries, amsOSRAM2024TMF8820}.
While this modulation can be useful~\cite{kitichotkul2025simultaneous, nousias2025opportunistic}, recovery of such time-varying radiance is highly challenging because flux constancy breaks down at time scales much faster than the inter-photon arrival time.
To tackle this problem, prior work recovers the time-varying flux of a pixel by estimating its Fourier coefficients from the photon timestamp realizations~\cite{wei2023passive}.
However, this method operates on each pixel independently, ignoring spatial correlations. As a result, these techniques fail when there are insufficient photons at each pixel, and spatial pooling is required to increase the signal-to-noise ratio~\cite{nousias2025opportunistic, lee2023caspi}. We refer to supplement Section~F for a detailed analysis.

\begin{figure}[t]
  \centering
  \includegraphics[width=0.9\textwidth]{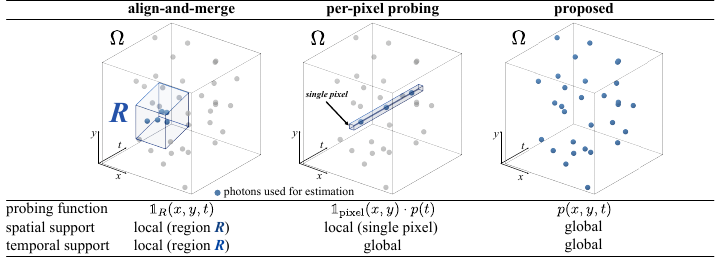}
  \caption{
  \textbf{Spatiotemporal support of existing and proposed methods.}
  Align-and-merge methods~\cite{ma2020quanta} aggregate photons within a local region~$R$, and per-pixel probing methods~\cite{wei2023passive} fix the spatial support to a single pixel while applying an arbitrary 1D temporal probe~$\probe(\timesym)$ to estimate its time-varying flux. Our framework admits a broader family of spatiotemporal probing functions~$\probespacetime$ that act on every photon globally.
  }
  \label{fig:probe-unify-compare}
\end{figure}

\paragraph{The gap: simultaneous ultrafast motion and illumination.}
We consider the imaging regime in which $\phi(x,y,t)$ varies rapidly due to both fast scene motion \emph{and} high-frequency illumination modulation. 
In this setting, the two classes of methods described above each fail in complementary ways.
Methods that rely on flux constancy cannot account for temporal intensity variations along a scene point's trajectory.
Conversely, per-pixel temporal methods that can handle such modulation fail to account for scene motion that extends beyond a single pixel's footprint.
As a scene point traverses a pixel, the observed temporal flux approximates a rect function with very high temporal bandwidth; in the low-photon-count regime, recovering such signals from few detections is severely ill-posed for pixel-wise methods~\cite{ingle2021passive, seets2021motion, wei2023passive}.
Both failures trace back to the same root cause: \emph{local spatiotemporal grouping of photons} (\cref{fig:probe-unify-compare}).

\section{Spatiotemporal Flux Probing}
\label{sec:stp-probing}

Spatiotemporal flux variations arise from two primary sources: temporal modulation of external illumination \cite{wei2023passive} and natural scene motion \cite{ma2020quanta}. Each produces a distinct signature in the underlying flux profile $\spacetimeflux$. Although we cannot directly observe $\spacetimeflux$, photon arrivals implicitly encode this structure. Here, we describe our method to extract this structure directly from photons, requiring only that $\spacetimeflux$ is well-behaved---\ie, nonnegative, bounded, and measurable (see supplement Section A). We consider the low-light regime, where photon detections are modeled by Poisson statistics.

\paragraph{Spatiotemporal flux probing.}
To recover different scene phenomena, we leverage spatiotemporal probing functions. A \textit{probing function} $\probespacetime$ is a deterministic function whose inner product with $\spacetimeflux$ informs the presence or absence of particular physical phenomena. However, this inner product---the \textit{flux probing integral}---cannot be computed directly because $\spacetimeflux$ is unknown. We prove in supplement Section B.2 that this projection can be estimated from the set of photon detections $\events$ through the probing measurements

\begin{equation}
    \underbrace{\sum_{(\spacex_i, \spacey_i, \timesym_i) \in \events} \probe(\spacex_i, \spacey_i, \timesym_i)}_{\text{probing measurements}} = \underbrace{\int_{0}^{\exposure}\int_0^\height\int_0^\width
    \probespacetime \, \spacetimeflux \,d\spacex\,d\spacey\,d\timesym}_{\text{flux probing integral }\langle\probe, \flux\rangle} +  \underbrace{M_\probe(\exposure)} _{\text{martingale noise}}.
    \label{eq:flux_probing_integral} 
\end{equation}

\noindent Intuitively, the probing measurements indicate how correlated the photon arrivals are with the probing function.

\paragraph{Noise model.} Approximation of the flux probing integral through probing is a noisy process due to randomness in photon arrivals~\cite{goodman_statistical_1985}. Equation (\ref{eq:flux_probing_integral}) decomposes the probing measurements into a deterministic integral and martingale noise~\cite{daley_introduction_2003}, which is a continuous stochastic process \cite{doob_stochastic_1991} caused by photon shot noise. We show in supplement Section B.3 that the probing measurements are approximately normal\footnote{This approximation holds with as few as 20 photons (supplement Section~I.1).} with mean equal to $\langle\probe, \flux\rangle$  and variance $\langle \probe^2, \flux\rangle$. As a result, probing measurements are an unbiased estimate of the flux probing integral. Characterization of the noise model enables statistical detection of the phenomena represented by different probing functions.
We also derive from first principles the decomposition in \cref{eq:flux_probing_integral} for spatiotemporal flux functions, extending prior work that considered temporal-only flux variations \cite{wei2023passive}. 

\paragraph{Spatiotemporal flux probing as a unifying lens.}
As illustrated in Fig.~\ref{fig:probe-unify-compare}, existing nonlearning methods can be thought of as special cases of spatiotemporal flux probing, differing only in their probing function and its support. Rather than fixing a local support or a restricted function class, our framework admits any bounded 3D probing function over the full spatiotemporal volume, so that every detected photon contributes jointly to estimation.

\subsection{Spatiotemporal Fourier Basis Functions}
\label{subsec:spatio_temporal_fourier_basis}
While our underlying probing theory is general, we consider the special case of 3D Fourier basis functions due to their inherent ability to simultaneously encode scene motion and illumination flicker. Specifically, a periodically flickering light source induces a frequency comb along the temporal frequency domain \cite{wei2023passive, nousias2025opportunistic}, whereas local linear motion of a bounded object generates a planar structure in the spatiotemporal frequency domain, with the plane's slope encoding object velocity \cite{Vernon2001FourierVS}. Detecting these physical phenomena corresponds to detecting the spatiotemporal frequencies $\freqvec$ associated with those phenomena.

\paragraph{Probing with Fourier basis.} A 3D Fourier basis function has the form $\probe_{\freqvec}(\spacetime)=e^{-j2\pi \freqvec^\top\spacetime}$, where $\spacetime=[\spacex, \spacey, \timesym]^\top$ is a point in space-time and $\freqvec=[\freqx, \freqy, \freqt]^\top$ is a spatiotemporal frequency. When probing with a single frequency $\freqvec$ of the Fourier basis, the resulting probing measurements

\begin{equation}
    \probemeas_{\freqvec} \triangleq \sum_{\spacetime \in \events} e^{-j2\pi \freqvec^{\top}\spacetime}
    \label{eq:fourier_probing}
\end{equation}

\noindent correspond to a phasor vector summation. If $\spacetimeflux$ contains frequency $\freqvec$, the phasors add coherently
and $|\probemeas_{\freqvec}|$ is large; if $\freqvec$ is absent, photon
arrivals are uncorrelated with the complex sinusoid, the phasors cancel out, and $|\probemeas_{\freqvec}|$ is small. However, due to the stochastic nature of photon arrivals, $|\probemeas_{\freqvec}|$ has nonzero fluctuations even if $\freqvec$ is absent. To detect a frequency $\freqvec$, the probing energy $|\probemeas_{\freqvec}|$ must be larger than the energy predicted by these noisy fluctuations.

\paragraph{Distribution of probing energy.} For 3D Fourier basis functions, we show in supplement Section B.4 that the probing measurements $\probemeas_\freqvec$ approximately follow a complex normal distribution with mean and covariance

\begin{equation}
    \boldsymbol{\mathrm{\mu}}
    =
    \begin{bmatrix}
    \left\langle \cos( 2\pi\freqvec^{\top}\spacetime), \flux(\spacetime) \right\rangle\\
    \left\langle -\sin(2\pi\freqvec^{\top}\spacetime), \phi(\spacetime) \right\rangle
    \end{bmatrix}, \,\,\,
    \boldsymbol{\mathrm{\Sigma}}
    =
    \begin{bmatrix}
    \left\langle \cos^2(2\pi\freqvec^{\top}\spacetime), \flux(\spacetime) \right\rangle & 0\\
    0 & \left\langle \sin^2(2\pi\freqvec^{\top}\spacetime), \flux(\spacetime) \right\rangle
    \end{bmatrix}.
    \label{eq:probing_distribution}
\end{equation}

\noindent The normalized probing energy

\begin{equation}
    |\probemeasnorm_{\freqvec}|^2 \triangleq 
        \text{Re}\left(\frac{\probemeas_{\freqvec}}{\sqrt{\boldsymbol{\mathrm{\Sigma}}_{1,1}}}\right)^2+\text{Im}\left(\frac{\probemeas_{\freqvec}}{{\sqrt{\boldsymbol{\mathrm{\Sigma}}_{2,2}}}}\right)^2
    \label{eq:fourier_energy}
\end{equation}

\noindent follows a noncentral $\chi^2$ distribution with two degrees of freedom, with noncentrality parameters determined by $\boldsymbol{\mathrm{\mu}}$ and $\boldsymbol{\mathrm{\Sigma}}$.

\paragraph{Frequency detection.} We use normalized Fourier probing functions $\probe_{\freqvec}(\spacetime)=\frac{1}{\sqrt{\spacevol}}e^{-j2\pi \freqvec^\top\spacetime}$ such that $\probe_{\freqvec}$ is orthonormal over the spatiotemporal volume $\setspacetime$, where $\spacevol=|\setspacetime|=\height\width\exposure$ represents the observation volume of photons. If frequency $\freqvec$ is absent from the flux, the probing measurement $\probemeas_{\freqvec}$ has zero mean. We can express this as a constant false alarm rate (CFAR) detector~\cite{scharf1991statistical}, where a frequency $\freqvec$ is detected if $|\probemeasnorm_{\freqvec}|^2 \geq \text{CDF}_{\chi^2_2}^{-1}(1-\alpha)$. Here, $\alpha$ denotes the significance value and false alarm probability. We show in supplement Section~B.5 that this relation can be written in terms of $|\probemeas_{\freqvec}|^2$ and critical value $\critvalue$ as follows:

\begin{equation}
\label{eq:CFAR3d}
|\probemeas_{\freqvec}|^2 \geq \underbrace{\text{CDF}_{\chi^2_2}^{-1}(1-\alpha) \frac{|\events|}{2\spacevol}}_{\critvalue}.
\end{equation}

\subsection{Velocity Detection}
\label{subsec:motion_detection}

Because linear motion induces planar structure in the spatiotemporal spectrum, we can use the probed spatiotemporal frequencies for motion analysis~\cite{Vernon2001FourierVS}. Consider a local patch whose intensity $I$ translates with approximately constant velocity $\mathbf{v}=[v_x, v_y]^\top$, such that $\spacetimeflux=I(x-v_xt, y-v_yt)$. Here, the spectral energy of $\spacetimeflux$ due to the patch motion is confined to the plane 

\begin{equation}
    v_xf_x + v_yf_y + f_t =0 
\label{eq:velocity_plane}
\end{equation}

\noindent in the spatiotemporal frequency domain \cite{Vernon2001FourierVS}. Thus, estimating linear motion reduces to identifying planar concentrations of energy in 3D frequency space. 

We formulate linear velocity detection as a plane-consistency problem. First, we scan a discrete set of $M$ velocity hypotheses $\{\mathbf{v}_i\}_{i=1}^M$. For each hypothesis, we measure the support of the corresponding velocity plane by accumulating the energy of the detected frequencies lying near that plane

\begin{equation}
E_i \;\triangleq\; \sum_{\mathbf{f}\,:\,| \mathbf{v}_i'^\top \mathbf{f}|
\,\leq\,\epsilon} |\probemeas_{\freqvec}|^2,
\label{eq:plane_score}
\end{equation}

\noindent where $\mathbf{v}'=[v_x, v_y,1]^\top$ and $\epsilon$ controls the tolerance to deviations from planarity. 
Then, velocity planes are detected based on their accumulated energies. Unlike the probing measurement, the plane scores do not follow a simple parametric
distribution because (1) the cumulative energy aggregates the energy across many different frequencies, and
(2) neighboring velocity hypotheses have overlapping frequencies, making their energies correlated. As a result, we use a nonparametric rank-CFAR test~\cite{HansenOlsen1971SignTestCFAR} to detect velocity planes. For each $\mathbf{v}_i$,
we compute its rank $r_i \triangleq |\{j \in \mathcal{N}_i : E_j < E_i\}|$ among
the $|\mathcal{N}_i|$ neighboring hypotheses,
and detect $\mathbf{v}_i$ only if $E_i$ falls in the top $\cfarsigvalvel$ fraction,
\begin{equation}
r_i \;\geq\; \lceil(1-\cfarsigvalvel)(|\mathcal{N}_i|+1)\rceil,
\label{eq:rank_cfar}
\end{equation}
which targets a false alarm rate $\cfarsigvalvel$. See supplement Section~C for derivations.

\subsection{Extension to Other Asynchronous Sensors}

Our probing theory can be extended beyond single-photon sensors to other asynchronous sensors whose events stochastically encode a continuous intensity signal. The only requirement is that the image formation model and noise properties of the sensor must be known. We discuss two cases here---spike and event cameras---and defer derivations to supplement Section~D.

\paragraph{Spike cameras.} Spike cameras output an event when the accumulated flux at a pixel $\int \flux(t)dt$ crosses a fixed threshold \cite{huang_1000_2023}. Therefore, the instantaneous spike rate is proportional to the flux $\spacetimeflux$ and spatiotemporal flux probing with the spike events yields a noisy estimate of the spatiotemporal Fourier coefficients of $\spacetimeflux$, up to a scale factor.

\paragraph{Event cameras.} Event cameras output an event with polarity $\eventpol=\pm1$ when the change in log-intensity $\log \spacetimeflux$ at a pixel exceeds a contrast threshold \cite{lichtsteiner_128times_2008}. Fourier probing on these events $(\spacex_i,\spacey_i,\timesym_i)\in\events$ is done via the following sum
\begin{equation}
\probemeas_\freqvec = \sum_{(\spacex_i,\spacey_i,\timesym_i)\in\events}\eventpol_i\probe_\freqvec(\spacex_i,\spacey_i,\timesym_i).
\label{eq:event-probing}
\end{equation}
This yields a noisy estimate of the Fourier coefficients of $\partial_t \log \spacetimeflux$ up to the scale of contrast threshold $C$.

\section{Videography by Spatiotemporal Flux Probing}
\label{sec:videography}
\begin{figure}[t]
    \centering
    \includegraphics[width=\textwidth]{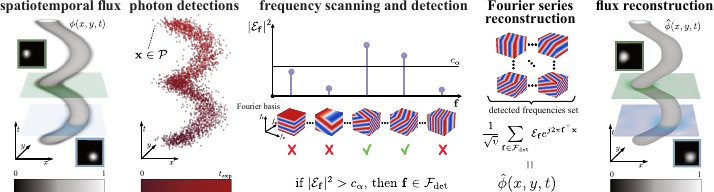}
    \caption{Overview of spatiotemporal flux probing.
    \textit{Spatiotemporal flux:} The goal of videography is to recover the spatiotemporal flux function $\spacetimeflux$ (insets visualize two time slices).
    \textit{Photon detections:} The observed photon detections $\events$ are a stochastic realization of an inhomogeneous Poisson process with rate $\spacetimeflux$. 
    \textit{Frequency scanning and detection:} We compute 3D Fourier probing measurements on $\events$ via Eq.~(\ref{eq:fourier_probing}) and declare $\freqvec$ detected (in $\freqscfar$) if its probing energy $|\probemeas_{\freqvec}|^2$ exceeds $c_{\alpha}$ (set for a target false alarm rate $\alpha$).
     \textit{Fourier series reconstruction:} We obtain $\hat{\phi}(x,y,t)$ by performing an inverse Fourier transform on the set of detected frequencies $\mathcal{F}_{\mathrm{det}}$.
    }
    \label{fig:methods}
\end{figure}

Our theory enables videography across asynchronous modalities such as single-photon, spike, and event cameras. We discuss the algorithms for each modality, with full implementation details in supplement Section~G.

\paragraph{Single-photon videography.}
\begin{wrapfigure}{r}{0.55\textwidth}
    \vspace{-\intextsep}
    \vspace{-\topskip}
    \vspace{-\baselineskip}
    \begin{minipage}{\linewidth} %
    \begin{algorithm}[H]
    \caption{Flux reconstruction}
    \label{alg:flux_recon}
    \footnotesize            %
    \begin{algorithmic}
        \Procedure{FluxRecon}{$\events, \spacevol, \alpha$}
            \State \textbf{// Frequency scanning}
            \State $\freqsall \gets$ $\{\freqvec_i\}_{i=1}^K$ up to Nyquist
                   \Comment{Supp.~G.1.1}
            \For{$\freqvec \in \freqsall$}
                \State $\probemeas_{\freqvec} \gets \frac{1}{\sqrt{v}}
                       \sum_{\spacetime \in \events}
                       e^{-j 2\pi \freqvec^\top \spacetime}$
                    \Comment{\cref{eq:fourier_probing}}
            \EndFor
            \State \textbf{// CFAR Detection}
            \State $\critvalue \gets \text{CDF}^{-1}_{\chi^2_2}\!(1-\alpha)
                   \cdot \frac{|\events|}{2\spacevol}$
                    \Comment{\cref{eq:CFAR3d}}
            \State $\freqscfar \gets \{\freqvec \mid
                   |\probemeas_{\freqvec}|^2 \geq \critvalue\}$
            \State \textbf{// Reconstruction}
            \State $\hat{\flux}(\spacetime) \gets \frac{1}{\sqrt{v}}
                   \sum_{\freqvec \in \freqscfar}
                   \probemeas_{\freqvec}\, e^{j 2\pi \freqvec^\top \spacetime}$
            \State \Return $\hat{\flux}$
        \EndProcedure
    \end{algorithmic}
    \end{algorithm}
    \end{minipage}
    \vspace{-\intextsep}
\end{wrapfigure}
Using our probing theory from Section~\ref{subsec:spatio_temporal_fourier_basis}, we can
recover the spatiotemporal flux $\spacetimeflux$ directly from a set of photon detections $\events$ in three steps, summarized in~\cref{alg:flux_recon}. First, we estimate $\probemeas_\freqvec$ from \cref{eq:fourier_probing} for all frequencies in a discrete spatiotemporal Fourier basis $\freqsall=\{\freqvec_1,\dots,\freqvec_K\}$---restricted to those frequencies recoverable under the given spatial and temporal sampling rates---and obtain the corresponding probing coefficients $\{\probemeas_{\freqvec_1},\dots, \probemeas_{\freqvec_K}\}$. Second, we apply the CFAR detection rule in \cref{eq:CFAR3d} to identify the set of detected frequencies $\freqscfar$. Finally,
we recover the spatiotemporal flux by Fourier series reconstruction over the detected frequencies.
The reconstructed flux can then be resampled at arbitrary temporal resolutions. An illustration of the pipeline is shown in \cref{fig:methods}.

\paragraph{Velocity-selective videography.}
Because we have an explicit mechanism for velocity detection (Section \ref{subsec:motion_detection}), we can reconstruct the video to focus on a particular linear velocity while blurring away others. This ``velocity focusing'' is performed directly in the spatiotemporal frequency domain. Given a detected velocity $\mathbf{v}_{\text{det}}$, we focus on that linear motion by shearing the 3D Fourier spectrum via the change of variables $f_t \leftarrow f_t+\mathbf{v}_{\text{det}}^\top \mathbf{f}_{x, y}$, with $\mathbf{f}_{x, y}=[f_x, f_y]^\top$. 
This shear maps the corresponding velocity plane to the temporal DC component, making linear motion at velocity $\mathbf{v}_{\text{det}}$ appear static in the reconstructed volume. Reconstruction is then performed in a sliding temporal window over the photon stream to generate video frames. Although related ideas have been explored in conventional videography~\cite{Vernon2001FourierVS} and single-photon imaging~\cite{sundar2023sodacam}, our method performs velocity focusing at ultra-low light levels and without requiring manual specification of the motion direction or magnitude as in previous work~\cite{sundar2023sodacam}.

\paragraph{Spike camera videography.} We compute the probing measurements from the spike stream and estimate the Fourier coefficients of the flux. However, in regions of constant flux, spike streams generate highly periodic spike trains, which introduce strong artificial high-frequency peaks. To suppress these spurious components, we apply a low-pass filter in addition to CFAR detection.

\paragraph{Event camera videography.}
The probing measurements of an event stream yield the Fourier coefficients of the temporal derivative of the log flux, $\partial_t \log \spacetimeflux$. To recover the log-intensity, we integrate this estimate of the derivative in the frequency domain by dividing the spectrum by $j 2\pi \freqt$ for $\freqt \neq 0$. In practice, low-frequency temporal components are generally dominated by bias and drift \cite{scheerlinck_continuous-time_2018}, so a high-pass filter is applied in addition to CFAR detection. The resulting reconstruction is determined up to an additive constant in log-intensity. This constant and lost low-frequency information can be obtained from intensity frames recorded alongside the events, which we fuse together using a Kalman filter \cite{kalman_new_1960}. 

\begin{figure*}[p]
  \centering
  \includegraphics[width=\textwidth]{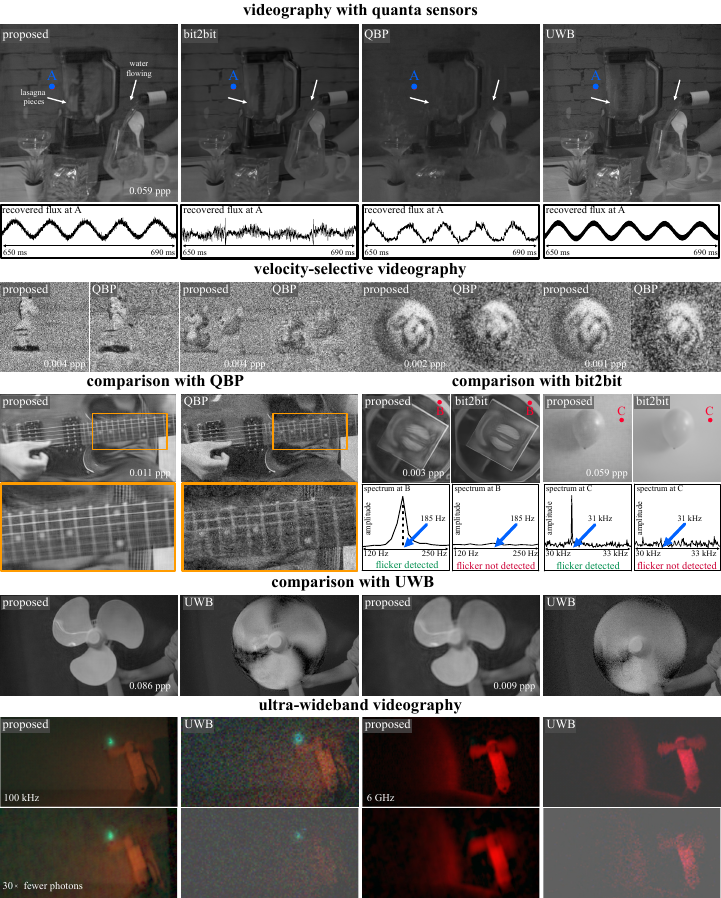}
  \caption{Spatiotemporal flux probing results. \textbf{Row 1}: Videography with quanta sensors. We reconstruct 100~kfps video of a scene with complex dynamics under 120~Hz illumination flicker. Prior methods fail to recover motion and flicker simultaneously; ours captures both. \textbf{Row 2:} Velocity-selective videography. Our method detects the velocities of multiple objects directly from the photon stream and produces higher-quality reconstructions than QBP~\cite{ma2020quanta}. \textbf{Row 3:} Comparison with QBP (left) and bit2bit~\cite{liu2024bit2bit} (right) on their respective datasets. Our method produces higher-quality reconstructions while simultaneously recovering the illumination flicker. \textbf{Row 4:} Comparison with UWB. Even in periodic motion scenarios favorable to UWB~\cite{wei2023passive}, our approach achieves higher-quality reconstructions at two different light levels. \textbf{Row 5:} Ultra-wideband videography. Our approach can recover the fan blade and the laser wavefront even with 30$\times$ fewer photons than previous work~\cite{wei2023passive}.}

  \label{fig:results}
\end{figure*}

\section{Results}
\label{sec:results}

We validate our theory on real captures from multiple sensing modalities. Our experiments demonstrate four capabilities:
(1) high-speed video recovery of dynamic scenes with simultaneous temporal flicker under extreme photon sparsity, a regime in which existing methods struggle,
(2) velocity-selective videography, where detected velocity planes enable reconstruction focused on individual moving objects,
(3) ultra-wideband videography in the GHz regime, achieved at 30$\times$ lower photon budgets than previous approaches, and
(4) extension of our spatiotemporal probing framework to event and spike cameras.
We provide additional results and videos in the supplementary materials. Supplement Sections H and I cover further experimental details and simulated-data results, respectively.

\paragraph{Sensor hardware.}
We use multiple sensing modalities---both synchronous and asynchronous---to validate our theory. For synchronous capture, we use a $512\times512$-pixel SPAD512 camera~\cite{spad512_datasheet} operating at $100$ kHz, which outputs binary frames where a value of $1$ indicates a photon detection. For this modality, we report light levels in photons per pixel (ppp)~\cite{ma2020quanta}.
For asynchronous sensing, we use the dataset from Wei et al.~\cite{wei2023passive}, acquired with a single-pixel SPAD providing 68~ps timing resolution and outputting a stream of photon timestamps. 
We additionally evaluate our framework on data from a DAVIS240C event camera and a spike camera using publicly available datasets~\cite{mueggler_event-camera_2017, zhu_retina-like_2020}.

\paragraph{Baselines.} We compare against several state-of-the-art techniques in high-speed single-photon imaging. \textit{Ultra-wideband imaging} (UWB) recovers temporal modulations at a pixel through flux probing in the time dimension only \cite{wei2023passive}. \textit{Quanta burst photography (QBP)} uses an ``align-and-merge'' strategy to aggregate photons along motion trajectories of scene points \cite{ma2020quanta}. \textit{bit2bit} is a self-supervised denoising technique that leverages the inductive smoothness bias of neural networks \cite{liu2024bit2bit}. Of these techniques, only UWB and our method are able to handle both asynchronous and synchronous photon streams. QBP and bit2bit can only handle synchronous frame data. Furthermore, these techniques produce qualitatively inferior results compared to our method, due to their limited ability to leverage global spatiotemporal photon correlations.

\paragraph{High-speed videography under extreme photon sparsity.} 
We demonstrate 100~kfps videography of two foam bullets striking and rupturing a balloon captured with a SPAD512 camera (Fig.~\ref{fig:teaser}). The scene is illuminated by a 120~Hz ceiling light and a 31~kHz modulated LED. Our method reconstructs the projectile trajectories, the balloon rupture, and both illumination flickers simultaneously. A visualization of the 31~kHz flicker is provided in supplement Section H.1.1.
Even under extreme photon sparsity, our approach continues to recover the white bullet trajectory and the ambient illumination flicker, demonstrating robust high-speed videography in photon-starved conditions. 

We further reconstruct a 100~kfps video of a scene in which lasagna is blended in the background while water is poured in the foreground under 120~Hz AC illumination. Despite photon-starved conditions, our method recovers fine spatial details of the flowing water and blender contents while faithfully capturing the global 120~Hz flicker (Fig.~\ref{fig:results}, row 1).
In contrast, methods based on local neighborhoods either smooth temporal modulation into spatial structure (\eg, bit2bit) or produce unstable modulation (\eg, QBP), while per-pixel approaches (\eg, UWB) fail to capture the dynamic motion.

\paragraph{Velocity-selective videography.}
We apply our velocity detection (Section~\ref{subsec:motion_detection}) to the scene shown in Fig.~\ref{fig:teaser}. Our method detects the velocities of multiple objects in the scene and reconstructs separate videos focused on each (Fig.~\ref{fig:teaser}, bottom), even for the gray high-speed foam bullet. To our knowledge, this is the first demonstration of post-capture velocity-selective video reconstruction from a single-photon stream with automatic velocity detection.

We evaluate velocity detection on the toys dataset from \cite{ma2020quanta}. We synthetically thin the timestamps to simulate extreme low-light conditions. Even under severe thinning, our method reliably detects object velocities and produces higher-quality reconstructions than QBP (Fig.~\ref{fig:results}, row 2, left). In this regime, local photon accumulation becomes unreliable, whereas velocity-aware aggregation provides a strong structural prior.
We also captured a scene of falling balls under AC illumination flicker, introducing simultaneous motion and high-frequency modulation. Our method accurately detects the ball’s velocity and reconstructs fine spatial detail despite extreme sparsity and flicker. In contrast, QBP blurs temporally varying intensity (Fig.~\ref{fig:results}, row 2, right). 

\paragraph{Comparison with baselines.} We evaluate our method against QBP on the guitar dataset~\cite{ma2020quanta} (Fig.~\ref{fig:results}, row 3, left). Our method is able to recover the fine features of the guitar strings at 0.011 ppp, whereas QBP produces noisy reconstructions. At this level of photon sparsity, local temporal aggregation cannot improve intensity estimates since there are not enough
photons to accumulate. 

We  compare with bit2bit on the Mandrill dataset~\cite{liu2024bit2bit}. Our approach resolves finer spatial details (Fig.~\ref{fig:results}, row 3, right) and faithfully recovers the faint CPU fan flicker present in the data. In contrast, bit2bit oversmooths both spatial texture and temporal variation, failing to capture the subtle flicker.
We additionally captured a scene of a static balloon illuminated by a localized 31~kHz sinusoidally modulated LED. Our method successfully recovers the high-frequency modulation, whereas bit2bit suppresses it as noise.

Finally, using the SPAD512 camera, we captured a scene of a fan rotating at approximately 15~Hz and compare against UWB. Although the motion is periodic---typically a favorable setting for UWB’s periodic priors---the rotating blades induce high temporal bandwidth in the flux at each pixel. Under extreme photon sparsity, UWB fails to recover this bandwidth due to the low per-pixel SNR and cannot reconstruct the blade structure (Fig.~\ref{fig:results}, row 4).
In contrast, our method accurately recovers both the fan blades and their specular reflections, even with 10$\times$ fewer detected photons, where UWB completely breaks down. This experiment demonstrates the advantage of estimating global spatiotemporal structure rather than treating each pixel’s temporal signal independently.

\paragraph{Ultra-wideband videography.} We evaluate our method on the same dataset used by Wei et al.~\cite{wei2023passive}, where a diffused laser pulse illuminates a scene containing a rotating fan. Our approach successfully reconstructs both the laser wavefront and the spinning fan blades (Fig.~\ref{fig:results}, row 5, top). In contrast, UWB blurs the fan motion at lower light levels due to the limited number of photons per pixel. Notably, our method yields high-quality reconstructions even with 30$\times$ less light (Fig.~\ref{fig:results}, row 5, bottom), demonstrating the advantage of exploiting global spatiotemporal correlations for reconstructing ultra-wideband videos.
\begin{figure}[t]
  \centering
  \includegraphics[width=\textwidth]{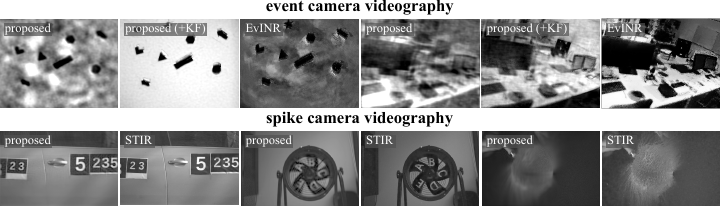}
  \caption{Videography with asynchronous sensors. \textbf{Top:} 
  Our method reconstructs 1~kfps video from raw event streams, achieving 
  competitive results with the learning-based method EvINR~\cite{wang_revisit_2024}. \textbf{Bottom:} We reconstruct 
  10~kfps video from binary spike streams with quality on par with STIR~\cite{fan_spatio-temporal_2024}.}
  \label{fig:eventspike-results} 
\end{figure}

\paragraph{Videography with asynchronous sensors.} 
We further demonstrate videography results on event and spike cameras. 
For event cameras, we evaluate on the dataset from~\cite{mueggler_event-camera_2017} captured with a DAVIS240C sensor. Direct integration of events alone produces blurred reconstructions due to accumulated drift and the lack of absolute intensity information. By combining our frequency-domain estimation of $\partial_t \log \spacetimeflux$ with an intensity image via Kalman filtering~\cite{kalman_new_1960}, we obtain reconstructions that preserve both spatial detail and temporal dynamics (Fig.~\ref{fig:eventspike-results}, top). The resulting quality is comparable to EvINR~\cite{wang_revisit_2024}, a state-of-the-art learning-based method for event-based videography.
We also evaluate our method on the spike camera dataset from~\cite{zhu_retina-like_2020}, covering scenes with fast (moving car), periodic (fan spinning), and highly nonperiodic motion (water balloon rupture) (Fig.~\ref{fig:eventspike-results}, bottom). Our reconstructions remain visually consistent and detailed across these varied dynamics. For reference, we compare against the state-of-the-art learning-based method STIR~\cite{fan_spatio-temporal_2024}.

Importantly, the core probing formulation transfers across quanta sensors, SPAD timestamp streams, event cameras, and spike cameras, while the image-formation models and reconstruction steps remain sensor-specific. To our knowledge, no existing method applies across all these sensing modalities within a single unified framework.

\section{Conclusion}

High-speed videography remains fundamentally limited by photon scarcity; our work does not change that physical constraint. What it changes is how the available photons are used. Existing methods confine estimation to local spatiotemporal supports, discarding information encoded in the broader pattern of photon arrivals. We have shown that the global structure of \emph{where and when} photons arrive carries sufficient information to jointly recover fast motion and rapid illumination dynamics, even in regimes where prior methods degrade.
We hope that viewing the full spatiotemporal photon stream as a coherent signal, rather than a collection of isolated detections, will inspire new approaches to imaging under extreme photon sparsity.

\subsection{Limitations}

Our current implementation has two main limitations. First, because we use global Fourier probing functions, sharp spatial edges and temporal discontinuities can spread energy across Fourier coefficients and produce Gibbs-like ringing artifacts in the reconstruction~\cite{harris_use_1978, gonzalez_digital_2018}. This limitation arises from the specific choice of probing basis rather than from our spatiotemporal probing theory itself. Since our theory only requires linear projections of the photon stream, alternative bases with better spatial or temporal localization---such as wavelets~\cite{mallat_theory_1989}---could reduce these artifacts without changing the underlying formulation.

Second, our velocity-selective videography assumes approximately constant linear motion over the reconstruction window. When motion is highly nonlinear, accelerating, or chaotic---for example in a balloon rupture or fluid splash---that concentration breaks down, leading to reduced selectivity and reconstruction blur. Extending the framework to richer motion models, such as higher-order kinematics using polynomial-phase~\cite{peleg_multicomponent_1996} or chirplet~\cite{mann_chirplet_1995} probing functions, is an important future direction.

\subsection{Future Directions}
There are several promising directions for future work. First, our formulation suggests a unified perspective on videography from asynchronous sensors. Whether events correspond to first-photon arrivals, intensity changes, or thresholded flux accumulation, the probing measurements reduce to sums over event timestamps. This provides a common mathematical framework for comparing sensing modalities in terms of the spatiotemporal information they preserve and the reconstruction quality they support under bandwidth constraints~\cite{sundar2023sodacam, sundar2024generalized}.

Second, our velocity-selective videography currently recovers image-plane motion, but these estimates could also support geometric inference. Under known camera motion, stereo, or multi-view capture, the recovered spectral shear may provide a route to depth estimation~\cite{scharstein_taxonomy_2002} and structure-from-motion~\cite{jungerman_radiance_2025} directly from sparse photon streams, linking our framework to core problems in dynamic 3D scene understanding in extreme low-light conditions.

Third, our framework may be useful for active and scientific imaging. In particular, the ability to handle fast motion together with high-frequency illumination changes could be valuable for single-photon lidar and Doppler imaging~\cite{kitichotkul2025simultaneous}. More broadly, because our method analyzes structure jointly in space and time, it may also be useful for vibration sensing \cite{sheinin_dual-shutter_2022, davis_visual_2014}, where subtle periodic motions must be recovered from extremely faint intensity signals.

\section*{Acknowledgements}

We are grateful to Kyros Kutulakos and David Lindell for providing the SPAD512 camera. Kyros Kutulakos acknowledges the support of NSERC under RTI-2023-00153.
We also thank the Marine Physical Laboratory at the Scripps Institution of Oceanography for providing access to their compute resources.

\bibliographystyle{splncs04} 
\bibliography{main} 

\end{document}


\maketitle
\let\svthefootnote\thefootnote
\let\thefootnote\relax\footnotetext{$^{*}$ Equal contribution: \texttt{\{yan569,mforlive\}@purdue.edu}}
\let\thefootnote\svthefootnote

{\small
\begin{spacing}{1.5} 
\hypersetup{linkcolor=black}
\tableofcontents
\end{spacing}
}
\clearpage

\appendix

\section{Preliminaries}

\textbf{Photon arrivals as a spatiotemporal point process.} While camera sensors physically consist of a discrete array of photodetectors, the photon arrival process can be modeled as a continuous process in space and time, registering the precise location and arrival time of each photon arrival event~\cite{goodman_statistical_1985}. Under this framework, over an exposure time $\exposure$, photon arrivals are modeled as an inhomogeneous Poisson point process $\events$ over the spatiotemporal domain $\setspacetime=\setspace\times[0,\exposure]$, where $\setspace=[0,\width]\times[0,\height]$ is the spatial area of the detector. Its underlying flux is given by $\flux$.

\textbf{Spatiotemporal flux.} While flux is normally modeled in photons/second as a temporal function where $\flux(\timesym)\geq 0$, this can be extended to the spatiotemporal domain. For a point $\spacetime=(\spacex, \spacey,\timesym)$, its instantaneous flux can be written as a function in spacetime, where $\flux(\spacetime)\geq0$. In terms of physical sensor units, this flux can be interpreted as the number of photon arrivals in the units of photons/second/pixel$^2$.

Importantly, we restrict the spatiotemporal flux functions that we consider to those that are \textbf{nonnegative}, \textbf{measurable}, and \textbf{bounded}. We also assume that we observe photon arrivals in a \textbf{finite} sensor area with finite exposure time. These are critical assumptions that allow spatiotemporal flux probing to hold mathematically.

\subsection{Mathematical definitions}

In order to rigorously characterize photon arrivals in space and time, we include definitions of the foundational measure-theoretic and probabilistic structures used throughout this work. These definitions formalize the notion that photons arrive randomly across both space and time, and that our uncertainty about future arrivals can be tracked as observations accumulate.

\begin{definition}[$\sigma$-algebra (Section \textbf{A1.4} in~\cite{daley_introduction_2003})]
A $\sigma$-algebra $\filt$ on a set $\setspacetime$ is a collection of subsets of $\setspacetime$ that contains the empty set and is closed under complement and countable unions. The pair $(\setspacetime, \filt)$ is called a measurable space.

\noindent In the context of photon arrivals, $\filt$ formalizes all spatiotemporal regions in $\setspacetime$ that are well-defined and observable---for example, the region corresponding to a given pixel area during a given time interval. Together, the elements of $\filt$ represent every possible collection of photon arrival events that a sensor could observe over its exposure.
\end{definition}

\begin{definition}[Measurability (Section \textbf{A1.4} in~\cite{daley_introduction_2003})]
Let $(\setspacetime, \filt)$ be a measurable space.
\begin{enumerate}[label=(\roman*)]
    \item A set $\setspacetimesub \subseteq \setspacetime$ is measurable if $\setspacetimesub \in \filt$.
    \item A function $f : \setspacetime \to \mathbb{R}$ is measurable if for any measurable set $B\subseteq\mathbb{R}$, the set $f^{-1}(B) = \{\spacetime \in \setspacetime : f(\spacetime) \in B\}$ is measurable.
\end{enumerate}

\noindent Measurability of flux function $\flux$ ensures that 
$\int_{\setspacetimesub} \flux(\spacetime)\,d\spacetime$ is well-defined 
for any measurable $\setspacetimesub \subseteq \setspacetime$, so that 
the expected number of photon arrivals in any observable region can be 
computed via integration against $\flux$.
\end{definition}

\subsubsection{Spatiotemporal Poisson point process definitions}

\begin{definition}[Spatiotemporal filtration]
A spatiotemporal filtration is a family $\{\filt_t\}_{t\in[0,\exposure]}$ of $\sigma$-algebras on $\Omega$, indexed by time $t$, such that $\filt_s \subseteq \filt_t$ for all $s \leq t$. Each $\filt_t$ encodes all observable information accumulated up to time $t$, across the full sensor domain $\setspace\subset \mathbb{R}^2$.

\noindent Intuitively, the filtration represents the growing history of photon arrivals on the sensor: as the exposure progresses, more photons have landed and $\filt_t$ grows to reflect all spatial arrival patterns observed up to time $t$.
\end{definition}

\begin{definition}[Spatiotemporal adapted process (Definition \textbf{3.2.14} in~\cite{jeanblanc_mathematical_2009})]
A stochastic process $\{X_t\}_{t\in[0,\exposure]}$, where each $X_t$ is a spatially-indexed random field over $\setspace$, is adapted to the filtration $\{\filt_t\}$ if for each $t$, the random field $X_t(\cdot)$ is $\filt_t$-measurable. That is, the value of $X_t$ at any spatial location $\spacexy \in \setspace$ is fully determined by the information available up to time $t$.

\noindent For photon arrivals, adaptedness means that our description of the arrival process at any spatial location on the sensor at time $t$ depends only on what has actually been observed up to that moment---not on future photon arrivals.
\end{definition}

\begin{definition}[Spatiotemporal $\filt$-martingale]
Let $\{\filt_t\}_{t\in[0,\exposure]}$ be a filtration. A process $M(\timesym) = M(\setspace \times [0,\timesym])$, aggregated over sensor domain $\setspace \subset \mathbb{R}^2$, 
is an $\filt$-martingale if it is $\filt$-adapted and:
\begin{enumerate}[label=(\roman*)]
    \item $\mathbb{E}[|M(\timesym)|] < \infty$ for all $t \in [0, \exposure]$,
    \item $\mathbb{E}[M(\timesym) \mid \filt_s] = M(s)$ \text{ a.s. for all } $s \leq t$.
\end{enumerate}

\noindent Intuitively, a spatiotemporal martingale describes a quantity defined across the sensor whose expected future value at any pixel location, given everything observed so far, is simply its current value. In the context of photon arrivals, a martingale arises when we correct the raw arrival counts by their expected flux contribution, leaving a residual process that carries no predictable trend forward in time.
\end{definition}

\begin{definition}[Inhomogeneous Poisson point process]
A spatiotemporal point process $\events$ on $\setspacetime=\setspace\times[0,\exposure] \subset \mathbb{R}^2\times\mathbb{R}^+$ 
is an inhomogeneous Poisson point process with flux function 
$\flux : \setspacetime \to \mathbb{R}^+$ if:
\begin{enumerate}[label=(\roman*)]
    \item For any measurable $\setspacetimesub \subseteq \setspacetime$, the count $\events(\setspacetimesub) \sim \mathrm{Poisson}(\fluxmeas(\setspacetimesub))$, where the intensity measure is
    \begin{equation}
        \fluxmeas(\setspacetimesub) = \int_\setspacetimesub \flux(\spacetime)\, d\spacetime
    \end{equation}
    \item For disjoint $\setspacetimesub_1, \ldots, \setspacetimesub_k \subseteq \setspacetime$, the counts $\events(\setspacetimesub_1), \ldots, \events(\setspacetimesub_k)$ are independent.
\end{enumerate}

\noindent For simplicity, we denote the total number of observed photon arrivals as $|\events| = \events(\setspacetime)$.

\noindent This models photons arriving at a sensor as random events in space and time, where the local flux $\flux(\spacexy, t)$ governs how frequently photons tend to land near location $\spacexy$ at time $t$. Brighter regions of a scene produce higher flux and therefore more photon arrivals, and arrivals in nonoverlapping regions or time intervals are statistically independent of one another.
\end{definition}

\clearpage
\section{Spatiotemporal Flux Probing Theory}

\subsection{Spatiotemporal photon counting process}

We start by showing the martingale decomposition of Poisson point process $\events$.

\begin{proposition}[Poisson point process decomposition]
Let $\events(\setspacetime)=\events(\setspace\times[0,\exposure])$ be the inhomogeneous Poisson point process with flux function $\flux(\spacetime)$, and let $\filt$ be a filtration such that $\events(\setspacetime)$ is $\filt$-adapted. Then for any measurable subset $\setspacetimesub = \setspace\times [0,\timesym] \subseteq \setspacetime$ where $0< \timesym \leq \exposure$,
    \begin{equation}
        M(t) 
        = \events(\setspacetimesub) 
        - \int_{\setspacetimesub} \flux(\spacetime)\,d\spacetime,
    \end{equation}
    where $M(\timesym) = M(\setspace \times [0,\timesym])$.

    In the special case that $\events$ is a 1D inhomogeneous Poisson point process, such that it is only defined on $[0,\exposure]$, this decomposition can instead be written in terms of counting process $N(t)=\events([0,t])$~\cite{wei2023passive}:
    \begin{equation}
        M(t) 
        = N(t) 
        - \int_0^t \flux(u)\,du.
        \label{eq:flux-decomp-1d}
    \end{equation}
    \label{prop:flux-decomp}
\end{proposition}

\begin{proof}[Proof Sketch of~\cref{prop:flux-decomp}]
The proof proceeds in two steps:

\textbf{Step 1:} We show the Poisson process $\events$ can be decomposed in 
terms of its flux $\flux(\spacetime)$ 
and zero-mean noise $M$. To do this, we make use of the following lemma:

\begin{lemma}[Poisson compensated integral (Equation \textbf{12.4} 
in~\cite{last_lectures_2017})]\label{lem:comp-integ}
    Define Poisson point process $\events$ with flux function $\flux$ on the domain $\setspacetime=\setspace\times[0,\exposure]$. For any deterministic function $f$ such that $\int_\setspacetime |f(\spacetime)\flux(\spacetime)|\,d\spacetime < \infty$, then $\int_\setspacetimesub |f(\spacetime)\flux(\spacetime)|\,d\spacetime < \infty$ on any measurable subset $\setspacetimesub = \setspace\times [0,\timesym] \subseteq \setspacetime$. The compensated integral of $f$ with respect to $\events$ is defined by
    \begin{equation}
        \int_{\setspacetimesub} f(\spacetime)\,dM(\spacetime) 
        = \int_{\setspacetimesub} f(\spacetime)\,d\events(\spacetime) 
        - \int_{\setspacetimesub} f(\spacetime)\,\flux(\spacetime)\,d\spacetime.
    \end{equation}
    
    such that $\mathbb{E}\left[\int_\setspacetimesub f(\spacetime)\,dM(\spacetime)\right] = 0$.
\end{lemma}

\textbf{Step 2:} We show that the stochastic process $M$ satisfies the conditions for an $\filt$-martingale, such that $\mathbb{E}[|M(\timesym)|] < \infty$ and $\mathbb{E}[M(\timesym) \mid \filt_s] = M(s)$.
\end{proof}

\begin{proof}[Proof of~\cref{prop:flux-decomp}]
    \textbf{Step 1:} Let $f(\spacetime)=1$. Since $\flux$ is measurable, nonnegative, and bounded on a finite domain, then $\int_\setspacetimesub|\flux(\spacetime)|\,d\spacetime=\int_\setspacetimesub\flux(\spacetime)\,d\spacetime<\infty$. Then we can write its noise term as
    \begin{equation}
        \int_{\setspacetimesub} dM(\spacetime) = M(\setspacetimesub)= \events(\setspacetimesub) 
        - \int_{\setspacetimesub} \flux(\spacetime)\,d\spacetime,\quad \mathbb{E}[M(\setspacetimesub)]=0.
        \label{eq:martin-zero}
    \end{equation}
    We can parametrize $M(\setspacetimesub)$ by time such that $M(t)=M(\setspace\times[0,t])$.

    \textbf{Step 2:} First, we show that $\mathbb{E}[|M(t)|]<\infty$. From~\cref{eq:martin-zero} and the triangle inequality,
    \begin{subequations}
        \begin{align}
            \mathbb{E}[|M(t)|] &= \mathbb{E}\left[\left|\events(\setspacetimesub) 
        - \int_\setspacetimesub\flux(\spacetime)\,d\spacetime\right|\right] \\
        &\leq \mathbb{E}[\underbrace{|\events(\setspacetimesub)|}_{\events(\setspacetimesub)\geq0}]
        + \underbrace{\mathbb{E}\left[\left|\int_\setspacetimesub\flux(\spacetime)\,d\spacetime\right|\right]}_{\flux(\spacetime)\geq0\text{ and deterministic}} \\
        &= \underbrace{\mathbb{E}[\events(\setspacetimesub)]}_{=\int_\setspacetimesub\flux(\spacetime)\,d\spacetime}
        + \int_\setspacetimesub\flux(\spacetime)\,d\spacetime \\
        &= 2\int_\setspacetimesub\flux(\spacetime)\,d\spacetime < \infty
        \end{align}
    \end{subequations}
    Next, we show $\mathbb{E}[M(t) \mid \filt_s] = M(s)$ for all $s \leq t$. Write
    \begin{subequations}
        \begin{align}
            \mathbb{E}[M(t) \mid \filt_s] 
            &= \mathbb{E}\left[\events(\setspace \times [0,t]) 
            - \underbrace{\int_{\setspace \times [0,t]} \flux(\spacetime)\,d\spacetime}_{\text{deterministic}} \;\middle|\; \filt_s\right] \\
            &= \mathbb{E}\left[\events(\setspace \times [0,t]) \mid \filt_s\right] 
            - \int_{\setspace \times [0,t]} \flux(\spacetime)\,d\spacetime,
        \end{align}
    \end{subequations}
    We decompose the count into the observed past and unobserved future,
    \begin{equation}
        \events(\setspace \times [0,t]) = \events(\setspace \times [0,s]) + \events(\setspace \times (s,t]),
    \end{equation}
    so that
    \begin{subequations}
        \begin{align}
            \mathbb{E}\left[\events(\setspace \times [0,t]) \mid \filt_s\right]
            &= \events(\setspace \times [0,s]) 
            + \mathbb{E}\left[\events(\setspace \times (s,t]) \mid \filt_s\right] \\
            &= \events(\setspace \times [0,s]) 
            + \int_{\setspace \times (s,t]} \flux(\spacetime)\,d\spacetime,
        \end{align}
    \end{subequations}
    where $\events(\setspace \times [0,s])$ is $\filt_s$-measurable and the increment $\events(\setspace \times (s,t])$ is independent of $\filt_s$ by the independent increments property of the Poisson process, with conditional expectation equal to its intensity integral. Substituting back,
    \begin{subequations}
        \begin{align}
            \mathbb{E}[M(t) \mid \filt_s] 
            &= \events(\setspace \times [0,s]) 
            + \int_{\setspace \times (s,t]} \flux(\spacetime)\,d\spacetime
            - \int_{\setspace \times [0,t]} \flux(\spacetime)\,d\spacetime \\
            &= \events(\setspace \times [0,s]) 
            - \int_{\setspace \times [0,s]} \flux(\spacetime)\,d\spacetime \\
            &= M(s).
        \end{align}
    \end{subequations}
    Since $\mathbb{E}[|M(t)|] < \infty$, $M(t)$ is $\filt_t$-measurable by construction, and $\mathbb{E}[M(t)\mid\filt_s]=M(s)$ for all $s\leq t$, we conclude that $M(t)$ is an $\filt$-martingale.
\end{proof}

\subsection{Spatiotemporal flux probing (proof of Equation 1)}
Using the martingale decomposition in~\cref{prop:flux-decomp}, we can prove the decomposition of the spatiotemporal flux probing measurements in Equation~1 in the main paper. We begin by restating a rigorous version of Equation~\ref{eq:flux_probing_integral}:

\textbf{Equation 1.} Define spatiotemporal Poisson point process $\events$ over the 
spatiotemporal volume $[0,\width]\times[0,\height] \times [0,\exposure]$ with 
underlying flux function $\flux(\spacex,\spacey,\timesym)$. Given a bounded, deterministic, and measurable probing 
function $\probe(\spacex,\spacey,\timesym)$, the following decomposition for probing 
measurements holds:
\begin{equation}
    \underbrace{\sum_{(\spacex_i, \spacey_i, \timesym_i) \in \events} \probe(\spacex_i, \spacey_i, \timesym_i)}_{\text{probing measurements}} = \underbrace{\int_{0}^{\exposure}\int_0^\height\int_0^\width
    \probespacetime \, \spacetimeflux \,d\spacex\,d\spacey\,d\timesym}_{\text{flux probing integral }\langle\probe, \flux\rangle} +  \underbrace{M_\probe(\exposure)} _{\text{martingale noise}}.
    \tag{1}\label{eq:flux_probing_integral}
\end{equation}
If $\events$ is a 1D Poisson point process, this becomes identical to the flux probing equation in~\cite{wei2023passive}:
\begin{equation}
    \sum_{\timesym_i \in \events} \probe(\timesym_i) = \int_{0}^{\exposure}
    \probe(\timesym) \, \flux(\timesym) \,dt +  M_\probe(\exposure).
    \label{eq:probe-meas-1d}
\end{equation}

We restate Equation~\ref{eq:flux_probing_integral} as a more succinct integral where $\setspacetime=[0,\width]\times[0,\height] \times [0,\exposure]$, which will be the notation used for the rest of the supplement:
\begin{equation}
    \sum_{\spacetime_i \in \events} \probe(\spacetime_i) 
    = \int_{\setspacetime} \probe(\spacetime)\,\flux(\spacetime)\,d\spacetime 
    + M_{\probe}(\exposure).
    \label{eq:flux-probe-integ-short}
\end{equation}

\begin{proof}[Proof Sketch of Equation~\ref{eq:flux_probing_integral}]
The proof proceeds in two steps:

\textbf{Step 1:} We substitute the function $f$ for our probing function $\probe$ in~\cref{lem:comp-integ}.

\textbf{Step 2:} We derive the flux probing measurements using the following 
lemma:

\begin{lemma}[Proposition \textbf{8.3.2.1.i} 
in~\cite{jeanblanc_mathematical_2009}]\label{lem:probe-martin}
    Let $\probe(\spacetime)$ be a bounded and deterministic function. Then the process
    \begin{equation}\label{eq:probe-martin}
        M_{\probe}(\timesym) 
        = \int_{\setspace \times [0,\timesym]} \probe(\spacetime)\,dM(\spacetime)
    \end{equation}
    is a $\filt$-martingale. The cited result establishes this for 1D processes; 
    the spatiotemporal case follows by fixing $\setspace$ and applying the result 
    to the marginal process $\timesym\mapsto M(\setspace\times[0,\timesym])$.
\end{lemma}
\end{proof}

\begin{proof}[Proof of Equation~\ref{eq:flux_probing_integral}]
\textbf{Step 1:} Let $f(\spacetime) 
= \probe(\spacetime)$ in~\cref{lem:comp-integ}. We obtain:
\begin{equation}
    \int_{\setspacetime} \probe(\spacetime)\,d\events(\spacetime) 
    = \int_{\setspacetime} \probe(\spacetime)\,dM(\spacetime) 
    + \int_{\setspacetime} \probe(\spacetime)\,\flux(\spacetime)\,d\spacetime.
\end{equation}

\textbf{Step 2:} 
The left-hand term equals $\sum_{\spacetime \in \events} \probe(\spacetime)$ 
by definition of integration against a counting measure. The martingale term
\begin{equation}
    \timesym \mapsto M_{\probe}(\timesym) 
    = \int_{\setspace \times [0,\timesym]} \probe(\spacetime)\,dM(\spacetime)
\end{equation}
is a $\filt$-martingale by~\cref{lem:probe-martin}. Setting $\timesym = \exposure$ and substituting $M_{\probe}(\exposure)$ gives us Equation~1.
\end{proof}

\subsection{Noise model for probing measurements in Equation 1}

From this section forward, we denote the probing measurements as $\probemeas_\probe \triangleq \sum_{\spacetime \in \events} \probe(\spacetime)$ and the probing flux integral as $\langle \probe,\flux\rangle=\int_{\setspacetime} \probe(\spacetime)\,\flux(\spacetime)\,d\spacetime$ for brevity.

\subsubsection{Mean and variance of probing measurements}

When describing the noise model for the flux probing measurements in the main paper, we claim the following:

\begin{proposition}
The probing measurements $\probemeas_\probe$ are unbiased with mean $\langle \probe,\flux\rangle$, variance $\langle \probe^2,\flux\rangle$, and approximately follow a normal distribution.
\label{prop:probe-dist}
\end{proposition}

\begin{proof}[Proof Sketch of~\cref{prop:probe-dist}]
\textbf{Step 1:} We show the mean of $\probemeas_\probe$ is $\langle \probe,\flux\rangle$, its variance is $\langle \probe^2,\flux\rangle$, and that it is unbiased via Campbell's theorem:
\begin{lemma}[Campbell's theorem (Section \textbf{3.2} in~\cite{kingman_poisson_1993})]\label{lemma:campbell}
For a Poisson point process $\events$ on $\setspacetime$ with flux intensity $\flux$ 
and intensity measure $d\fluxmeas(\spacetime) = \flux(\spacetime)\,d\spacetime$, 
and a measurable function $\probe : \setspacetime \to \mathbb{R}$:
\begin{enumerate}[label=(\roman*)]
    \item If $\int_\setspacetime |\probe(\spacetime)|\flux(\spacetime)\,d\spacetime < \infty$, then
    \begin{equation}    
        \mathbb{E}\left[\sum_{\spacetime \in \events} \probe(\spacetime)\right] 
        = \int_\setspacetime \probe(\spacetime)\flux(\spacetime)\,d\spacetime.
    \end{equation}
    \item If $\int_\setspacetime \probe(\spacetime)^2\flux(\spacetime)\,d\spacetime < \infty$, then
    \begin{equation}    
        \mathrm{Var}\left(\sum_{\spacetime \in \events} \probe(\spacetime)\right) 
        = \int_\setspacetime \probe(\spacetime)^2\flux(\spacetime)\,d\spacetime.
    \end{equation}
\end{enumerate}
Since we choose probing functions $\probe$ such that they are bounded and measurable, and evaluate them on a bounded domain, these integrals naturally converge.
\end{lemma}

\textbf{Step 2:} We show $\probemeas_\probe$ is approximately normal via the central limit theorem.
\end{proof}

\begin{proof}[Proof of ~\cref{prop:probe-dist}]
\textbf{Step 1:} The result follows directly from Campbell's theorem:
\begin{equation}
    \mathbb{E}[\probemeas_\probe] = \langle \probe,\flux\rangle,
\end{equation}
\begin{equation}
    \text{Var}(\probemeas_\probe) = \langle \probe^2,\flux\rangle.
\end{equation}
Naturally, this means our estimator $\probemeas_\probe$ is unbiased.

\textbf{Step 2:} The probing measurements are a sum of independent identically-distributed random variables, where each photon observation location is a random variable. This approximates a normal distribution under the central limit theorem~\cite{papoulis_probability_1991}. We verify this empirically via simulation in Section~\ref{sec:probe-normality}.
\end{proof}

\subsubsection{Covariance of probing measurements}

\begin{proposition}
    For probing functions $\probe_1,\probe_2$, their probing measurement covariance is:
    \begin{equation}
        \text{Cov}(\probemeas_{\probe_1},\probemeas_{\probe_2}) = \langle \probe_1\probe_2,\flux\rangle
        \label{eq:probe-cov}
    \end{equation}
    \label{prop:probe-cov}
\end{proposition}

\begin{proof}[Proof Sketch of~\cref{prop:probe-cov}]
We introduce the concept of second factorial moment measures for a more generalizable proof that is more succinct and does not require properties of martingales.

\begin{definition}[Second factorial moment measure (Definition \textbf{4.9} in~\cite{last_lectures_2017})]\label{def:2nd-facmm}
    For a general point process $\events$ and nonnegative measurable function $f$, its second factorial moment measure is defined as follows:
    \begin{equation}
        \mathbb{E}[\sum_{\spacetime,\spacetime'\in\events,\spacetime\neq \spacetime'}f(\spacetime,\spacetime')] = \int_{\setspacetime\times \setspacetime} f(\spacetime,\spacetime')\,M^{(2)}(d\spacetime,d\spacetime').
    \end{equation}
\end{definition}
Intuitively, the second factorial moment measure informs us, on average, how many distinct pairs of photons appear inside $\spacetime$. For general point processes, this can tell us whether points tend to cluster or spread out from each other. However, since photon arrivals are independent, knowing one photon arrival does not inform us where other photons arrive. This naturally leads to the following lemma:
\begin{lemma}[Corollary \textbf{4.10} in~\cite{last_lectures_2017}]
    For a Poisson point process with flux function $\flux$, its 2nd factorial moment measure simplifies to the following:
    \begin{equation}
        \int_{\setspacetime\times \setspacetime} f(\spacetime,\spacetime')\,M^{(2)}(d\spacetime,d\spacetime')=\int_{\setspacetime\times \setspacetime} f(\spacetime,\spacetime')\flux(\spacetime)\flux(\spacetime')\,d\spacetime d\spacetime'.
    \end{equation}
    \label{lem:2nd-facmm-pois}
\end{lemma}
\end{proof}

\begin{proof}[Proof of~\cref{prop:probe-cov}]
The covariance by definition is
\begin{equation}
    \text{Cov}(\probemeas_{\probe_1},\probemeas_{\probe_2})=\mathbb{E}[\probemeas_{\probe_1}\probemeas_{\probe_2}]-\mathbb{E}[\probemeas_{\probe_1}]\mathbb{E}[\probemeas_{\probe_2}].
    \label{eq:probe-cov-expand}
\end{equation}
We first evaluate $\mathbb{E}[\probemeas_{\probe_1}\probemeas_{\probe_2}]$:
\begin{equation}
    \mathbb{E}[\probemeas_{\probe_1}\probemeas_{\probe_2}] = \mathbb{E}[\sum_{\spacetime\in\events}\probe_1(\spacetime)\probe_2(\spacetime)] + \mathbb{E}[\sum_{\spacetime,\spacetime'\in\events,\spacetime\neq\spacetime'}\probe_1(\spacetime)\probe_2(\spacetime')]
    \label{eq:cov-cross-expect}
\end{equation}
Letting $f(\spacetime)=\probe_1(\spacetime)\probe_2(\spacetime)$, we have from Campbell's theorem (\cref{lemma:campbell}) that
\begin{equation}
    \mathbb{E}[\sum_{\spacetime\in\events}\probe_1(\spacetime)\probe_2(\spacetime)] = \int_\setspacetime\probe_1(\spacetime)\probe_2(\spacetime)\flux(\spacetime)\,d\spacetime = \langle \probe_1\probe_2,\flux\rangle.
\end{equation}
Then from Definition~\ref{def:2nd-facmm} and Lemma~\ref{lem:2nd-facmm-pois}, the second term can be written as a second factorial moment measure, and simplified as follows:
\begin{subequations}
\begin{align}
    \mathbb{E}[\sum_{\spacetime,\spacetime'\in\events,\spacetime\neq\spacetime'}\probe_1(\spacetime)\probe_2(\spacetime')] &= \int_{\setspacetime\times \setspacetime} \probe_1(\spacetime)\probe_2(\spacetime')\,M^{(2)}(d\spacetime,d\spacetime') \\
    &= \int_{\setspacetime\times \setspacetime} \probe_1(\spacetime)\probe_2(\spacetime')\flux(\spacetime)\flux(\spacetime')\,d\spacetime d\spacetime' \\
    &= \int_\setspacetime\probe_1(\spacetime)\flux(\spacetime)\underbrace{\left(\int_\setspacetime\probe_2(\spacetime')\flux(\spacetime')d\spacetime'\right)}_{\text{constant w.r.t. }\spacetime}\,d\spacetime \\
    &= \left(\int_\setspacetime\probe_1(\spacetime)\flux(\spacetime)\,d\spacetime \right)\left(\int_\setspacetime\probe_2(\spacetime')\flux(\spacetime')d\spacetime'\right) \\
    &= \mathbb{E}[\probemeas_{\probe_1}]\mathbb{E}[\probemeas_{\probe_2}].
\end{align}
\label{eq:probe-cov-derivation}
\end{subequations}
Substituting~\cref{eq:probe-cov-derivation} in~\cref{eq:cov-cross-expect} yields
\begin{equation}
    \mathbb{E}[\probemeas_{\probe_1}\probemeas_{\probe_2}] = \langle \probe_1\probe_2,\flux\rangle + \mathbb{E}[\probemeas_{\probe_1}]\mathbb{E}[\probemeas_{\probe_2}]
    \label{eq:cov-cross-expect-solved}
\end{equation}
Then substituting~\cref{eq:cov-cross-expect-solved} in~\cref{eq:probe-cov-expand} results in our desired result~\cref{eq:probe-cov}:
\begin{equation}
    \text{Cov}(\probemeas_{\probe_1},\probemeas_{\probe_2}) = \langle \probe_1\probe_2,\flux\rangle + \mathbb{E}[\probemeas_{\probe_1}]\mathbb{E}[\probemeas_{\probe_2}] - \mathbb{E}[\probemeas_{\probe_1}]\mathbb{E}[\probemeas_{\probe_2}] = \langle \probe_1\probe_2,\flux\rangle.
\end{equation}
\end{proof}

\subsection{Noise model for Fourier probing measurements (Equation 2)}

In this section, we derive the noise model for our probing measurements distribution described in Equations 2, 3, and 4.

\textbf{Equation 2.} Fix spatiotemporal frequency $\freqvec=[\freqx,\freqy,\freqt]$. The Fourier basis probing functions take the form $\probe_{\freqvec}(\spacetime)=e^{-j2\pi\freqvec^\top\spacetime}$, where the probing measurement $\probemeas_{\freqvec}$ is defined as
\begin{equation}
    \probemeas_{\freqvec} \triangleq \sum_{\spacetime\in\events}e^{-j2\pi\freqvec^\top\spacetime}.
    \tag{2}
\end{equation}

\subsubsection{Mean and covariance matrix of Fourier probing measurements (proof of Equation 3)}

\textbf{Equation 3.} For 3D Fourier basis functions, the probing 
measurements $\probemeas_\freqvec$ approximately follow a complex 
normal distribution with mean and covariance
\begin{equation}
    \probemean
    =
    \begin{bmatrix}
    \left\langle \cos( 2\pi\freqvec^{\top}\spacetime), \flux(\spacetime) \right\rangle\\
    \left\langle -\sin(2\pi\freqvec^{\top}\spacetime), \flux(\spacetime) \right\rangle
    \end{bmatrix}, \quad
    \probecov
    =
    \begin{bmatrix}
    \left\langle \cos^2(2\pi\freqvec^{\top}\spacetime), \flux(\spacetime) \right\rangle & 0\\
    0 & \left\langle \sin^2(2\pi\freqvec^{\top}\spacetime), \flux(\spacetime) \right\rangle
    \end{bmatrix},
    \tag{3}
    \label{eq:probe-dist-fourier}
\end{equation}
where the off-diagonal terms are negligible for natural videos by the 
power law decay of their power spectra.

\begin{corollary}\label{cor:cov-weaksig}
Fix frequency $\freqvec$. Suppose $2\freqvec$ contributes little signal to $\flux$ (\ie, $\langle p_{2\freqvec},\flux\rangle\approx 0$), or constitutes a high frequency in a natural video. Its covariance matrix can be approximated as
\begin{equation}
    \probecov
    \approx
    \frac{\langle 1,\flux\rangle}{2}I_2 = \underbrace{\frac{\mathbb{E}[|\events|]}{2}}_{\approx |\events|/2}I_2,
\end{equation}
where $I_2$ is the $2\times 2$ identity matrix. Equality holds if $2\freqvec$ contributes no signal to $\flux$ (\ie, $\langle \probe_{2\freqvec},\flux\rangle=\mathbb{E}[\probemeas_{2\freqvec}]=0$).
\end{corollary}

\begin{proof}[Proof Sketch of Equation~\ref{eq:probe-dist-fourier}]
\textbf{Step 1:} We separate the real and imaginary parts using Euler's formula, and derive their mean.

\textbf{Step 2:} We derive the covariance matrix between the real and imaginary parts of the Fourier probing measurements, and show that for natural videos, it is approximately diagonal, since their power spectra generally follow the power law~\cite{dong_statistics_1995}. This lets us conclude the real and imaginary components are approximately independent normal distributions.
\end{proof}

\begin{proof}[Proof of Equation~\ref{eq:probe-dist-fourier}]
\textbf{Step 1:} Euler's formula provides
\begin{equation}
    \probe_\freqvec(\spacetime)=\cos(2\pi\freqvec^\top\spacetime)-j\sin(2\pi\freqvec^\top\spacetime).
\end{equation}
Since $\text{Re}[p_\freqvec(\spacetime)]=\cos(2\pi\freqvec^\top\spacetime)$ and $\text{Im}[p_\freqvec(\spacetime)]=-\sin(2\pi\freqvec^\top\spacetime)$,~\cref{prop:probe-dist} gives us
\begin{subequations}
    \begin{align}
        \mathbb{E}[\text{Re}[\probemeas_{\freqvec}]]&=\langle \cos(2\pi\freqvec^\top\spacetime),\flux(\spacetime)\rangle, \\
        \mathbb{E}[\text{Im}[\probemeas_{\freqvec}]]&=\langle -\sin(2\pi\freqvec^\top\spacetime),\flux(\spacetime)\rangle.
    \end{align}
\end{subequations}

\textbf{Step 2:} Similarly, the variances are given by
\begin{subequations}
    \begin{align}
    \text{Var}(\text{Re}[\probemeas_{\freqvec}])&=\langle \cos^2(2\pi\freqvec^\top\spacetime),\flux(\spacetime)\rangle \\ &
    = \frac{1}{2}\underbrace{\langle 1,\flux(\spacetime)\rangle}_{\approx|\events|} + \frac{1}{2}\langle \cos(4\pi\freqvec^\top\spacetime),\flux(\spacetime)\rangle,
    \end{align}
\end{subequations}
\begin{subequations}
    \begin{align}
    \text{Var}(\text{Im}[\probemeas_{\freqvec}])&=\langle \sin^2(2\pi\freqvec^\top\spacetime),\flux(\spacetime)\rangle \\ &
    = \frac{1}{2}\langle 1,\flux(\spacetime)\rangle - \frac{1}{2}\langle \cos(4\pi\freqvec^\top\spacetime),\flux(\spacetime)\rangle.
    \end{align}
\end{subequations}

From~\cref{prop:probe-cov}, the covariance $\text{Cov}(\text{Re}[\probemeas_{\freqvec}],\text{Im}[\probemeas_{\freqvec}])$ is given by
\begin{subequations}
\begin{align}
    \text{Cov}(\text{Re}[\probemeas_{\freqvec}],\text{Im}[\probemeas_{\freqvec}]) &= -\langle \cos(2\pi\freqvec^\top\spacetime)\sin(2\pi\freqvec^\top\spacetime),\flux(\spacetime)\rangle \\
    &= -\frac{1}{2}\langle \sin(4\pi\freqvec^\top\spacetime),\flux(\spacetime)\rangle.
\end{align}
\label{eq:probe-cov-exact}
\end{subequations}

We note that~\cref{eq:probe-cov-exact} is the exact expression for the covariance. Since both $\langle \sin(4\pi\freqvec^\top\spacetime),\flux(\spacetime)\rangle$ and $\langle \cos(4\pi\freqvec^\top\spacetime),\flux(\spacetime)\rangle$ are bounded above by $A_{2\freqvec}/2$, where $A_{2\freqvec}$ is the amplitude of frequency $2\freqvec$ in $\flux$, if the frequency $2\freqvec$ is not present in the flux, then $A_{2\freqvec}=0$ (or equivalently, whenever
$\langle \sin(4\pi\freqvec^\top \spacetime),\flux(\spacetime)\rangle=0$ and
$\langle \cos(4\pi\freqvec^\top \spacetime),\flux(\spacetime)\rangle=0$), and the covariance becomes diagonal.

More generally, when the projection of $\flux$ onto frequency $2\freqvec$ is small, these
terms are small relative to $\langle 1,\flux\rangle$, and the covariance matrix is
well-approximated as diagonal. In practice, high-frequency components of
$\flux$ often decay, so the $2\freqvec$ projections become small as $\|\freqvec\|$ increases, since natural videos tend to follow the power law~\cite{dong_statistics_1995}.

More specifically, in the domain of high temporal frequencies $\freqt$ and low spatial frequencies $\freqvec_s=(\freqx,\freqy)$, the power spectrum $R(\freqvec_s,\freqt)$ follows the asymptotic relation~\cite{dong_statistics_1995}
\begin{equation}
    R(\freqvec_s,\freqt)\sim \frac{1}{\|\freqvec_s\|^{m-1}\freqt^2}
\end{equation}
In practice, high frequency components $\freqvec$ often decay quickly, thus their covariances quickly become small. We can approximate $\langle \sin(4\pi\freqvec^\top\spacetime),\flux(\spacetime)\rangle\approx 0$ and $\langle \cos(4\pi\freqvec^\top\spacetime),\flux(\spacetime)\rangle\approx 0$. This means the covariance terms of $\probemeas_\freqvec$ can be approximated as follows:
\begin{equation}
\text{Cov}(\text{Re}[\probemeas_{\freqvec}],\text{Im}[\probemeas_{\freqvec}]) \approx 0, \quad
\text{Var}(\text{Re}[\probemeas_\freqvec])\approx \text{Var}(\text{Im}[\probemeas_\freqvec])\approx \frac{\langle1,\flux\rangle}{2}.    
\end{equation}
Where equality holds if $2\freqvec$ contributes no energy to $\flux$.

From~\cref{prop:probe-dist}, this distribution is approximately normal.
\end{proof}

\subsubsection{Fourier probing energy (proof of Equation 4)}
\label{sec:fourier-probe-energy}

\textbf{Equation 4.} The total normalized Fourier probing energy is given by
\begin{equation}
    |\probemeasnorm_{\freqvec}|^2 \triangleq 
        \text{Re}\left(\frac{\probemeas_{\freqvec}}{\sqrt{\boldsymbol{\mathrm{\Sigma}}_{1,1}}}\right)^2+\text{Im}\left(\frac{\probemeas_{\freqvec}}{{\sqrt{\boldsymbol{\mathrm{\Sigma}}_{2,2}}}}\right)^2.
    \tag{4}
    \label{eq:fourier_energy}
\end{equation}
Under the diagonal covariance described in Equation~\ref{eq:probe-dist-fourier}, it follows a noncentral $\chi^2$ distribution with two degrees of freedom.

\begin{corollary}[Probing energy under the null hypothesis]
    Under the null hypothesis that both $\freqvec$ and $2\freqvec$ 
    contribute no energy to $\flux$ (\ie, $\langle 
    \probe_\freqvec,\flux\rangle=0$ and $\langle 
    \probe_{2\freqvec},\flux\rangle=0$), the covariance matrix reduces 
    to $\probecov = \frac{\langle 1,\flux \rangle}{2}I_2$, which means
    \begin{equation}
        |\widetilde{\probemeas_{\freqvec}}|^2 = \frac{2}{\langle 1,\flux\rangle}|\probemeas_{\freqvec}|^2.
    \end{equation}
    Thus, $|\probemeas_{\freqvec}|^2$ follows a scaled $\chi^2$ distribution with two degrees of freedom:
    \begin{equation}
    |\probemeas_{\freqvec}|^2 \sim \frac{\langle 1,\flux \rangle}{2}
    \chi^2_2
    \approx \frac{|\events|}{2}
    \chi^2_2
    \label{eq:probe-energy-null}
    \end{equation}
\end{corollary}

\begin{proof}[Proof of Equation~\ref{eq:fourier_energy}]
    $|\probemeasnorm_{\freqvec}|^2$ is the sum of two independent squared normal distributions with unit variance, which by definition follows a noncentral $\chi^2$ distribution with noncentrality parameter $\lambda$:
    \begin{equation}
    |\widetilde{\probemeas_{\freqvec}}|^2\sim\chi_2^2(\lambda),\quad
        \lambda=\frac{\boldsymbol{\mathrm{\mu}}_1^2}{\boldsymbol{\mathrm{\Sigma}}_{1,1}}+\frac{\boldsymbol{\mathrm{\mu}}_2^2}{\boldsymbol{\mathrm{\Sigma}}_{2,2}}.
    \end{equation}
    Under the null hypothesis, $\lambda = 0$ and the distribution is 
    central $\chi^2_2$.
\end{proof}

\subsection{CFAR frequency detection (proof of Equation 5)}
\label{sec:flux-fourier-cfar}

\textbf{Equation 5.} We use normalized probing 
function $\probe_{\freqvec}(\spacetime)=\frac{1}{\sqrt{\spacevol}}
e^{-j2\pi\freqvec^\top\spacetime}$, where $\spacevol = |\setspacetime| 
= \width\cdot\height\cdot\exposure$ is the spatiotemporal volume of 
$\setspacetime$, such that they are orthonormal over $\setspacetime$. 
Then frequency is selected with significance level $\cfarsigval$ if
\begin{equation}
    \tag{5}
    \label{eq:CFAR3d}
|\probemeas_{\freqvec}|^2 \geq \textsc{CDF}_{\chi_2^2}^{-1}(1-\cfarsigval) \frac{|\events|}{2\spacevol}.
\end{equation}

\begin{proof}[Proof of Equation~\ref{eq:CFAR3d}]
    Under the null hypothesis that $\freqvec$ and $2\freqvec$ both 
    contribute no energy to $\flux$, the energy 
    $|\probemeas_{\freqvec}|^2$ follows a scaled central $\chi_2^2$ 
    distribution from~\cref{eq:probe-energy-null}, where the scale 
    factor includes the spatiotemporal volume $\spacevol$ from the 
    normalized probing function:
    \begin{equation}
        |\probemeas_\freqvec|^2\sim \frac{\langle 1,\flux\rangle}{2\spacevol}
    \chi^2_2.
    \end{equation}
    When $\langle p_{2\freqvec},\flux\rangle\neq 0$, the covariance is 
    no longer isotropic and the test is approximate. However, since 
    natural video power spectra decay as a power law with frequency, 
    energy at $2\freqvec$ is suppressed relative to $\freqvec$, and 
    the isotropic approximation holds well in practice (see Section~\ref{sec:probe-normality}).

    The exact detection threshold $\textsc{CDF}_{\chi_2^2}^{-1}(1-\cfarsigval)\frac{\langle 
    1,\flux\rangle}{2\spacevol}$ 
    depends on the unknown flux $\flux$ and is therefore unobservable. 
    Approximating via the observed photon count $|\events|\approx
    \mathbb{E}[|\events|]=\langle 1,\flux\rangle$ gives Equation~\ref{eq:CFAR3d}. We show in Section~\ref{sec:cfar-robust} that this approximation performs well in practice.
\end{proof}

\clearpage

\section{Velocity Detection}
\label{sec:vel-det}
In this section, we provide the derivations and proofs for the velocity detection framework presented in Section 3.2 of the main paper.

\subsection{Fourier planar structures induced by linear velocity (proof of Equation 6)}
For convenience, we restate Equation 6 from the main paper.

\textbf{Equation 6.} Let a local patch of spatial intensity $I$ translate at a constant velocity $\mathbf{v}=[v_x,v_y]^\top$ such that
\begin{equation*}
\flux(x,y,t)=I(x-v_xt,y-v_yt).
\label{eq:eq6_restate_signal}
\end{equation*}
Then the Fourier coefficients lie on the plane
\begin{equation}
v_x f_x + v_y f_y + f_t = 0
\tag{6}
\label{eq:motion_plane}
\end{equation}

\begin{proof}[Proof Sketch of Equation 6.]
    The proof proceeds in two steps.
    
    \textbf{Step 1.} We write the Fourier transform of the translated pattern as a space-time integral and perform a change of variables that aligns the coordinates with the motion.
    
    \textbf{Step 2.} We evaluate the transformed integral and show that it vanishes unless the frequencies satisfy Equation~6
    
\end{proof}

\begin{proof}[Proof of Equation 6.]
\mbox{}\\

\textbf{Step 1.} Let
\begin{equation}
\flux(x,y,t)=I(x-v_xt,y-v_yt),
\label{eq:motion_model_proof}
\end{equation}
for $(x,y,t)\in\mathbb{R}^3$.

The Fourier transform of $\flux(x,y,t)$ is
\begin{equation}
\fftflux(f_x,f_y,f_t)
=
\iiint_{(x,y,t)\in\mathbb{R}^3}
\flux(x,y,t)\,
e^{-i2\pi(f_xx+f_yy+f_tt)}
\,dx\,dy\,dt.
\label{eq:fourier_motion_def}
\end{equation}

Substituting \eqref{eq:motion_model_proof} into \eqref{eq:fourier_motion_def} gives
\begin{equation}
\fftflux(f_x,f_y,f_t)
=
\iiint_{(x,y,t)\in\mathbb{R}^3}
I(x-v_xt,y-v_yt)\,
e^{-i2\pi(f_xx+f_yy+f_tt)}
\,dx\,dy\,dt.
\label{eq:fourier_motion_start}
\end{equation}

We now introduce the change of variables
\begin{equation}
\xi = x-v_xt,\qquad \eta = y-v_yt,\qquad \tau = t,
\label{eq:cov_forward}
\end{equation}
for $(\xi,\eta,\tau)\in\mathbb{R}^3$. Equivalently,
\begin{equation}
x=\xi+v_x\tau,\qquad y=\eta+v_y\tau,\qquad t=\tau.
\label{eq:cov_inverse}
\end{equation}

The Jacobian matrix of this transformation is
\begin{equation}
\frac{\partial(x,y,t)}{\partial(\xi,\eta,\tau)}
=
\begin{pmatrix}
1 & 0 & v_x\\
0 & 1 & v_y\\
0 & 0 & 1
\end{pmatrix},
\label{eq:jacobian_matrix}
\end{equation}
whose determinant is
\begin{equation}
\left|\frac{\partial(x,y,t)}{\partial(\xi,\eta,\tau)}\right|=1.
\label{eq:jacobian_det}
\end{equation}
Therefore,
\begin{equation}
dx\,dy\,dt = d\xi\,d\eta\,d\tau.
\label{eq:jacobian_measure}
\end{equation}

Applying the change of variables \eqref{eq:cov_forward}--\eqref{eq:jacobian_measure} to \eqref{eq:fourier_motion_start}, we obtain
\begin{equation}
\fftflux(f_x,f_y,f_t)
=
\iiint_{(\xi,\eta,\tau)\in\mathbb{R}^3}
I(\xi,\eta)\,
e^{-i2\pi\left[f_x(\xi+v_x\tau)+f_y(\eta+v_y\tau)+f_t\tau\right]}
\,d\xi\,d\eta\,d\tau.
\label{eq:fourier_motion_cov_a}
\end{equation}
Rearranging the exponent gives
\begin{equation}
\fftflux(f_x,f_y,f_t)
=
\iiint_{(\xi,\eta,\tau)\in\mathbb{R}^3}
I(\xi,\eta)\,
e^{-i2\pi(f_x\xi+f_y\eta)}
e^{-i2\pi(v_xf_x+v_yf_y+f_t)\tau}
\,d\xi\,d\eta\,d\tau.
\label{eq:fourier_motion_cov_b}
\end{equation}

\textbf{Step 2.} The integrand in \eqref{eq:fourier_motion_cov_b} separates into a spatial term depending only on $(\xi,\eta)$ and a temporal term depending only on $\tau$. Hence,
\begin{equation}
\fftflux(f_x,f_y,f_t)
=
\left(
\iint_{(\xi,\eta)\in\mathbb{R}^2}
I(\xi,\eta)\,e^{-i2\pi(f_x\xi+f_y\eta)}
\,d\xi\,d\eta
\right)
\left(
\int_{\tau\in\mathbb{R}}
e^{-i2\pi(v_xf_x+v_yf_y+f_t)\tau}
\,d\tau
\right).
\label{eq:fourier_motion_factor}
\end{equation}

Define
\begin{equation}
\widehat{I}(f_x,f_y)
=
\iint_{(\xi,\eta)\in\mathbb{R}^2}
I(\xi,\eta)\,e^{-i2\pi(f_x\xi+f_y\eta)}
\,d\xi\,d\eta
\label{eq:spatial_ft_def}
\end{equation}
and
\begin{equation}
J(f_x,f_y,f_t)
=
\int_{\tau\in\mathbb{R}}
e^{-i2\pi(v_xf_x+v_yf_y+f_t)\tau}
\,d\tau.
\label{eq:J_def}
\end{equation}
Then \eqref{eq:fourier_motion_factor} becomes
\begin{equation}
\fftflux(f_x,f_y,f_t)=\widehat{I}(f_x,f_y)\,J(f_x,f_y,f_t).
\label{eq:fourier_motion_factor_compact}
\end{equation}

Using the distributional identity
\begin{equation}
\int_{\tau\in\mathbb{R}} e^{-i2\pi \alpha \tau}\,d\tau = \delta(\alpha),
\label{eq:delta_identity}
\end{equation}
with
\begin{equation}
\alpha = v_xf_x + v_yf_y + f_t,
\label{eq:alpha_def}
\end{equation}
we obtain
\begin{equation}
J(f_x,f_y,f_t)=\delta(v_xf_x+v_yf_y+f_t).
\label{eq:J_delta}
\end{equation}

Substituting \eqref{eq:J_delta} into \eqref{eq:fourier_motion_factor_compact} yields
\begin{equation}
\fftflux(f_x,f_y,f_t)
=
\widehat{I}(f_x,f_y)\,\delta(v_xf_x+v_yf_y+f_t).
\label{eq:fourier_motion_final_delta}
\end{equation}
Therefore, $\fftflux(f_x,f_y,f_t)$ can be nonzero only when
\begin{equation}
v_xf_x + v_yf_y + f_t = 0.
\label{eq:fourier_plane_final}
\end{equation}
That is, the 3D Fourier coefficients of $\flux$ are supported on a plane through the origin with normal vector $[v_x,v_y,1]^\top$.
This concludes the proof.
\end{proof}

\begin{remark}[Finite observation window]
The identity in \eqref{eq:delta_identity} assumes integration over $\tau\in\mathbb{R}$. For a finite observation interval $\tau\in[0,\exposure]$, the delta distribution is replaced by a sinc-like kernel of width on the order of $1/\exposure$, and the Fourier energy concentrates near the plane \eqref{eq:fourier_plane_final} rather than exactly on it. Thus, the plane constraint is exact in the limit $\exposure\to\infty$ and approximate for finite $\exposure$.
\end{remark}

\subsection{Parametric velocity detection model}

We next describe a parametric CFAR model for the local velocity energies. Let $\Pi_i$ denote the set of Fourier coefficients lying on the candidate plane associated with velocity hypothesis $\velocityVec_i$, and let
\begin{equation}
E_i \triangleq \sum_{\freqvec\in\Pi_i} |\probemeas_\freqvec|^2
\label{eq:plane_energy}
\end{equation}
be the corresponding plane energy. Under the null hypothesis $H_0$, assume that the estimate of the Fourier coefficients on the plane are independent circular complex normal random variables with zero mean,
\begin{equation}
\probemeas_\freqvec\mid H_0 \sim \mathcal{CN}(0,\sigma_i^2),
\qquad \freqvec\in\Pi_i.
\label{eq:null_fourier_coeff}
\end{equation}
Writing each estimated coefficient as
\begin{equation}
\probemeas_\freqvec=X_{\freqvec}+ \mathbf{i}Y_{\freqvec},
\label{eq:coeff_decomposition}
\end{equation}
with
\begin{equation}
X_{\freqvec},Y_{\freqvec}\sim\normaldist(0,\sigma_i^2/2),
\label{eq:real_imag_distribution}
\end{equation}
we have
\begin{equation}
|\probemeas_\freqvec|^2 = X_{\freqvec}^2+Y_{\freqvec}^2,
\label{eq:magnitude_squared}
\end{equation}
and therefore
\begin{equation}
\frac{2|\probemeas_\freqvec|^2}{\sigma_i^2}\sim \chi_2^2.
\label{eq:single_coeff_chi2}
\end{equation}
If the plane contains $|\Pi_i|$ estimated coefficients, then summing over all the ones on the plane gives
\begin{equation}
\frac{2E_i}{\sigma_i^2}
=
\sum_{\freqvec\in\Pi_i}
\frac{2|\probemeas_\freqvec|^2}{\sigma_i^2}
\sim \chi_{2|\Pi_i|}^2.
\label{eq:plane_energy_chi2}
\end{equation}
Hence, under $H_0$, the plane energy $E_i$ follows a scaled central chi-squared distribution with $2|\Pi_i|$ degrees of freedom.

Under the alternative hypothesis $H_1$, the plane contains true signal energy, so the estimated Fourier coefficients have nonzero means:
\begin{equation}
\probemeas_\freqvec\sim \mathcal{CN}(\mu_{\freqvec},\sigma_i^2).
\label{eq:alt_fourier_coeff}
\end{equation}
In this case,
\begin{equation}
\frac{2E_i}{\sigma_i^2}\sim \chi_{2|\Pi_i|}^2(\lambda_i),
\qquad
\lambda_i=\sum_{\freqvec\in\Pi_i}\frac{2|\mu_{\freqvec}|^2}{\sigma_i^2},
\label{eq:plane_energy_noncentral}
\end{equation}
so that $E_i$ follows a scaled noncentral chi-squared distribution~\cite{kay1993statistical}.

\begin{remark}[Gamma and Gaussian forms under $H_0$]
Since
\begin{equation}
\frac{1}{2}\chi^2_{2k}\sim\mathrm{Gamma}(k,1),
\label{eq:gamma_identity}
\end{equation}
the null distribution can equivalently be written as
\begin{equation}
E_i \sim \mathrm{Gamma}(|\Pi_i|,\,\sigma_i^2),
\label{eq:null_gamma}
\end{equation}
with
\begin{equation}
\mathbb{E}[E_i\mid H_0]=|\Pi_i|\sigma_i^2,
\qquad
\variance{E_i\mid H_0}=|\Pi_i|\sigma_i^4.
\label{eq:null_moments}
\end{equation}
As $|\Pi_i|$ grows, the classical CLT for i.i.d. summands gives the normal approximation
\begin{equation}
E_i \mid H_0 \approx \normaldist\!\left(|\Pi_i|\sigma_i^2,\;|\Pi_i|\sigma_i^4\right).
\label{eq:null_gaussian}
\end{equation}
Therefore, when the number of coefficients accumulated on the plane is large, the null distribution of the plane energy may be approximated by a normal distribution with the mean and variance above.
\end{remark}

\subsection{Limitation of the parametric velocity detection model}

We next discuss the assumptions underlying a parametric velocity-plane detector and motivate the use of a nonparametric alternative. Implementing a parametric CFAR detector would require estimating at least the local scale parameter $\sigma_i^2$, and in the nonzero-mean case also the noncentrality parameter $\lambda_i$, from neighboring training cells.

A natural approach under this model is to use a window-based CFAR detector, such as CA-CFAR or GO-CFAR, in which a local neighborhood is defined around each candidate velocity $\velocityVec_i$, a guard region is excluded around the cell under test, and the remaining cells are used to estimate the local background level and set a detection threshold. This yields a parametric test whose false alarm probability is controlled only insofar as the assumed distributional model is accurate.

However, these assumptions are often too strong in our setting. In particular, the Fourier coefficients accumulated along nearby velocity planes are generally not independent, since nearby planes overlap substantially in Fourier space and are computed from the same underlying data. As a result, nearby plane energies are also correlated, and the assumption that the local plane energies are independent and identically distributed can be difficult to justify. For this reason, a fully parametric CFAR model is brittle when applied to the velocity detection problem.

This motivates a weaker nonparametric null model. Rather than assuming that the local plane energies are independent and identically distributed, we assume only that under $H_0$ the collection
$$
E_i,\{E_j\}_{j\in\neighborhood_i}
$$
is exchangeable and has a continuous joint distribution~\cite{meng2020rank}. Exchangeability means that the joint distribution is invariant under any permutation of the variables, so that under the null hypothesis no candidate in the local neighborhood is statistically privileged over any other. Continuity ensures that ties occur with probability zero. These assumptions are sufficient to derive the uniform rank law used by the rank-based CFAR detector, while avoiding the stronger and less realistic requirement of independence among neighboring plane energies.

\subsection{Nonparametric velocity detection model (proof of Equation 8)}
\label{sec:velocity_detection}
In this section, we prove Equation 8 of the main paper, after providing notation.

Let $\velocityVec_i=[\velocity_x,\velocity_y]^\top$ denote a candidate velocity hypothesis in a discrete velocity grid, and let $E_i$ be its associated energy. For each hypothesis $\velocityVec_i$, we define a local neighborhood $\neighborhood_i$ by centering a square window around $\velocityVec_i$ and excluding a smaller guard region around the cell under test. The guard region prevents the energy of the tested hypothesis and its immediate neighbors from contaminating the background set, while the remaining cells in the window serve as neighboring hypotheses for comparison. We denote the number of such neighboring hypotheses by $|\neighborhood_i|$.

\textbf{Equation 8.} Let $E_i$ denote the energy of a candidate velocity hypothesis $\velocityVec_i$, and let $\neighborhood_i$ be its set of neighboring hypotheses used as training cells. Define the rank
\begin{equation*}
r_i \triangleq \left|\left\{j\in\neighborhood_i : E_j < E_i \right\}\right|.
\label{eq:rank_def}
\end{equation*}
Then detecting $E_i$ by requiring it to lie in the top $\alpha_{\mathrm{vel}}$ fraction of its local neighborhood is equivalent to the test
\begin{equation}
\tag{8}
r_i \ge \left\lceil (1-\alpha_{\mathrm{vel}})(|\neighborhood_i|+1)\right\rceil.
\label{eq:rank_cfar_threshold}
\end{equation}
Moreover, under the null hypothesis $H_0$ that the collection
\begin{equation*}
E_i,\{E_j\}_{j\in\neighborhood_i}
\label{eq:exchangeable_collection}
\end{equation*}
is exchangeable and has a continuous joint distribution, the rank $r_i$ is uniformly distributed on $\{0,1,\dots,|\neighborhood_i|\}$. Consequently, the false alarm probability of the detector depends only on the selected threshold and the neighborhood size, thereby yielding a CFAR test.

\begin{proof}[Proof sketch of Equation 8.]
The proof proceeds in four steps:

\textbf{Step 1:} We relate the rank $r_i$ to the position of $E_i$ in the sorted list formed by the cell under test together with its neighboring hypotheses.

\textbf{Step 2:} We show that requiring $E_i$ to belong to the top $\alpha_{\mathrm{vel}}$ fraction of that list is equivalent to the threshold in Equation~8.

\textbf{Step 3:} Under $H_0$, we show that exchangeability implies that all positions of the cell under test in the sorted list are equally likely, while continuity guarantees that ties occur with probability zero.

\textbf{Step 4:} We compute the probability of detection under $H_0$ and show that it depends only on the threshold and the neighborhood size.
\end{proof}

\begin{proof}[Proof of Equation 8.]
\mbox{}\\

\textbf{Step 1:} Consider the set consisting of the cell under test and its neighboring hypotheses,
\begin{equation}
\{E_i\}\cup\{E_j : j\in\neighborhood_i\}.
\label{eq:cut_and_neighbors}
\end{equation}
Its total size is
\begin{equation}
|\neighborhood_i|+1.
\label{eq:total_set_size}
\end{equation}
Let these values be sorted in ascending order. Since
\begin{equation}
r_i = \left|\left\{j\in\neighborhood_i : E_j < E_i \right\}\right|,
\label{eq:rank_def_repeat}
\end{equation}
the quantity $r_i$ counts exactly how many neighboring energies are strictly smaller than $E_i$. Therefore, the position of $E_i$ in the ascending ordering is
\begin{equation}
r_i+1.
\label{eq:rank_position}
\end{equation}

\textbf{Step 2:} We require that $E_i$ lies in the top $\alpha_{\mathrm{vel}}$ fraction of the $|\neighborhood_i|+1$ values. This means that the ascending position of $E_i$ must satisfy
\begin{equation}
r_i + 1 > (1-\alpha_{\mathrm{vel}})(|\neighborhood_i|+1),
\label{eq:top_fraction_condition}
\end{equation}
\ie,
\begin{equation}
r_i > (1-\alpha_{\mathrm{vel}})(|\neighborhood_i|+1) - 1.
\label{eq:rank_inequality_preceil}
\end{equation}
Since $r_i$ is integer-valued, this is equivalent to
\begin{equation}
r_i \ge \left\lceil (1-\alpha_{\mathrm{vel}})(|\neighborhood_i|+1)\right\rceil.
\label{eq:rank_inequality_ceil}
\end{equation}
This is exactly \eqref{eq:rank_cfar_threshold}. Consequently, detecting a hypothesis only when its energy lies in the top $\alpha_{\mathrm{vel}}$ fraction of the combined set is equivalent to thresholding its rank by \eqref{eq:rank_cfar_threshold}.

\textbf{Step 3:} Under the null hypothesis $H_0$, the collection
\begin{equation}
E_i,\{E_j\}_{j\in\neighborhood_i}
\label{eq:exchangeable_collection_repeat}
\end{equation}
is exchangeable and has a continuous joint distribution. Since the joint distribution is continuous, ties occur with probability zero. Therefore, with probability one, these values can be arranged in a strict ascending order.

Because the collection is exchangeable, its joint distribution is invariant under permutation of the samples. Hence, no sample is favored over any other in the ordering, and $E_i$ is equally likely to occupy any of the $|\neighborhood_i|+1$ positions in the ascending ordering. For any fixed $k\in\{0,1,\dots,|\neighborhood_i|\}$, the event $r_i=k$ means that exactly $k$ neighboring energies are smaller than $E_i$, or equivalently, that $E_i$ occupies position $k+1$ in the ascending ordering. Since each position is equally likely, we have
\begin{equation}
\probability{r_i = k \mid H_0}
=
\frac{1}{|\neighborhood_i|+1},
\qquad k=0,1,\dots,|\neighborhood_i|.
\label{eq:uniform_rank_null}
\end{equation}
Therefore, under $H_0$, the rank $r_i$ is uniformly distributed on $\{0,1,\dots,|\neighborhood_i|\}$ ~\cite{lehmann2005testing}.

\textbf{Step 4:} Let
\begin{equation}
T_i=\left\lceil (1-\alpha_{\mathrm{vel}})(|\neighborhood_i|+1)\right\rceil.
\label{eq:threshold_def}
\end{equation}
Under $H_0$, the detector declares a detection whenever $r_i\ge T_i$. Since $r_i$ is uniformly distributed on $\{0,1,\dots,|\neighborhood_i|\}$, we obtain
\begin{equation}
\probability{\text{detect}\mid H_0}
=
\probability{r_i\ge T_i\mid H_0}
=
\sum_{k=T_i}^{|\neighborhood_i|}\probability{r_i=k\mid H_0}
=
\sum_{k=T_i}^{|\neighborhood_i|}\frac{1}{|\neighborhood_i|+1}.
\label{eq:fa_sum}
\end{equation}
Evaluating the sum gives
\begin{equation}
\probability{\text{detect}\mid H_0}
=
\frac{|\neighborhood_i|-T_i+1}{|\neighborhood_i|+1}.
\label{eq:fa_with_threshold}
\end{equation}
Substituting the definition of $T_i$, we obtain
\begin{equation}
\probability{\text{detect}\mid H_0}
=
\frac{|\neighborhood_i|-\left\lceil (1-\alpha_{\mathrm{vel}})(|\neighborhood_i|+1)\right\rceil+1}{|\neighborhood_i|+1}.
\label{eq:fa_closed_form}
\end{equation}
Therefore, under the null hypothesis, the false alarm probability depends only on the selected threshold and the neighborhood size, and not on the underlying background distribution. This establishes the CFAR property and concludes the proof.
\end{proof}

\begin{remark}[Exact false alarm level on integer-compatible grids]
In general, the false alarm probability is
\begin{equation}
\probability{\text{detect}\mid H_0}
=
\frac{|\neighborhood_i|-\left\lceil (1-\alpha_{\mathrm{vel}})(|\neighborhood_i|+1)\right\rceil+1}{|\neighborhood_i|+1}.
\label{eq:remark_fa_general}
\end{equation}
In particular, this expression reduces exactly to
\begin{equation}
\probability{\text{detect}\mid H_0}=\alpha_{\mathrm{vel}}
\label{eq:remark_fa_exact}
\end{equation}
whenever $\alpha_{\mathrm{vel}}(|\neighborhood_i|+1)$ is an integer. Indeed, in that case,
\begin{equation}
\left\lceil (1-\alpha_{\mathrm{vel}})(|\neighborhood_i|+1)\right\rceil
=
(|\neighborhood_i|+1)-\alpha_{\mathrm{vel}}(|\neighborhood_i|+1),
\label{eq:remark_ceiling_identity}
\end{equation}
and substituting into \eqref{eq:remark_fa_general} gives
\begin{equation}
\probability{\text{detect}\mid H_0}
=
\frac{|\neighborhood_i|-\left((|\neighborhood_i|+1)-\alpha_{\mathrm{vel}}(|\neighborhood_i|+1)\right)+1}{|\neighborhood_i|+1}
=
\alpha_{\mathrm{vel}}.
\label{eq:remark_exact_substitution}
\end{equation}
Therefore, when $\alpha_{\mathrm{vel}}(|\neighborhood_i|+1)$ is integer-valued, the rank threshold yields exactly the desired false alarm probability.
\end{remark}

\clearpage
\section{Extension to Other Asynchronous Sensors}

\subsection{Spike Cameras: Spatiotemporal Probing Derivation}
In this section, we derive the extension of spatiotemporal flux probing to spike cameras. We start with some notation and derive the extension to 1D temporal probing.

\paragraph{Integrate-and-fire mechanism} The spike camera pixel integrates the photon arrivals into an internal accumulator. A spike is emitted each time the accumulator reaches a threshold $\Theta > 0$ (in units of photon counts), and the accumulator resets to zero~\cite{huang_1000_2023}. Let $\{\tau_k(\spacexy)\}_{k=1}^{N^{\mathrm{sp}}(\spacexy)}$ be the spike times, with $\tau_0(\spacexy) \triangleq 0$. Define the spike counting process
\begin{equation}
    N_{\spacexy}^{\mathrm{sp}}(t) \triangleq \big|\{k : \tau_k(\spacexy) \leq t\}\big|.
\end{equation}

By construction, the $k$-th spike fires when exactly $\Theta$ photons have arrived since the $(k{-}1)$-th spike:
\begin{equation}
    N_{\spacexy}^{\mathrm{ph}}\big(\tau_k(\spacexy)\big) - N_{\spacexy}^{\mathrm{ph}}\big(\tau_{k-1}(\spacexy)\big) = \Theta, \qquad k = 1,2,\dots
    \label{eq:threshold_photon}
\end{equation}
and the residual photon count since the last spike satisfies
\begin{equation}
    \varepsilon_{\spacexy}(t) \triangleq N_{\spacexy}^{\mathrm{ph}}(t) - N_{\spacexy}^{\mathrm{ph}}\big(\tau_{N_{\spacexy}^{\mathrm{sp}}(t)}(\spacexy)\big) < \Theta.
    \label{eq:residual_count}
\end{equation}
This gives the fundamental identity
\begin{equation}
    N_{\spacexy}^{\mathrm{ph}}(t) = \Theta \, N_{\spacexy}^{\mathrm{sp}}(t) + \varepsilon_{\spacexy}(t).
    \label{eq:photon_spike_identity}
\end{equation}

\subsubsection{Temporal spike probing}

\begin{proposition}[Spike probing decomposition]
\label{prop:unified_1d}
Let $p$ be a deterministic, bounded, and differentiable probing function. Then
\begin{equation}
    \Theta \sum_{k=1}^{N^{\mathrm{sp}}(\spacexy)} p\big(\tau_k(\spacexy)\big)
    \;=\;
    \int_0^{\exposure} p(t)\,\phi(\spacexy,t)\,d t
    \;+\;
    M_{p,\spacexy}
    \;+\;
    \eta_{p,\spacexy},
    \label{eq:unified_probing}
\end{equation}
where:
\begin{itemize}
    \item $M_{p,\spacexy}$ is martingale noise inherited from~\cref{eq:probe-meas-1d}, the flux probing measurements,
    \item $\eta_{p,\spacexy}$ is a bounded quantization residual satisfying
    \begin{equation}
        |\eta_{p,\spacexy}| \leq \Theta\big(|p(\exposure)| + |p(0)|\big) + \Theta \int_0^{\exposure} |p'(t)|\,d t.
        \label{eq:eta_bound}
    \end{equation}
\end{itemize}
\end{proposition}

\begin{proof}[Proof Sketch of~\cref{prop:unified_1d}]
\textbf{Step 1:} We substitute the 1D martingale decomposition from the photon counting process described in~\cref{eq:flux-decomp-1d}, and show it can be represented in differential form.

\textbf{Step 2:} We integrate the differential form of the decomposition from step 1, and evaluate the individual terms.

\textbf{Step 3:} We derive the bounds of the quantization error.
\end{proof}

\begin{proof}
\textbf{Step 1:}
The martingale decomposition described in~\cref{eq:flux-decomp-1d} can be substituted directly into the photon--spike identity~\eqref{eq:photon_spike_identity}:
\begin{equation}
    \int_0^t \phi(\spacexy,s)\,\mathrm{d}s + M_{\spacexy}(t) = \Theta\, N_{\spacexy}^{\mathrm{sp}}(t) + \varepsilon_{\spacexy}(t).
    \label{eq:combined_identity}
\end{equation}
Both sides of~\eqref{eq:combined_identity} are functions of $t$ that change only by jumps (at photon arrivals on the left, at spike emissions and photon arrivals on the right) or by smooth accumulation (the flux integral on the left). In particular, every term is a well-defined increasing or piecewise-constant process whose value at any time is determined by finitely many photon and spike events up to that time. This regularity is sufficient to take increments on both sides~\cite{cohen_stochastic_2015}, giving the differential form
\begin{equation}
    \phi(\spacexy,t)\,\mathrm{d}t + \mathrm{d}M_{\spacexy}(t) = \Theta\,\mathrm{d}N_{\spacexy}^{\mathrm{sp}}(t) + \mathrm{d}\varepsilon_{\spacexy}(t).
    \label{eq:stieltjes_diff}
\end{equation}
\textbf{Step 2:}
Multiply both sides of~\eqref{eq:stieltjes_diff} by $p(t)$ and integrate over $[0,\exposure]$:
\begin{equation}
    \int_0^{\exposure} p(t)\,\phi(\spacexy,t)\,d t + \int_0^{\exposure} p(t)\,d M_{\spacexy}(t)
    =
    \Theta \int_0^{\exposure} p(t)\,d N_{\spacexy}^{\mathrm{sp}}(t) + \int_0^{\exposure} p(t)\,d \varepsilon_{\spacexy}(t).
    \label{eq:probed_both_sides}
\end{equation}
Since $\,d N_{\spacexy}^{\mathrm{sp}}(t) = \sum_k \delta(t - \tau_k(\spacexy))\,d t$, the integral $\int_0^{\exposure} p(t)\,d N_{\spacexy}^{\mathrm{sp}}(t)$ evaluates $\probe$ along a step process, which evaluates to the sum
\begin{equation}
    \int_0^{\exposure} p(t)\,d N_{\spacexy}^{\mathrm{sp}}(t) = \sum_{k=1}^{N^{\mathrm{sp}}(\spacexy)} p\big(\tau_k(\spacexy)\big).
\end{equation}
Rearranging~\eqref{eq:probed_both_sides} gives
\begin{equation}
    \Theta \sum_{k=1}^{N^{\mathrm{sp}}(\spacexy)} p\big(\tau_k(\spacexy)\big)
    =
    \int_0^{\exposure} p(t)\,\phi(\spacexy,t)\,d t
    +
    \underbrace{\int_0^{\exposure} p(t)\,d M_{\spacexy}(t)}_{\displaystyle M_{p,\spacexy}}
    +
    \underbrace{\left(-\int_0^{\exposure} p(t)\,d \varepsilon_{\spacexy}(t)\right)}_{\displaystyle \eta_{p,\spacexy}},
\end{equation}
such that $M_{p,\spacexy}$ is a martingale by~\cref{lem:probe-martin}.

\textbf{Step 3:}
Integration by parts gives
\begin{align}
    \eta_{p,\spacexy} &= -\int_0^{\exposure} p(t)\,d \varepsilon_{\spacexy}(t) \notag\\
    &= -\Big[p({\exposure})\varepsilon_{\spacexy}({\exposure}) - p(0)\varepsilon_{\spacexy}(0)\Big] + \int_0^{\exposure} \varepsilon_{\spacexy}(t)\,p'(t)\,d t.
\end{align}
Taking absolute values and using $|\varepsilon_{\spacexy}(t)| < \Theta$:
\begin{equation}
    |\eta_{p,\spacexy}| \leq \Theta\big(|p(\exposure)| + |p(0)|\big) + \Theta\int_0^{\exposure} |p'(t)|\,d t. \qedhere
\end{equation}
\end{proof}

\begin{remark}[Limiting regimes]
\label{rem:limits}
The decomposition~\eqref{eq:unified_probing} unifies the photon and spike probing frameworks:
\begin{enumerate}
    \item \textbf{Single-photon limit} ($\Theta = 1$): Every photon triggers a spike, so $N^{\mathrm{sp}} = N^{\mathrm{ph}}$, the residual $\varepsilon_{\spacexy}(t) \equiv 0$ (since it can only be 0), and $\eta_{p,\spacexy} = 0$. The equation reduces to the photon flux probing equation:
    \begin{equation}
        \sum_{k} p(\tau_k) = \int_0^{\exposure} p(t)\flux(\spacexy,t)\,d t + M_{p,\spacexy}.
    \end{equation}
    \item \textbf{Deterministic (high-flux) limit}: When the mean photon count per threshold window $\Theta/\bar\flux$ is large, each inter-spike interval concentrates around its deterministic value by the law of large numbers. The spike times become approximately deterministic functionals of $\flux$, and $M_{p,\spacexy}/\Theta \to 0$ in a suitable sense. The equation reduces to the following spike probing equation:
    \begin{equation}
        \Theta\sum_k p(\tau_k) = \int_0^{\exposure} p(t)\flux(\spacexy,t)\,d t + \eta_{p,\spacexy}.
    \end{equation}
\end{enumerate}
\end{remark}

\subsubsection{Spatiotemporal spike probing}
In this section, we extend the temporal analysis to the spatiotemporal domain.

\begin{proposition}[Spatiotemporal unified spike probing]
\label{prop:unified_st}
Let $\spikes = \{(\spacexy_i, t_i)\}_{i=1}^N$ be the full spike event stream over the sensor domain $\setspace$ and time interval $[0,\exposure]$, with each pixel following an independent integrate-and-fire model driven by its own Poisson photon stream with common threshold $\Theta$. For any probing function $p(\spacexy,t)$ that is deterministic, bounded, and differentiable in $t$ for each $\spacexy$:
\begin{equation}
    \Theta \sum_{i=1}^{N} p(\spacexy_i, t_i)
    =
    \int_\setspace \int_0^{\exposure} p(\spacexy,t)\,\flux(\spacexy,t)\,d t\,d\spacexy
    + M_p + \eta_p,
    \label{eq:unified_st}
\end{equation}
where $|\eta_p| \leq \int_\setspace \Big[\Theta\big(|p(\spacexy,\exposure)| + |p(\spacexy,0)|\big) + \Theta\int_0^{\exposure} |\partial_t p(\spacexy,t)|\,d t\Big]\,d\spacexy$ and $M_p$ is the martingale inherited from Equation~\ref{eq:flux_probing_integral}.
\end{proposition}

\begin{proof}
Apply Proposition~\ref{prop:unified_1d} at each pixel $\spacexy$ using the temporal probe $t \mapsto p(\spacexy,t)$, then integrate over $\spacexy \in \setspace$. From~\cref{lem:probe-martin}, $M_p$ is a martingale.
\end{proof}

\subsubsection{Fourier probing: null distribution}

We now restrict to Fourier probing functions
\begin{equation}
    p_\freqvec(\spacexy,t) = e^{-j2\pi \freqvec^\top [\spacexy, t]},
\end{equation}
where $\mathbf{f} = (f_x, f_y, f_t)$ is the spatiotemporal frequency.

From~\eqref{eq:unified_st}, the (scaled) spike Fourier probing measurement is
\begin{equation}
    \probemeas_\freqvec = \Theta \sum_{i=1}^N e^{-j2\pi \mathbf{f}^\top [\spacexy_i, t_i]}
    =
    \underbrace{\langle p_\freqvec,\flux\rangle}_{\text{flux probing integral}} + M_{p_\freqvec} + \eta_{p_\freqvec},
    \label{eq:fourier_probing_spike}
\end{equation}

To make the CFAR analysis tractable, we work in the regime where the quantization residual $\eta_{p_\freqvec}$ is negligible relative to the stochastic term. This is justified when:
\begin{itemize}
    \item The threshold $\Theta$ is moderate (not too large), so the quantization error is small, or
    \item A low-pass pre-filter has been applied to suppress the $O(\Theta |f_t|)$ growth of $\eta_{p_\freqvec}$ at high frequencies (as noted in the remark on periodic spike trains).
\end{itemize}

\begin{proposition}\label{prop:spike-nulldist}
    Under the null hypothesis $H_0$, the flux $\flux$ has no energy at frequency $\mathbf{f}$ and $2\mathbf{f}$, thus $\probemeas_\freqvec$ approximates a complex normal distribution with
    \begin{equation}
        \probemean = \begin{bmatrix}
    0\\
    0
    \end{bmatrix} ,\quad
        \probecov = \frac{\bar{N}}{2}I_2.
    \end{equation}
\end{proposition}
\begin{proof}
Under this negligible quantization error $\eta_{p_\freqvec}$, $\probemeas_\freqvec$ approximately follows the probing measurements derived in Equation~\ref{eq:flux_probing_integral}, thus the distribution of $\probemeas_\freqvec$ can be approximated by Equation~\ref{eq:probe-dist-fourier}.

Since $\freqvec$ does not contribute to $\flux$, the flux probing integral $\langle p_\freqvec,\flux\rangle=0$. Evaluating Equation~\ref{eq:probe-dist-fourier} using this integral yields $\mathbb{E}[\probemeas_\freqvec]=0$. \cref{cor:cov-weaksig} provides our covariance $\probecov= (\bar{N}/2)I_2$ under those same conditions, where $\bar{N}=\langle 1,\flux\rangle$ is the expected number of photon arrivals.
\end{proof}

\subsubsection{CFAR detector for spike cameras}

Here, we derive the CFAR detector for spike cameras.

\begin{corollary}[Normalized probing energy under $H_0$]
Define the normalized spike probing energy as follows from the normalized Fourier flux probing energy in Equation~\ref{eq:fourier_energy}:
    \begin{equation}
        |\probemeasnorm_{\freqvec}|^2 = 
        \text{Re}\left(\frac{\probemeas_{\freqvec}}{\sqrt{\boldsymbol{\mathrm{\Sigma}}_{1,1}}}\right)^2+\text{Im}\left(\frac{\probemeas_{\freqvec}}{{\sqrt{\boldsymbol{\mathrm{\Sigma}}_{2,2}}}}\right)^2,
    \end{equation}
    Since we assume $H_0$, then $|\probemeasnorm_{\freqvec}|^2\sim\chi_2^2$.
    Substituting in the covariance and mean from~\cref{prop:spike-nulldist}
    \begin{equation}
    |\probemeasnorm_{\freqvec}|^2 = \frac{2}{\bar{N}} \left|\probemeas_\freqvec\right|^2,
    \label{eq:spike-test_stat}
\end{equation}
\end{corollary}

\begin{proposition}[CFAR detector for spike camera Fourier probing]\label{prop:cfar-spike}
Given a false alarm probability $\alpha$, we detect spatiotemporal frequency $\freqvec\neq 0$ (reject $H_0$) when
\begin{equation}
    \left|\probemeas_\freqvec\right|^2
    \;\geq\;
    \mathrm{CDF}_{\chi^2_2}^{-1}(1-\alpha)\;\cdot\;\frac{\bar{N}}{2}.
    \label{eq:cfar_spike}
\end{equation}

Since $\bar{N}$ (the expected total photon count) is not directly observed, the conservative CFAR bound is used instead:
\begin{equation}
    \left|\probemeas_\freqvec\right|^2
    \;\geq\;
    \mathrm{CDF}_{\chi^2_2}^{-1}(1-\alpha)\;\cdot\;\frac{\Theta N + \Theta |A|}{2},
    \label{eq:cfar_practical}
\end{equation}
where $N$ is the observed number of spikes.
\end{proposition}

\begin{proof}[Proof of~\cref{prop:cfar-spike}]
The CFAR detection threshold for $|\probemeasnorm_{\freqvec}|^2$ is
\begin{equation}
    |\probemeasnorm_{\freqvec}|^2 \geq \mathrm{CDF}_{\chi^2_2}^{-1}(1-\alpha).
\end{equation}
Substituting~\cref{eq:spike-test_stat} yields~\cref{eq:cfar_spike}.

To determine a conservative bound, we have the identity~\eqref{eq:photon_spike_identity} evaluated at $t = \exposure$:
\begin{equation}
    N^{\mathrm{ph}}_{\mathrm{total}} = \Theta \cdot N^{\mathrm{sp}}_{\mathrm{total}} + \varepsilon_{\mathrm{residual}},
\end{equation}
where $N^{\mathrm{sp}}_{\mathrm{total}} = \sum_{\spacexy} N^{\mathrm{sp}}_{\spacexy}(\exposure) = N$ is the total spike count and $\varepsilon_{\mathrm{residual}} = \sum_{\spacexy} \varepsilon_{\spacexy}(\exposure)$.

Since $0 \leq \varepsilon_{\spacexy}(\exposure) < \Theta$ for each pixel, we have
\begin{equation}
    \Theta N \leq N^{\mathrm{ph}}_{\mathrm{total}} < \Theta N + \Theta|\setspace|,
\end{equation}
where $|\setspace|$ is the number of pixels. The quantity $\Theta N$ is a lower-biased proxy for the realized total photon count.
We therefore use $\hat{\bar N}=\Theta N$ as an estimate
in the high-count regime where $\Theta |\setspace| \ll \bar N$, yielding an approximate CFAR threshold.
A conservative alternative is to replace $\Theta N$ by $\Theta N+\Theta |\setspace|$. The practical CFAR criterion becomes:
\begin{equation}
    \left|\probemeas_\freqvec\right|^2
    \;\geq\;
    \mathrm{CDF}_{\chi^2_2}^{-1}(1-\alpha)\;\cdot\;\frac{\Theta N + \Theta |A|}{2}.
    \label{eq:spike-cfar_practical}
\end{equation}
\end{proof}

\begin{remark}[Comparison with the photon CFAR detector]
In the single-photon case ($\Theta = 1$), the spike count equals the photon count ($N = N^{\mathrm{ph}}_{\mathrm{total}}$) and the criterion~\eqref{eq:spike-cfar_practical} becomes
\begin{equation}
    \left|\probemeas_\freqvec\right|^2 \geq \mathrm{CDF}_{\chi^2_2}^{-1}(1-\alpha)\cdot\frac{N}{2},
\end{equation}
which matches Equation~\ref{eq:CFAR3d}, after accounting for the $1/\sqrt{\spacevol}$ normalization convention.
\end{remark}

\begin{remark}[Role of the quantization residual in detection]
The CFAR analysis above assumed $\eta_{p_\freqvec} \approx 0$. When this does not hold (\eg, at high temporal frequencies where $|\eta_{p_\freqvec}|$ can be $O(\Theta |f_t| T)$), the quantization residual acts as an additional bounded bias under $H_0$. This inflates the effective false alarm rate: the test statistic picks up ``energy'' from the sawtooth quantization pattern even when the true flux has no content at frequency $\freqvec$.

This causes artificial harmonic peaks from nearly-periodic spike trains. To maintain the desired false alarm rate, applying a temporal low-pass filter to the spike probing measurement can suppress the high-frequency quantization artifacts before computing $\probemeas_{\freqvec}$.
\end{remark}

\subsection{Event Cameras: Spatiotemporal Probing Derivation}
\label{sec:event_probing_supp}
In this section, we establish the connection of asynchronous event sensors to spatiotemporal probing of the spectrum of $\partial_t \log \flux$. For clarity, we focus on the temporal-only case here and extend to spatiotemporal probing in the next section.

To model events as an estimator for the derivative of the log flux, we adopt the mathematical representation presented in~\cite{scheerlinck_continuous-time_2018}.

\paragraph{Event model (per pixel)}
Fix a pixel location $\mathbf{r}=(x,y)$ and let
\begin{equation}
L_{\spacexy}(t) \;\triangleq\; \log \flux(\spacexy,t),
\end{equation}
denote the log-flux (or log-intensity) over an exposure window $t\in[0,\exposure]$.
In the standard contrast-threshold model, the event camera outputs an event
whenever the change in log-flux reaches a contrast threshold $C>0$. For simplicity, we consider the contrast threshold equal for both ON and OFF events.
Equivalently, there exists a piecewise-constant quantized process
$\widetilde{L}_{\mathbf{r}}(t)$ such that:
\begin{enumerate}
\item \textbf{(Quantization error)} $\;|L_{\mathbf{r}}(t)-\widetilde{L}_{\mathbf{r}}(t)| \le C$ for all $t\in[0,\exposure]$;
\item \textbf{(Jump structure)} $\widetilde{L}_{\mathbf{r}}(t)$ changes only via jumps of size $\pm C$ at the event times
$\{\tau_k(\mathbf{r})\}_{k=1}^{N(\mathbf{r})}$:
\begin{equation}
\Delta \widetilde{L}_{\mathbf{r}}\!\big(\tau_k(\mathbf{r})\big)
\;\triangleq\;
\widetilde{L}_{\mathbf{r}}\!\big(\tau_k(\mathbf{r})^+\big)
-
\widetilde{L}_{\mathbf{r}}\!\big(\tau_k(\mathbf{r})^-\big)
\;=\; \sigma_k(\mathbf{r})\,C,
\qquad \sigma_k(\mathbf{r})\in\{+1,-1\}.
\end{equation}
\end{enumerate}
Define the residual (bounded) error
\begin{equation}
\varepsilon_{\mathbf{r}}(t) \;\triangleq\; L_{\mathbf{r}}(t)-\widetilde{L}_{\mathbf{r}}(t),
\qquad\text{so that}\qquad
|\varepsilon_{\mathbf{r}}(t)|\le C.
\end{equation}

\subsubsection{Temporal event probing}

\begin{definition}[Event field and probing measurements]
For a fixed pixel $\spacexy$, define the signed event measure (also denoted as an event field in Equation 7 from~\cite{scheerlinck_continuous-time_2018})
\begin{equation}
dS_{\spacexy}(t) \;\triangleq\; \sum_{k=1}^{N(\spacexy)} \sigma_k(\spacexy)\,\delta\!\big(t-\tau_k(\spacexy)\big)\,dt,
\end{equation}
so that, for any probing function $p$,
\begin{equation}
\int_{0}^{\exposure} p(t)\,dS_{\mathbf{r}}(t) \;=\; \sum_{k=1}^{N(\mathbf{r})} \sigma_k(\mathbf{r})\,p\!\big(\tau_k(\mathbf{r})\big).
\end{equation}
This is the quantized log-intensity signal with respect to the probing function in Equation 8 from~\cite{scheerlinck_continuous-time_2018}.
\end{definition}

\begin{proposition}[Event probing of $\partial_t \log\phi$]
\label{prop:event_probing_1d}
Assume the contrast-threshold event model above and let $p$ be deterministic, bounded, and differentiable.
Then the signed event probing sum is a linear projection of $\partial_t L_{\mathbf{r}}(t)$ up to a bounded error:
\begin{equation}
C \sum_{k=1}^{N(\mathbf{r})} \sigma_k(\mathbf{r})\,p\!\big(\tau_k(\mathbf{r})\big)
\;=\;
\int_{0}^{\exposure} \probe(t)\,\partial_t L_{\spacexy}(t)\,dt
\;+\; \eta_{p,\mathbf{r}},
\label{eq:event_probing_1d}
\end{equation}
where the residual term satisfies the deterministic bound
\begin{equation}
|\eta_{p,\mathbf{r}}|
\;\le\;
C\Big(|p(\exposure)|+|p(0)|\Big)
\;+\;
C\int_{0}^{\exposure} |p'(t)|\,dt.
\label{eq:event_error_bound_1d}
\end{equation}
This is the analogue to Equation 9 from~\cite{scheerlinck_continuous-time_2018}, the main difference being that the quantization error in our case depends on the probing function.
\end{proposition}

\begin{proof}
Since $L_{\mathbf{r}}(t)=\widetilde{L}_{\mathbf{r}}(t)+\varepsilon_{\mathbf{r}}(t)$, we write the Stieltjes integral
\begin{equation}
\int_{0}^{\exposure} p(t)\,dL_{\mathbf{r}}(t)
\;=\;
\int_{0}^{\exposure} p(t)\,d\widetilde{L}_{\mathbf{r}}(t)
\;+\;
\int_{0}^{\exposure} p(t)\,d\varepsilon_{\mathbf{r}}(t).
\label{eq:split_integral}
\end{equation}
Because $\widetilde{L}_{\mathbf{r}}(t)$ is piecewise constant with jumps $\Delta \widetilde{L}_{\mathbf{r}}(\tau_k)=\sigma_k C$,
\begin{equation}
\int_{0}^{\exposure} p(t)\,d\widetilde{L}_{\mathbf{r}}(t)
\;=\;
\sum_{k=1}^{N(\mathbf{r})} p\!\big(\tau_k(\mathbf{r})\big)\,\Delta \widetilde{L}_{\mathbf{r}}\!\big(\tau_k(\mathbf{r})\big)
\;=\;
C\sum_{k=1}^{N(\mathbf{r})}\sigma_k(\mathbf{r})\,p\!\big(\tau_k(\mathbf{r})\big).
\label{eq:jump_sum}
\end{equation}
For the residual term, apply integration by parts to the Stieltjes integral (valid since both $p$ and $\varepsilon_{\spacexy}$ are bounded and measurable):
\begin{equation}
\int_{0}^{\exposure} p(t)\,d\varepsilon_{\mathbf{r}}(t)
\;=\;
p(\exposure)\varepsilon_{\mathbf{r}}(\exposure)-p(0)\varepsilon_{\mathbf{r}}(0)
\;-\;
\int_{0}^{\exposure} \varepsilon_{\mathbf{r}}(t)\,p'(t)\,dt.
\label{eq:ibp}
\end{equation}
Using $|\varepsilon_{\mathbf{r}}(t)|\le C$ gives
\begin{equation}
\left|\int_{0}^{T} p(t)\,d\varepsilon_{\mathbf{r}}(t)\right|
\;\le\;
C\Big(|p(T)|+|p(0)|\Big)
\;+\;
C\int_{0}^{T} |p'(t)|\,dt.
\end{equation}
Finally, since $dL_{\mathbf{r}}(t)=\partial_t L_{\mathbf{r}}(t)\,dt$ for differentiable $L_{\mathbf{r}}$,
we have $\int_0^{\exposure} p(t)\,dL_{\mathbf{r}}(t)=\int_0^{\exposure} p(t)\partial_t L_{\mathbf{r}}(t)\,dt$.
Combining with \eqref{eq:split_integral}--\eqref{eq:jump_sum} yields the event probing sum~\cref{eq:event_probing_1d}
by setting $\eta_{p,\mathbf{r}} \triangleq \int_0^{\exposure} p(t)\,d\varepsilon_{\mathbf{r}}(t)$ as the bound \eqref{eq:event_error_bound_1d}.
\end{proof}

\subsubsection{Spatiotemporal event probing (proof of Equation 9)}
\label{sec:event_probing_st}

In this section, we extend the temporal analysis to the spatiotemporal domain and derive Equation 9 of the main paper.

\textbf{Equation 9.} The Fourier probing measurements for event cameras are
\begin{equation}
    \probemeas_\freqvec = \sum_{(\spacex_i,\spacey_i,\timesym_i)\in\events}\eventpol_i\probe_\freqvec(\spacex_i,\spacey_i,\timesym_i).
    \tag{9}
    \label{eq:event-probing}
\end{equation}

\paragraph{Event stream over the sensor}
Let $\setspace$ denote the set of pixel locations (or a continuous sensor domain with pixel area $|\setspace|$).
The event camera outputs a set of events
\begin{equation}
\dvsevs \;=\; \{(\mathbf{r}_i,t_i,\sigma_i)\}_{i=1}^{N},
\qquad \mathbf{r}_i=(x_i,y_i)\in\setspace,\;\; t_i\in[0,\exposure],\;\; \sigma_i\in\{+1,-1\}.
\end{equation}
Define the global signed event measure on $\setspace\times[0,T]$:
\begin{equation}
dS(\mathbf{r},t)
\;\triangleq\;
\sum_{i=1}^{N}\sigma_i\,\delta(\mathbf{r}-\mathbf{r}_i)\,\delta(t-t_i)\,d\mathbf{r}\,dt,
\end{equation}
so that for any probing function $p(\mathbf{r},t)$,
\begin{equation}
\int_{\setspace}\int_{0}^{\exposure} p(\mathbf{r},t)\,dS(\mathbf{r},t)
\;=\;
\sum_{i=1}^{N}\sigma_i\,p(\mathbf{r}_i,t_i).
\label{eq:global_event_sum}
\end{equation}
Substituting $p(\mathbf{r}, t)=\probe_{\freqvec}(\spacetime)=e^{-j2\pi\freqvec^\top\spacetime}$ in~\cref{eq:global_event_sum} yields Equation 9 of the main paper.

\begin{proposition}[Spatiotemporal event probing]
\label{prop:event_probing_st}
Assume the per-pixel contrast-threshold model holds at every $\mathbf{r}\in\setspace$, \ie,
$L_{\mathbf{r}}(t)=\log\phi(\mathbf{r},t)$ admits a quantized approximation $\widetilde{L}_{\mathbf{r}}(t)$ with
$|L_{\mathbf{r}}(t)-\widetilde{L}_{\mathbf{r}}(t)|\le C$ and jumps of size $\pm C$ at event times with polarities.
Let $p(\mathbf{r},t)$ be bounded and differentiable in $t$ for each $\mathbf{r}$.
Then
\begin{equation}
C\sum_{i=1}^{N}\sigma_i\,p(\mathbf{r}_i,t_i)
\;=\;
\int_{\setspace}\int_{0}^{\exposure} p(\mathbf{r},t)\,\partial_t \log\phi(\mathbf{r},t)\,dt\,d\mathbf{r}
\;+\; \eta_p,
\label{eq:event_probing_st}
\end{equation}
where the residual term $\eta_p$ is bounded by the spatial integral of the per-pixel bounds:
\begin{equation}
|\eta_p|
\;\le\;
\int_{\setspace}
\left[
C\Big(|p(\mathbf{r},T)|+|p(\mathbf{r},0)|\Big)
+
C\int_{0}^{\exposure} \big|\partial_t p(\mathbf{r},t)\big|\,dt
\right]
d\mathbf{r}.
\label{eq:event_probing_st_bound}
\end{equation}
\end{proposition}

\begin{proof}
Apply~\cref{prop:event_probing_1d} independently to each pixel $\mathbf{r}$ with probing function $t\mapsto p(\mathbf{r},t)$,
then sum the resulting identities over $\mathbf{r}\in\setspace$ (or integrate over $\Omega$ in the continuous-domain view).
The left-hand side becomes $\sum_i \sigma_i p(\mathbf{r}_i,t_i)$ as in \eqref{eq:global_event_sum}, while the right-hand side becomes
the space--time integral of $p(\mathbf{r},t)\partial_t \log\phi(\mathbf{r},t)$ plus the sum (or integral) of residual terms.
The bound \eqref{eq:event_probing_st_bound} follows by summing the per-pixel bound \eqref{eq:event_error_bound_1d}.
\end{proof}

\subsubsection{Simplified noise model for events}

While~\cref{prop:event_probing_st} provides an upper bound for the error of the probing measurements, it does not represent a stochastic model to derive a CFAR detector. We introduce background activity (BA) as the main source of shot noise, where circuit junction leakage and temperature periodically produce ON events~\cite{lichtsteiner_128times_2008}.

To keep the model simple, we make the assumption that BA events originate from a homogeneous Poisson point process, where the events are biased towards ON events. This is a common assumption used in event denoising methods~\cite{feng_event_2020,khodamoradi_onon-space_2021} and event camera simulators~\cite{hu_v2e_2021}.

\begin{definition}[Biased background activity model]
\label{def:ba_model}
Given a pixel $\spacexy\in\setspace$, the background activity model assumes the following:
\begin{enumerate}[label=(\roman*)]
    \item Background events arrive at $\spacexy$ with homogeneous Poisson process rate $\baeventrate$ events/second. For simplicity, we assume all pixels adopt the same rate.
    \item Each background activity event has polarity $\sigma_i=\pm1$, where $\sigma_i=+1$ with probability $q>1/2$ due to ON bias, and $\sigma_i=-1$ with probability $1-q$, independently drawn.
    \item Background activity events are independent of other events and between pixels, which is a standard simplification that neglects refractory-period interactions between the two streams.
\end{enumerate}
We denote the events coming from real signal as $\dvsevs_{\text{sig}}$, and those coming from background activity as $\dvsevs_{\text{BA}}$, thus we have that
\begin{equation}
    \dvsevs = \dvsevs_{\text{sig}}\cup \dvsevs_{\text{BA}}.
\end{equation}
The total expected number of background activity events over the sensor and observation window is
\begin{equation}
    \mathbb{E}[\dvsevs_{\text{BA}}]=\bar{N}_{\text{BA}} = \int_{\setspace}\int_0^{\exposure}\baeventrate\,dt\,d\spacexy = |\setspace|\,\baeventrate\,\exposure.
\end{equation}
\end{definition}

\begin{corollary}
\label{cor:polarity_moments}
For any BA event $(\spacexy,t,\sigma)\in\dvsevs_{\text{BA}}$, its polarity $\sigma\in\{+1,-1\}$ satisfies:
\begin{align}
    \mathbb{E}[\sigma] &= 2q - 1 \triangleq b, \label{eq:polarity_mean}\\
    \mathbb{E}[\sigma^2] &= 1, \label{eq:polarity_second}\\
    \mathrm{Var}(\sigma) &= 1 - b^2 = 4q(1-q). \label{eq:polarity_var}
\end{align}
We refer to $b\in(-1,1]$ as the \emph{polarity bias}. Note that $\mathbb{E}[\sigma^2]=1$ identically since $\sigma^2=1$ for all events regardless of polarity.
\end{corollary}

\subsubsection{Fourier probing: null distribution}

Here, we derive the distribution of the null hypothesis of the probing measurements $H_0$ for the purposes of CFAR detection.

To make the CFAR analysis tractable, we work in the regime where the quantization error $\eta_p$ is small. That is, since the bound of $\eta_p$ scales linearly with frequency $\freqvec$ (due to the $\int_{0}^{\exposure} \big|\partial_t p(\mathbf{r},t)\big|\,dt$ in~\cref{eq:event_probing_st_bound}), the quantization error is small if the scene does not vary too quickly. Naturally, this makes faster scenes harder to reconstruct.

Using the Fourier basis as our probing function, denote our polarity-weighted probing measurements as
\begin{equation}
    \probemeas_\freqvec = \probemeas_\freqvec^{\text{sig}} + \probemeas_\freqvec^{\text{BA}},
\end{equation}
such that $\probemeas_\freqvec^{\text{sig}} = C\sum_{i=1}^{N_{\text{sig}}}\sigma_i e^{-j2\pi\freqvec^\top(\spacexy_i,t_i)}$ and $\probemeas_\freqvec^{\text{BA}} = C\sum_{i=1}^{N_{\text{BA}}}\sigma_i e^{-j2\pi\freqvec^\top(\spacexy_i,t_i)}$.

Note that from~\cref{prop:event_probing_st}, $\probemeas_\freqvec^{\text{sig}}$ is deterministic up to a bounded error.

Under the null hypothesis $H_0$ that $\freqvec$ is not a component of $\partial_t\log\flux(\spacexy,t)$, we expect $\probemeas_\freqvec^{\text{sig}}=0$, therefore $\probemeas_\freqvec=\probemeas_\freqvec^{\text{BA}}$.

\begin{proposition}[Mean and variance of $P_\freqvec^{\text{BA}}$ under $H_0$]
\label{prop:ba_meanvar}
Under the background activity model (\cref{def:ba_model}) with symmetric contrast threshold $C$, the background activity probing measurement $\probemeas_\freqvec^{\text{BA}}$ can be modeled as a functional of a marked Poisson point process~\cite{last_lectures_2017}, since $\dvsevs_{\text{BA}}$ itself is a homogeneous Poisson point process. For $\freqvec\neq 0$, its mean and variance are as follows:
\begin{align}
    \mathbb{E}\big[\probemeas_\freqvec^{\text{BA}}\big] &= 0\\
    \mathrm{Var}\big(\probemeas_\freqvec^{\text{BA}}\big) &= C^2\bar{N}_{\text{BA}}.
\end{align}
We restrict frequencies to values that are integer multiples of the grid resolution ($1/\width$,$1/\height$,$1/\exposure$ for $\freqx,\freqy,\freqt$ respectively), so that the Fourier bases are orthonormal and integrate over full periods over the domain.
\end{proposition}

\begin{proof}[Proof of~\cref{prop:ba_meanvar}]
By Campbell's theorem (\cref{lemma:campbell}), the mean and variance of a sum over a Poisson process are:
\begin{align}
    \mathbb{E}\big[\probemeas_\freqvec^{\text{BA}}\big]
    &= \lambda_{\text{BA}}\sum_{\sigma\in\{+1,-1\}} g(\sigma)
    \int_\setspace\!\int_0^{\exposure} C\sigma\,e^{-j2\pi\freqvec^\top(\spacexy,t)}\,dt\,d\spacexy, \label{eq:campbell_mean}\\
    \mathrm{Var}\big(\probemeas_\freqvec^{\text{BA}}\big)
    &= \lambda_{\text{BA}}\sum_{\sigma\in\{+1,-1\}} g(\sigma)
    \int_\setspace\!\int_0^{\exposure} \big|C\sigma\,e^{-j2\pi\freqvec^\top(\spacexy,t)}\big|^2\,dt\,d\spacexy, \label{eq:campbell_var}
\end{align}
where $g(+1)=q$ and $g(-1)=1-q$.

\textbf{Mean.} Evaluating \eqref{eq:campbell_mean}, and using the fact that polarity $\sigma$ is independent of the event location:
\begin{equation}
    \mathbb{E}\big[\probemeas_\freqvec^{\text{BA}}\big]
    = C\lambda_{\text{BA}}\,\underbrace{\mathbb{E}[\sigma]}_{=\,b}
    \int_\setspace\!\int_0^{\exposure} e^{-j2\pi\freqvec^\top(\spacexy,t)}\,dt\,d\spacexy.
    \label{eq:ba_mean}
\end{equation}
We restrict to frequencies
that are integer multiples of the grid resolution $(1/\width,1/\height,1/\exposure)$, so that the Fourier basis functions complete full periods over the observation volume. As a consequence, the Fourier basis is orthonormal and
$\int_\setspace\!\int_0^{\exposure} e^{-j2\pi\freqvec^\top(\spacexy,t)}\,dt\,d\spacexy=0$, therefore
\begin{equation}
    \mathbb{E}\big[\probemeas_\freqvec^{\text{BA}}\big] = 0, \qquad \freqvec\neq\mathbf{0}.
    \label{eq:ba_mean_zero}
\end{equation}
For the DC component $\freqvec=\mathbf{0}$, the mean is $Cb\,\baeventrate\,|\setspace|\,\exposure = Cb\,\bar{N}_{\text{BA}}$.

\textbf{Variance.} Evaluating \eqref{eq:campbell_var}:
\begin{align}
    \mathrm{Var}\big(\probemeas_\freqvec^{\text{BA}}\big)
    &= C^2\baeventrate\,\underbrace{\sum_{\sigma\in\{+1,-1\}} g(\sigma)\,\sigma^2}_{=\,\mathbb{E}[\sigma^2]\,=\,1}\,
    \int_\setspace\!\int_0^{\exposure}
    \underbrace{\big|e^{-j2\pi\freqvec^\top(\spacexy,t)}\big|^2}_{=\,1}
    \,dt\,d\spacexy \nonumber\\
    &= C^2\baeventrate\,|\setspace|\,\exposure
    = C^2\bar{N}_{\text{BA}}.
    \label{eq:ba_var}
\end{align}
\end{proof}

\begin{proposition}[Covariance of real and imaginary parts]\label{prop:ba_cov}
Let $\probemeas_\freqvec^{\text{BA}} = X_\freqvec + jY_\freqvec$ such that
$X_\freqvec = C\sum_{i\in\dvsevs_{\text{BA}}}\sigma_i\cos(2\pi\freqvec^\top(\spacexy_i,t_i))$ and
$Y_\freqvec = -C\sum_{i\in\dvsevs_{\text{BA}}}\sigma_i\sin(2\pi\freqvec^\top(\spacexy_i,t_i))$, we have that
\begin{equation}
    \Sigma = \mathrm{Var}(X_\freqvec) = \mathrm{Var}(Y_\freqvec) = \frac{C^2\bar{N}_{\text{BA}}}{2}, \qquad
    \mathrm{Cov}(X_\freqvec,Y_\freqvec) = 0.
    \label{eq:equal_variance}
\end{equation}
\end{proposition}

\begin{proof}[Proof of~\cref{prop:ba_cov}]
Campbell's theorem applied separately gives:
\begin{align}
    \mathrm{Var}(X_\freqvec) &= C^2\lambda_{\text{BA}}
    \int_\setspace\!\int_0^{\exposure} \cos^2\!\big(2\pi\freqvec^\top(\spacexy,t)\big)\,dt\,d\spacexy, \label{eq:var_X}\\
    \mathrm{Var}(Y_\freqvec) &= C^2\lambda_{\text{BA}}
    \int_\setspace\!\int_0^{\exposure} \sin^2\!\big(2\pi\freqvec^\top(\spacexy,t)\big)\,dt\,d\spacexy, \label{eq:var_Y}\\
    \mathrm{Cov}(X_\freqvec,Y_\freqvec) &= -C^2\lambda_{\text{BA}}
    \int_\setspace\!\int_0^{\exposure} \cos(2\pi\freqvec^\top(\spacexy,t))\sin(2\pi\freqvec^\top(\spacexy,t))\,dt\,d\spacexy. \label{eq:cov_XY}
\end{align}
We restrict to frequencies that are orthonormal and integrate over full periods in the domain. As a consequence, the integrals of the Fourier basis evaluate to
\begin{align}
    \int_\setspace\!\int_0^{\exposure} \cos^2\!\big(2\pi\freqvec^\top(\spacexy,t)\big)\,dt\,d\spacexy &= \frac{|\setspace|\,\exposure}{2}, \label{eq:cos2_exact}\\
    \int_\setspace\!\int_0^{\exposure} \sin^2\!\big(2\pi\freqvec^\top(\spacexy,t)\big)\,dt\,d\spacexy &= \frac{|\setspace|\,\exposure}{2}, \label{eq:sin2_exact}\\
    \int_\setspace\!\int_0^{\exposure} \cos\!\big(2\pi\freqvec^\top(\spacexy,t)\big)\sin\!\big(2\pi\freqvec^\top(\spacexy,t)\big)\,dt\,d\spacexy &= 0, \label{eq:cossin_exact}
\end{align}
using the following properties: $\cos^2\theta = \frac{1}{2}(1+\cos 2\theta)$, $\sin^2\theta = \frac{1}{2}(1-\cos 2\theta)$, $\cos\theta\sin\theta = \frac{1}{2}\sin 2\theta$, and the double-frequency terms integrating to zero on the grid. Substituting and evaluating $\bar{N}_{\text{BA}} = \lambda_{\text{BA}}|\setspace|\,\exposure$:
\begin{equation}
    \mathrm{Var}(X_\freqvec) = \mathrm{Var}(Y_\freqvec) = \frac{C^2\bar{N}_{\text{BA}}}{2}, \qquad
    \mathrm{Cov}(X_\freqvec,Y_\freqvec) = 0.
\end{equation}
\end{proof}

\begin{proposition}[Asymptotic normality]\label{prop:ba_normal}
For large $\bar{N}_{\text{BA}}$, the CLT for Poisson integrals gives:
\begin{equation}
    \probemeas_\freqvec^{\text{BA}} \;\overset{d}{\approx}\; \mathcal{CN}\!\left(0,C^2 \bar{N}_{\text{BA}}\right), \qquad \freqvec\neq\mathbf{0},
    \label{eq:ba_clt}
\end{equation}
\ie, $X_\freqvec$ and $Y_\freqvec$ are approximately independent each with $\mathcal{N}(0,C^2 \bar{N}_{\text{BA}}/2) = \mathcal{N}(0,\Sigma)$.
\end{proposition}

\begin{proof}[Proof of~\cref{prop:ba_normal}]
Conditioned on $N_{\mathrm{BA}}=n$, the quantity
\[
\probemeas_f^{\mathrm{BA}} = C\sum_{i=1}^{n} \sigma_i e^{-j2\pi \mathbf{f}^\top(\mathbf{r}_i,t_i)}
\]
is a sum of $n$ i.i.d.\ bounded complex random variables. For nonzero Fourier frequencies, these summands have mean zero, and their real and imaginary parts have equal variance. Therefore, for sufficiently large $n$, the multivariate central limit theorem implies
\[
\probemeas_f^{\mathrm{BA}} \mid (N_{\mathrm{BA}}=n)
\;\overset{d}{\approx}\;
\mathcal{CN}(0,\, C^2 n).
\]
Taking expectation over $N_{\text{BA}}$ and replacing $n$ with its 
expectation $\bar{N}_{\text{BA}}$ gives the unconditional approximation 
in~\cref{eq:ba_clt}.
\end{proof}

With the null distribution established, we now derive the CFAR detector.

\subsubsection{CFAR detector for event cameras}

\begin{corollary}[Normalized probing energy under $H_0$]
\label{cor:energy_null}
Define the normalized probing energy at frequency $\freqvec\neq\mathbf{0}$:
\begin{equation}
    \probemeasnorm_\freqvec \;\triangleq\;
    \frac{X_\freqvec^2}{\Sigma} + \frac{Y_\freqvec^2}{\Sigma}
    = \frac{|\probemeas_\freqvec|^2}{\Sigma},
    \quad \Sigma = \frac{C^2\bar{N}_{\text{BA}}}{2}
    \label{eq:normalized_energy}
\end{equation}
Under $H_0(\freqvec)$, where no signal is present at frequency $\freqvec$, we have $\probemeas_\freqvec = \probemeas_\freqvec^{\text{BA}}$, and thus $\probemeasnorm_\freqvec \sim \chi^2_2$ (chi-squared with 2 degrees of freedom) asymptotically.
\end{corollary}

\begin{proposition}[CFAR detector for event camera Fourier probing]
\label{prop:event-cfar}
Given a false alarm probability $\alpha$, we detect spatiotemporal frequency $\freqvec\neq\mathbf{0}$ (reject $H_0$) if and only if
\begin{equation}
    |\probemeas_\freqvec|^2 \;\geq\; \text{CDF}_{\chi^2_2}^{-1}(1-\alpha) \frac{C^2 \bar{N}_{\text{BA}}}{2}.
\label{eq:event-cfar}
\end{equation}
However, since $\bar{N}_{\text{BA}}$ is not directly observable, we instead use the total event count $N$ as a conservative upper bound ($\bar{N}_{\text{BA}} \leq N$ under most cases).
\begin{equation}
    |\probemeas_\freqvec|^2 \;\geq\; \text{CDF}_{\chi^2_2}^{-1}(1-\alpha) \frac{C^2N}{2}.
\label{eq:event-cfar-overest}
\end{equation}
This more conservative bound produces fewer detections, where the actual false alarm rate is at most $\alpha$.
\end{proposition}

\begin{proof}
Under $H_0$, $\probemeasnorm_\freqvec = |\probemeas_\freqvec|^2/\Sigma \sim \chi^2_2$. Thus, evaluating with respect to the CDF of $\chi_2^2$ gives
\begin{equation}
    |\probemeas_\freqvec|^2 \geq \text{CDF}_{\chi^2_2}^{-1}(1-\alpha)\Sigma = \text{CDF}_{\chi^2_2}^{-1}(1-\alpha) \frac{C^2 \bar{N}_{\text{BA}}}{2}.
\end{equation}
\end{proof}

\clearpage
\section{Photon-Location Prediction and Global Flicker: an Intuitive Analysis for bit2bit}
\label{sec:supp_bit2bit_flicker}

In our experiments, bit2bit~\cite{liu2024bit2bit} often preserves fine spatial detail, yet can attenuate weak scene-wide illumination flicker. We do not view this as a fundamental limitation of the method. Rather, we offer the following intuition for why a photon-location prediction objective, when trained on short spatiotemporal crops, may provide a stronger and more consistent learning signal for spatial structure than for weak global temporal modulation.

\paragraph{Setup}
Consider a simple global-flicker model
\begin{equation}
  \phi(x,y,t) = A(t)\,\lambda(x,y),
  \label{eq:bit2bit_flux_sep}
\end{equation}
where $\lambda(x,y)$ is a fixed spatial pattern and $A(t)$ is a time-varying gain shared by all pixels. Under this model, flicker changes primarily \emph{when} photons arrive, while leaving their relative \emph{spatial distribution} essentially unchanged.

\paragraph{A simple distributional view}
Let $p(i,t)$ denote the normalized distribution of held-out target detections over a crop, where $i$ indexes spatial location and $t$ indexes time bin. Under~\eqref{eq:bit2bit_flux_sep}, this distribution factorizes as
\begin{equation}
  p(i,t) = p_t\,s_i,
  \label{eq:bit2bit_factorized_target}
\end{equation}
where $p_t$ is the temporal marginal induced by $A(t)$ and $s_i$ is the normalized spatial pattern induced by $\lambda(x,y)$.

For intuition, suppose the model's predicted distribution can be viewed in the form
\begin{equation}
  q_\theta(i,t) = q_\theta(t)\,q_\theta(i\mid t).
  \label{eq:bit2bit_factorized_pred}
\end{equation}
Then the expected negative log-likelihood of $M$ held-out detections can be written as
\begin{equation}
  \mathbb{E}[\mathcal{L}(\theta)]
  =
  \mathrm{const}
  +
  M\,\mathrm{KL}\!\bigl(p_t \,\|\, q_\theta(t)\bigr)
  +
  M\sum_t p_t\,\mathrm{KL}\!\bigl(s \,\|\, q_\theta(\cdot\mid t)\bigr).
  \label{eq:bit2bit_kl_view}
\end{equation}
Here, $M=\sum_{t=1}^{T}\sum_{i=1}^{N} w_{i,t}$ is the number of held-out target detections in the crop, $w_{i,t}\in\{0,1\}$ indicates whether voxel $(i,t)$ contains a held-out target detection, and $\mathrm{KL}(p\|q)=\sum_x p(x)\log\!\frac{p(x)}{q(x)}$ denotes the Kullback--Leibler divergence. 
Equation~(\ref{eq:bit2bit_kl_view}) should be read as an \emph{interpretive} decomposition rather than an exact statement about the implementation of bit2bit~\cite{liu2024bit2bit}. Its purpose is to highlight that, under global flicker, the temporal signal is carried only by the low-dimensional marginal $p_t$, whereas the spatial term reflects the full spatial arrangement of detections within each crop.

\paragraph{Why weak flicker may be de-emphasized}
Under global flicker, the spatial pattern $s_i$ is approximately time-invariant, so every held-out detection contributes consistent supervision for spatial prediction. By contrast, the only cue for flicker is the variation of the 1D temporal marginal $p_t$ across the crop. If that within-crop variation is weak, then a temporally smoothed prediction $q_\theta(t)$ may incur only a modest penalty, even when the model fits spatial structure well.

This effect is expected to be strongest when the crop duration is short relative to the flicker period. For example, if
\begin{equation}
  A(t) = 1 + \epsilon \sin(2\pi f t),
\end{equation}
then over a crop of duration $\Delta_{\mathrm{win}} = T\Delta t$, the maximum change in gain is
\begin{equation}
  \Delta A_{\max}
  =
  2\epsilon \left|\sin\!\left(\pi f \Delta_{\mathrm{win}}\right)\right|.
  \label{eq:bit2bit_crop_mod}
\end{equation}
For $f=120\,\mathrm{Hz}$, frame rate $100$~kHz, and $T=32$ (this is bit2bit's default temporal crop~\cite{liu2024bit2bit}), we have $\Delta_{\mathrm{win}}=0.32\,\mathrm{ms}$, which gives
\begin{equation}
  \Delta A_{\max} \approx 0.048
  \qquad \text{when } \epsilon=0.2.
\end{equation}
Thus, even a $20\%$ sinusoidal flicker produces less than $5\%$ variation within such a crop, so the temporal histogram of detections can appear nearly flat at crop scale.

\paragraph{Interpretation}
The above analysis does not imply that bit2bit cannot represent flicker. Rather, it suggests that, in the short-crop low-photon regime considered in this work, the objective may offer a stronger and more stable incentive to refine spatial photon placement than to preserve weak crop-wide temporal modulation. This provides one possible explanation for why bit2bit can produce visually sharp reconstructions while still under-emphasizing subtle global flicker in our experiments (see Fig.~1, top row, and Fig.~4, rows~1 and~3 of the main
paper; the temporal smoothing is most clearly visible in the videos on the
supplemental webpage).

\clearpage
\section{Limitations of Per-Pixel Temporal Analysis Under Motion}
\label{sec:1dvs3d}

\begin{figure}[b]
  \centering
  \includegraphics[width=\textwidth]{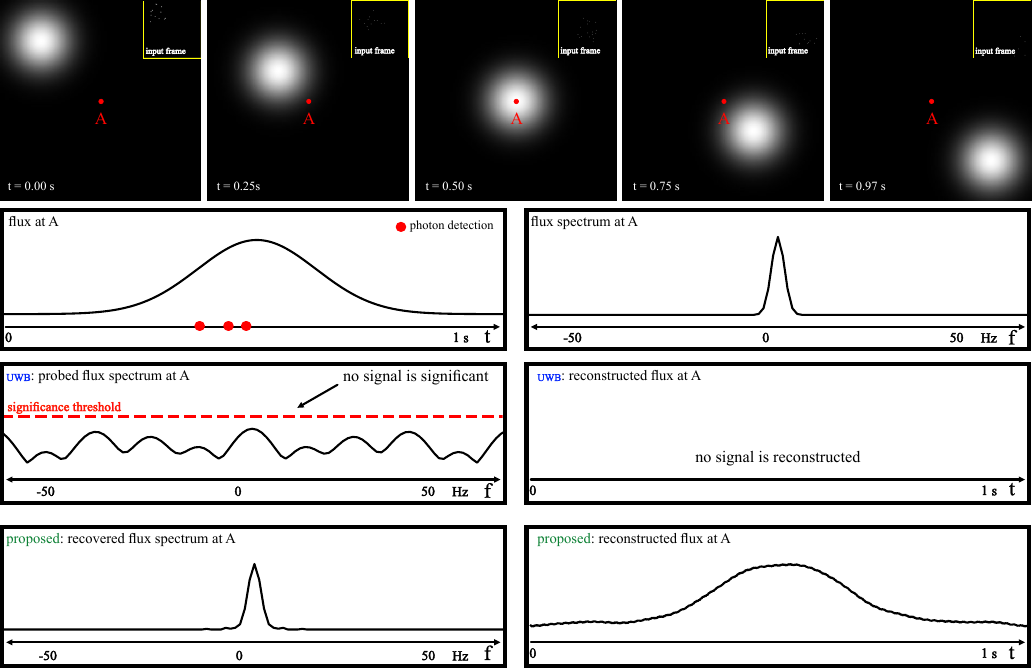}
  \caption{\textbf{Failure of pixelwise temporal probing under motion and extreme photon sparsity.}
  \textbf{Row 1:} Spatiotemporal flux frames of a translating Gaussian blob, with corresponding binary input photon frames shown in the insets. As the blob passes through point A, the local flux varies smoothly in time, while the observed binary frames contain only a few isolated photon arrivals.
  \textbf{Row 2:} Ground truth temporal flux at pixel A, along with the three detected photons (left), and the ground truth flux spectrum (right).
  \textbf{Row 3:} UWB~\cite{wei2023passive} applied at pixel A: the per-pixel temporal spectrum remains below the significance threshold at all frequencies, so no component is detected and the temporal signal is not reconstructed. This is due to extreme photon sparsity at pixel A and the absence of strong periodicity in the local signal.
  \textbf{Row 4:} By aggregating photons across space and time rather than relying solely on detections at pixel $A$, our method recovers a clean spectrum and reconstructs a temporal signal that closely matches the ground-truth flux.}
  \label{fig:1duwb-fail}
\end{figure}

\paragraph{Why per-pixel temporal methods fail under motion}
When a scene moves, photon detections are spread across many pixels
rather than concentrated at a fixed location. At any single pixel, the
object is visible only during the brief interval in which it passes
through, so only a small fraction of the total photon budget contributes
to the local temporal signal. Under limited photon counts, this
sparsity can be severe enough that the per-pixel temporal spectrum never
exceeds the CFAR threshold~\cite{wei2023passive} at any frequency, causing the reconstruction
to fail entirely.

\paragraph{Illustration}
Figure~\ref{fig:1duwb-fail} demonstrates this effect using a
translating Gaussian blob. Although the blob is smooth and its temporal
profile at a fixed pixel is a simple pulse (Row~2, left), only three
photons are detected at pixel A over the full acquisition. The
resulting per-pixel spectrum (Row~3, left) is noise-dominated and
remains below the significance threshold at all frequencies, so UWB~\cite{wei2023passive}
reconstructs no signal at A.

\paragraph{Global spatiotemporal probing}
The key observation is that per-pixel methods discard the spatial
dimension entirely: each pixel is processed independently using only
its own detections. By contrast, spatiotemporal probing computes 3D
Fourier coefficients by summing over all photon arrivals across both
space and time, aggregating evidence from the entire sensor rather than
a single pixel. Even though pixel A receives only three photons, the
global photon count is orders of magnitude larger. Our method recovers
a clean spectrum (Row~4, left) and reconstructs a temporal signal at
A that closely matches the ground truth (Row~4, right).

\clearpage
\section{Implementation Details}

\subsection{Single-photon videography}

We discuss implementation details of performing spatiotemporal flux probing on different single-photon cameras.

\subsubsection{Frequency selection and spatiotemporal normalization}

The spatiotemporal frequency grid is determined by the physical extent of
the acquisition window. The frequency resolution in each dimension is
$1/\width$, $1/\height$, and $1/\exposure$ in $x$, $y$, and $t$ respectively,
and we probe integer multiples of these resolutions up to the Nyquist
limits set by the sensor's spatial and temporal sampling rates.

To ensure orthonormality of the Fourier basis with respect to the integral over the physical spatiotemporal volume, we normalize each probing function by $1/\sqrt{\spacevol}$, where $\spacevol = \width \cdot \height \cdot \exposure$ is the total physical volume of the window. The normalized basis functions are then
\begin{equation}
    \probe_{\freqvec}(\spacetime) = \frac{1}{\sqrt{\spacevol}} e^{-j 2\pi \freqvec^\top \spacetime},
\end{equation}
which satisfy
\begin{equation}
    \langle \probe_{\freqvec}, \probe_{\freqvec'} \rangle = \begin{cases}
        0 & \freqvec\neq\freqvec' \\
        1 & \freqvec = \freqvec'
    \end{cases}
\end{equation}
under the physical-domain integral. Crucially, this normalization ensures that the reconstructed flux $\hat{\flux}(\spacetime)$ is expressed in the correct physical units. 

The computation of the Fourier probing measurements depends on the
sensor type. For quanta sensors~\cite{pi_imaging_spad_2021} the binary readout produces regularly sampled frames, so the
probing measurements can be computed efficiently via the fast Fourier transform (FFT)~\cite{cooley_algorithm_1965}. For SPAD
arrays~\cite{micro_photon_devices_pdm_nodate}, photon arrivals are irregularly sampled in time, so we compute
the probing measurements via direct sum of the terms, which evaluates the Fourier sum at the irregular photon
timestamps without binning into discrete frames.

\subsubsection{Selection of the CFAR significance level $\cfarsigval$}
The significance level $\cfarsigval$ controls the tradeoff between sensitivity to weak frequencies and robustness to noise. A larger $\cfarsigval$ lowers the detection threshold, admitting more frequencies and recovering finer spatiotemporal detail, but increases the expected number of false alarms. A smaller $\cfarsigval$ raises the threshold, suppressing spurious detections at the cost of potentially missing weaker signal components.

A natural choice is $\cfarsigval = 1/|\mathcal{F}|$, where $|\mathcal{F}|$ is the number of probed frequencies. Under the CFAR model this corresponds to controlling the expected number of false detections across the entire frequency grid to approximately one. In practice, however, we found this choice to be overly conservative: the resulting threshold suppresses many high-frequency components and produces over-smoothed reconstructions.

In all experiments we therefore use $\cfarsigval = 10^{-4}$, which provides a good balance between recovering fine spatiotemporal structure and limiting spurious frequency detections.

\subsubsection{Short-time Fourier transform for quanta sensors}
\label{sec:stft}

The spatiotemporal flux probing framework places no constraints on how
the probing functions partition or aggregate the spatiotemporal volume---it
holds for any bounded, measurable $\probe$, including globally supported
ones. In principle, one can probe the entire acquisition at once, and for
SPAD arrays this is what we do: the smaller sensor format and shorter
acquisition windows make global Fourier probing practical.

For quanta sensors, however, larger spatial extent and longer acquisition windows make a single global Fourier model both less practical and less well matched to the nonstationary structure of natural scenes. Sharp
spatiotemporal boundaries produce broadband spectral leakage under a
global Fourier basis, and hard frequency thresholding then introduces
Gibbs-like ringing artifacts in the
reconstruction~\cite{harris_use_1978, gonzalez_digital_2018}. This is a
well-known limitation of the Fourier basis under nonstationarity, not of
the probing theory itself. We therefore adopt the short-time Fourier
transform (STFT)~\cite{crochiere_weighted_1980,allen_unified_1977} with
Hann windows~\cite{harris_use_1978} and 75\% overlap. Our default window
size is $100 \times 100$ pixels $\times\;10{,}000$ frames. This trades
global frequency resolution for spatiotemporal localization, and
empirically produces more coherent reconstructions of scenes with
spatially and temporally varying structure.

Even with windowing, the spatiotemporal context of each STFT block far
exceeds that of existing methods. Each window aggregates
$100 \times 100 \times 10{,}000 = 10^8$ spatiotemporal samples, providing
roughly $300\times$ more temporal context than bit2bit~\cite{liu2024bit2bit} and $100\times$
more spatial context than per-pixel temporal analysis like UWB\footnote{Quanta image sensors output $\sim$100k binary frames per
second, so UWB~\cite{wei2023passive} probes each pixel independently
using all $\sim10^5$ temporal samples. Our STFT window aggregates
$100\times100\times10{,}000 = 10^8$ spatiotemporal samples---roughly
$1000\times$ more total measurements per probed region, at the cost of
using $10\times$ fewer temporal frames per pixel.}. The STFT therefore
preserves the core advantage of spatiotemporal probing---pooling photon
evidence across a large spatiotemporal volume---while adapting to the
local structure of natural scenes. In other words, the STFT is not a theoretical necessity but a modeling
choice: natural scenes are spatiotemporally nonstationary, and local
windows assume only that the flux is locally well-approximated by a
superposition of sinusoids---a much weaker assumption than global
stationarity.

\subsubsection{Windowed CFAR detection and reconstruction}
When using the short-time Fourier transform for quanta image sensors, the global CFAR detector
and reconstruction must be adapted to operate within each local window.
Define $w(\spacetime)$ as the window function.
Within each spatiotemporal window $\setspacetimesub_k$ of volume $\spacevol_k = |\setspacetimesub_k|$, frequency
detection proceeds exactly as in the global setting (Equation~\ref{eq:CFAR3d}), but with the probing function
$\probe_{\freqvec,k}(\spacetime) =
\frac{1}{\sqrt{\spacevol_k}}w(\spacetime-\spacetime_k)e^{-j2\pi\freqvec^\top\spacetime}$.
Approximating its covariance matrix as the diagonal matrix in~\cref{cor:cov-weaksig}, we obtain
\begin{equation}
    \text{Var}\!\left(\text{Re}[\probemeas_{\freqvec,k}]\right)
    \approx
    \text{Var}\!\left(\text{Im}[\probemeas_{\freqvec,k}]\right)
    \approx
    \frac{\langle w(\spacetime - \spacetime_k)^2, \flux \rangle}{2\spacevol_k}
    \approx
    \frac{|\events|_{w,k}}{2\spacevol_k},
\end{equation}
where the window-weighted photon count
\begin{equation}
    |\events|_{w,k} = \sum_{\spacetime \in \events}
    w(\spacetime - \spacetime_k)^2
\end{equation}
is an empirical estimate of the effective photon count within window $k$,
with each photon weighted by the squared window value at its location.
We detect frequency $\freqvec$ within window $k$ with
significance level $\cfarsigval$ if
\begin{equation}
    |\probemeas_{\freqvec,k}|^2 \geq
    \textsc{CDF}_{\chi_2^2}^{-1}(1-\cfarsigval)
    \frac{|\events|_{w,k}}{2\spacevol_k}.
    \label{eq:windowed-cfar}
\end{equation}
This is a windowed CFAR detector: the threshold adapts to the local 
noise level within each window, as estimated from the window-weighted photon count $|\events|_{w,k}$, ensuring a consistent false alarm rate across all windows regardless of local photon density.

\paragraph{Weighted overlap-add}
To reconstruct a continuous-time flux estimate from overlapping windowed
measurements, we use a weighted overlap-add (WOLA)
procedure~\cite{crochiere_weighted_1980}. Let $\hat{\flux}_k(\spacetime)$
denote the local flux reconstruction within window $k$, obtained by
summing the detected Fourier coefficients. The
spatiotemporal flux $\hat{\flux}(\spacetime)$ is then reconstructed by the window-weighted sum
\begin{equation}
    \hat{\flux}(\spacetime) =
    \frac{\sum_k w(\spacetime - \spacetime_k)\,
    \hat{\flux}_k(\spacetime)}
    {\sum_k w(\spacetime - \spacetime_k)^2},
\end{equation}
where the denominator normalizes for the fact that the window is applied
twice: once during analysis (via the windowed probing function) and once
during synthesis (via the numerator weighting). Defining $w^2$ as the Hann window with 75\% overlap, the squared window sum
$\sum_k w(\spacetime-\spacetime_k)^2$ is approximately constant over
the interior of $\setspacetime$~\cite{harris_use_1978}, so the
reconstruction simplifies to a uniform weighted combination of
overlapping windows.

\subsection{Velocity-selective videography}
In this section, we describe the implementation details of velocity detection and velocity-selective videography, corresponding to Sections 3.2 and 4 of the main paper, respectively.

\subsubsection{Velocity detection hyperparameters}
Our velocity detection algorithm has the following hyperparameters:
\begin{enumerate}
    \item velocity plane thickness $\epsilon$, fixed at 0.5 unless specified differently;
    \item velocity search range $(\velocity_{\min}, \velocity_{\max}, \num{N}_v)$;
    \item neighborhood window $\window$;
    \item neighborhood guard $g$, fixed at 3 unless specified differently;
    \item CFAR rate $\cfarsigvalvel$ fixed at 0.001 unless differently specified.
\end{enumerate}

The plane thickness $\epsilon$ determines how much spectral mass is counted as supporting a given motion plane. Larger values improve robustness to discretization, noise, and slight model mismatch, but also reduce selectivity by allowing more off-plane coefficients to contribute. Smaller values yield sharper energy concentration but can become brittle when the spectral support is broadened by finite sampling, noise, or deviations from constant-velocity motion.

The number of sampled velocities per axis, \(\num{N}_v\), controls the discretization of the search space: a finer grid improves velocity resolution but increases the cost of plane aggregation quadratically. 

The effects of neighborhood geometry and CFAR rate $\cfarsigvalvel$ can be summarized as contributing to the control of overall false detection rate.

\subsubsection{Velocity detection with single-photon sensors}

Given a candidate velocity $\mathbf{v}=[v_x,v_y]^\top$, we measure how much
spectral energy is consistent with constant motion at that velocity,
and detect significant velocities using a nonparametric CFAR test.
Starting from the power spectrum
$\mathcal{E}^2(\freqx,\freqy,\freqt)$ of the input volume, detection
proceeds in three steps:
\begin{enumerate}
    \item For each candidate velocity, compute the signed distance of
    every frequency-domain point to the corresponding velocity plane,
    \begin{equation}
        \xi(\freqx,\freqy,\freqt; \mathbf{v}) =
        \freqt + v_x\,\freqx + v_y\,\freqy,
    \end{equation}
    and sum the power within thickness $\epsilon$ of the plane to obtain
    the energy map,
    \begin{equation}
        E(\mathbf{v}) = \sum_{(\freqx,\freqy,\freqt)}
        \mathcal{E}^2(\freqx,\freqy,\freqt)\,
        \mathbf{1}\!\left[\,|\xi(\freqx,\freqy,\freqt;\mathbf{v})| \le \epsilon \,\right].
    \end{equation}
    Velocities are swept over a uniform grid of $\num{N}_v$ values per
    axis, normalized as described in
    Section \ref{subsubsec:velocity-selective-video} to match the discrete
    Fourier grid.
    \item For each cell $i$ in the energy map, build a training set
    $\mathcal{T}_i$ from the neighborhood of half-width $\window$
    around $i$, excluding an inner guard region of half-width
    $g<\window$.
    \item Compute the rank statistic, the number of training cells
    exceeded by the CUT,
    \begin{equation}
        r_{i} = \sum_{e\,\in\,\mathcal{T}_{i}} \mathbf{1}[E_i \ge e].
    \end{equation}
    Under $H_0$, $r_i$ is uniform on $\{0,\dots,|\mathcal{T}_i|\}$, as proved in
    Equation 8. A detection is declared at $i$ when
    \begin{equation}
        r_i \ge \left\lceil (1-\alpha_{\mathrm{vel}})(|\mathcal{T}_i|+1)\right\rceil.
    \end{equation}
\end{enumerate}
For a spatiotemporal crop of size $\width \times \height \times \frames$ and a velocity grid of size $\num{N}_v \times \num{N}_v$,
computing the energy map requires comparing every spectral sample
against every candidate plane, giving
$O(\frames\,\height\,\width\,\num{N}_v^2)$ time. The subsequent CFAR
test operates only on the $\num{N}_v \times \num{N}_v$ energy map,
adding $O(\num{N}_v^2\,\window^2)$ time using a sliding-window
implementation.

\subsubsection{Velocity-selective videography with single-photon sensors}
 \label{subsubsec:velocity-selective-video}
After selecting a velocity $\velocityVec = [\velocity_x, \velocity_y]^\top$,
we form a velocity-selective video by compensating the corresponding
motion in the Fourier domain. Under constant velocity, the 3D spectrum
of the scene is supported on a plane through the origin (Equation 6 in the main paper). The goal is to
de-shear this plane so that it aligns with $\freqt = 0$, effectively
freezing the selected motion. 

The compensation is a three-step procedure. Starting from the 3D
spectrum $\fftflux(\freqx, \freqy, \freqt)$ of the input volume:
\begin{enumerate}
    \item Apply an inverse FFT along the temporal axis, producing a
    mixed representation in $[\freqx, \freqy, \timesym]^\top$.
    \item Multiply by the phase ramp
\begin{equation}
e^{\,i\,2\pi\,\timesym\left(
\velocity_x^{\mathrm{sc}}\,\freqx
+\velocity_y^{\mathrm{sc}}\,\freqy
\right)}
\end{equation}
    which shifts each spatial frequency by a temporal offset
    proportional to the velocity.
    \item Apply a forward FFT along time to obtain the de-sheared
    spectrum, from which the compensated video is reconstructed.
\end{enumerate}

Reconstruction is then performed in a sliding temporal window over the photon stream to generate video frames. Although related ideas have been explored in conventional videography~\cite{Vernon2001FourierVS} and single-photon imaging~\cite{sundar2023sodacam}, our framework performs velocity focusing at ultra-low light levels and without requiring manual specification of the motion direction or magnitude as in previous work~\cite{sundar2023sodacam}.

The scaled velocities $\velocity_x^{\mathrm{sc}}$ and
$\velocity_y^{\mathrm{sc}}$ account for the fact that the candidate
velocities are specified in pixels per frame, while the phase ramp
operates on the discrete Fourier grid. The rescaling matches the two
coordinate systems:
\begin{equation}
\velocity_x^{\mathrm{sc}}
= \velocity_x
\frac{\max(\freqt)-\min(\freqt)}{\max(\freqx)-\min(\freqx)},
\qquad
\velocity_y^{\mathrm{sc}}
= \velocity_y
\frac{\max(\freqt)-\min(\freqt)}{\max(\freqy)-\min(\freqy)}.
\end{equation}

For a spatiotemporal crop of size
$\width \times \height \times \frames$, each compensated video requires
one inverse and one forward FFT along time plus one element-wise
multiplication, giving $O(\frames\,\height\,\width\log\frames)$ time
and $O(\frames\,\height\,\width)$ memory. All operations are dense
tensor computations and run efficiently on a GPU.

\subsection{Spike camera videography}

Spike cameras produce spikes at regularly sampled frames at 40k fps~\cite{huang_1000_2023}, so we take the STFT approach from single-photon videography, instead using the conservative CFAR bound defined in~\cref{eq:spike-cfar_practical}.

To suppress harmonic artificial frequencies produced from constant flux spike trains, we apply a low-pass filter to the probing measurements by dropping frequencies $\freqvec$ with large temporal frequencies $\freqt$.

\subsection{Event camera videography}

We discuss implementation details of reconstructing log-flux from event cameras, both from pure integration using only events, and a basic fusion method using a Kalman filter~\cite{kalman_new_1960}.

\subsubsection{Log-flux estimation from only events}

Using the spatiotemporal event probing measurements described in~\cref{prop:event_probing_st}, we apply CFAR detection using the conservative bound in~\cref{eq:event-cfar-overest} with $\alpha=10^{-4}$. This yields a noisy estimate of $\partial_t\log\flux(\spacexy, t)$, which we can then integrate over $\exposure$ for an estimate of $\log\flux(\spacexy, t)$ up to a constant.

Since our reconstructed function takes the form of

\begin{equation}
    \widehat{\partial_t\log\flux(\spacexy, t)} = C\sum_{\freqvec\in\freqscfar}\probemeas_\freqvec e^{j2\pi\freqvec^\top[\spacexy, t]},
\end{equation}

integration over time can be performed by dividing the spectrum by $j2\pi\freqt$ for $\freqt\neq0$:

\begin{equation}
    \widehat{\log\flux(\spacexy, t)} = C\sum_{\freqvec\in\freqscfar,\freqt\neq0}\frac{1}{j2\pi\freqt}\probemeas_\freqvec e^{j2\pi\freqvec^\top[\spacexy, t]}.
\end{equation}

We discard all temporal DC components ($\freqt = 0$), which are generally dominated by bias and drift~\cite{scheerlinck_continuous-time_2018}, and apply a high-pass filter in $\freqt$ to suppress additional low-frequency artifacts in the reconstructed scene.

\subsubsection{Kalman filter with intensity frames}

The event data in~\cite{mueggler_event-camera_2017} was recorded with a DAVIS240C camera~\cite{ini_davis240c}, which also records intensity images at approximately 20 fps. This can be combined with the estimated log-flux derivative $\widehat{\partial_t\log\flux(\spacexy, t)}$ via the Kalman filter~\cite{kalman_new_1960}.

We model the log-flux $x(\spacexy, t) = \log \flux(\spacexy, t)$ as a scalar state per pixel. The process model integrates the event-derived derivative estimate as a control input over a timestep $\Delta t$:
\begin{equation}
    x(\spacexy, t) = x(\spacexy, t - \Delta t) + \Delta t \cdot \widehat{\partial_t \log \flux(\spacexy, t)} + w, \qquad w \sim \mathcal{N}(0, Q),
\end{equation}
and the intensity measurement model treats each APS frame as a direct noisy observation of the log-flux:
\begin{equation}
    z(\spacexy) = \log I(\spacexy) = x(\spacexy) + v, \qquad v \sim \mathcal{N}(0, R).
\end{equation}
The same filter applies identically in the Fourier domain: replacing $x(\spacexy, t)$ with $\hat{x}(\freq, t)$ and $z(\spacexy)$ with $\hat{z}(\freq)$, the process and measurement models retain their linear-Gaussian structure, since the Fourier transform is a linear operation.
In our experiments, $Q$ and $R$ were set empirically without formal
tuning. A principled calibration of these noise parameters is left to
future work.

\clearpage
\section{Experiments}
We use the SPAD512 camera from Pi Imaging to capture the new datasets introduced with this work. This camera produces binary frames, which are single-bit images recording a 1 at a pixel location if it receives one or more photons during a frame exposure. Unless otherwise stated, each binary frame has a temporal resolution (exposure) of 10 microseconds, and frames are taken at a regular rate of 97.1 kfps for a total of 100000 frames. Scene illumination is controlled to ensure that photon detections are unlikely to be affected by SPAD dead time. In other words, 1s in the binary frames are unlikely to correspond to more than a single photon arrival. Under this operating condition, treating detections as individual photon arrivals is effectively equivalent to sampling from the underlying 3D inhomogeneous Poisson process governed by the photon arrival rate at the sensor. To further ensure this regime, we additionally employ ND filters and adjust the aperture of the focusing lens when necessary.

\paragraph{Scene specific details}
\begin{itemize}
    \item \textbf{Balloon Popping with 31~kHz LED (Fig.~1).} The scene contains two sources of illumination flicker: ceiling lights powered by the AC mains at approximately 120~Hz, and an additional light source flickering at 31~kHz. An ND filter with optical density OD=0.7 was used for this capture. The average detection rate was 2458 photons per pixel per second (ppps), varying from 14625 ppps in bright regions to 34 ppps in dark regions.
    
    \item \textbf{Blender (Fig.~4).} The illumination flicker in this scene is also caused by AC-powered lighting, resulting in a natural flicker at approximately 120~Hz. An ND filter with optical density $\mathrm{OD}=0.7$ is used for this scene. The average detection rate was 5977 photons per pixel per second (ppps), varying from 98885 ppps in bright regions to 75 ppps in dark regions.

    \item \textbf{Static Balloon with high-speed light (Fig.~4).} The 31~kHz flickering light source is generated using an LED–resistor circuit driven by a function generator with a sinusoidally varying voltage. Ambient lighting with 120~Hz AC flicker is also present in the scene. An ND filter with optical density $\mathrm{OD}=0.7$ is used. The average detection rate was 5933 photons per pixel per second (ppps), varying from 14937 ppps in bright regions to 496 ppps in dark regions.

    \item \textbf{Dropping Ball (Fig.~4).} A static ball is released from a fixed height and falls with minimal rotational motion. An ND filter with optical density $\mathrm{OD}=0.3$ is used for this scene. The average detection rate was 3278 photons per pixel per second (ppps), varying from 18105 ppps in bright regions to 45 ppps in dark regions.

    \item \textbf{Balloon Popping with 20~kHz LED (supplemental webpage).} We captured an additional scene similar to Fig.~1, in which a 20~kHz flickering LED is directed at the balloon to make the flicker more visible. The background illumination flicker in this scene originates from ceiling lights powered by the AC mains, producing a flicker at approximately 120~Hz. An ND filter with optical density OD=0.7 was used for this capture. The average detection rate was 3653 photons per pixel per second (ppps), varying from 33937 ppps in bright regions to 59 ppps in dark regions.

\end{itemize}

\paragraph{Thinning while maintaining dark counts}
To simulate a lower photon flux, we perform thinning of the photon detections by independently applying Bernoulli sampling to each detected event, retaining it with probability $p$. It is well known that thinning an inhomogeneous Poisson process in this manner produces another inhomogeneous Poisson process whose rate is scaled by $p$ relative to the original process~\cite{snyder2012random}. Note that uniformly thinning all timestamps by a constant factor does not change the SNR, since both
signal photons and dark counts are reduced proportionally. To vary the SNR, we thin the original set of captured timestamps
while maintaining the same level of dark counts.

We estimate the dark count rate of the SPAD512 single-photon camera by capturing binary frames with the lens cap on. After
removing hot pixels following~\cite{liu2024bit2bit}, we compute the average dark count rate as the total number of detected photons divided
by the total acquisition time and number of pixels, yielding 25 counts per second (cps) for our camera. We then thin the original timestamps by a desired factor $p$ and add timestamps from a homogeneous Poisson process
with rate $(1-p)\times 25$ cps. This reduces the signal by approximately factor $p$ while maintaining a constant dark count rate of 25 cps, effectively lowering the SNR.

\paragraph{Hot pixel correction}
Hot pixel correction is performed through device calibration and then using statistical off-board correction as implemented in the \texttt{spadtools}~\cite{spadtools} library.

\subsection{Single-photon videography}

For all scenes captured with the SPAD512 single-photon camera (binary frame quanta sensor), we use
the STFT with $100 \times 100 \times 10{,}000$ windows, Hann windowing,
and 75\% overlap, as described in Section~\ref{sec:stft}. Within each window,
we probe spatial frequencies $\freqx$ and $\freqy$ up to 256 cycles and
temporal frequency $\freqt$ up to 50\,kHz. We set $\cfarsigval=10^{-4}$
and reconstruct the spatiotemporal flux from the detected frequencies.
Since the reconstructed flux is a continuous function, it can be sampled
at any desired frame rate up to the Nyquist limit set by the sensor's
timestamp resolution.
We perform gamma correction to all videos with $\gamma=2.2$.
The supplemental webpage shows videos rendered at various frame rates for the scenes in Figure~1 and Figure~4 (top) of the main paper.
\begin{remark}
Frequency detection is essential to the reconstruction. Applying the
Fourier transform and inverting it without frequency selection
effectively reproduces the original binary sequence, placing
approximately a Dirac impulse at every photon location in space and
time.
\end{remark}

\subsubsection{Additional details for Balloon popping with 31~kHz LED (Figure 1 scene)}

\begin{figure}[t]
  \centering
  \includegraphics[width=0.9\textwidth]{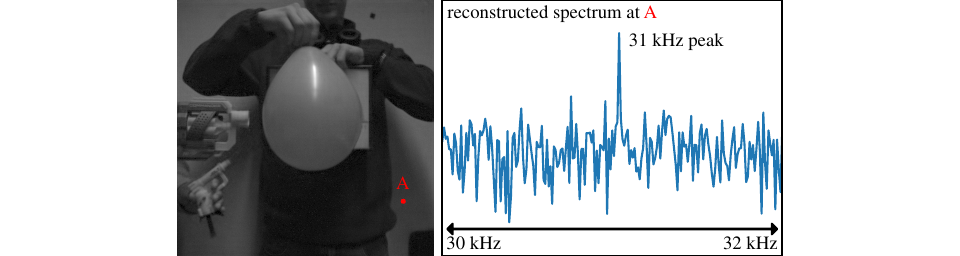}
  \caption{\textbf{Recovered 31 kHz temporal flicker in Figure 1.} Our method successfully recovers the faint 31~kHz LED flicker, alongside the 120~Hz ceiling light flicker and the scene dynamics shown in Figure 1.}
  \label{fig:teaser-fft}
\end{figure}

 Our method is able to recover the 31~kHz modulated LED in Figure 1 simultaneously with the projectiles, 120 Hz ceiling light, and balloon rupture. We show the recovered temporal spectrum at a pixel in~\cref{fig:teaser-fft}.

\subsubsection{Additional details for ultra-wideband videography (Figure 4, row 5 scene)}

Following UWB~\cite{wei2023passive}, we probe temporal frequencies
corresponding to the known harmonics of the fan (108~Hz) and the MHz
laser (20~MHz), since an exhaustive 3D Fourier transform up to GHz
temporal frequencies is computationally infeasible. For each selected
$\freqt$, we probe all spatial frequencies and apply CFAR detection to
reject outlier spatial frequencies.

\subsubsection{Runtime for single-photon videography}

\textbf{Computational complexity.} On quanta data, the binary frames form a regularly sampled volume of $\width \times \height$ pixels over $T$ time bins spanning a physical exposure $\exposure$, giving $N = \width \height T$ total samples. Global 3D FFT+CFAR+IFFT costs $O(N \log N)$. With the STFT, let $K$ denote the number of spatiotemporal windows, each of size $N_k = \width_k \height_k T_k$ samples; the total cost is $O(K N_k \log N_k)$, where $K$ grows with overlap and shrinking window size, which incurs higher cost than the global FFT.

\textbf{Runtime.} All reconstructions run on a single NVIDIA V100 GPU (32\,GB VRAM) and
require no training. For SPAD512 scenes, reconstructing $10{,}000$
consecutive frames takes under 15 minutes, while reconstructing
$1{,}000$ frames sparsely sampled across all $100{,}000$ quanta frames
takes approximately one hour. For reference, bit2bit requires
approximately 8 hours of training on $40{,}000$ binary frames before
inference can begin.

\subsection{Velocity-selective videography on the Figure 1 scene}
\label{sec:exp:velocity-sel}

We demonstrate velocity-selective videography on this real scene
containing motions of varying speed, contrast, and spatial extent. The
velocity search grid spans $300 \times 300$ hypotheses (half-width 150
in each direction), yielding $90{,}000$ candidate velocities. The
neighborhood window is $\window = 150$ and the significance level is
$\cfarsigvalvel = 10^{-4}$. For each event of interest, we select the
velocity range and reference frame to match the speed and timing of the
target motion.

\begin{figure}[t]
    \centering
    \includegraphics[width=\linewidth]{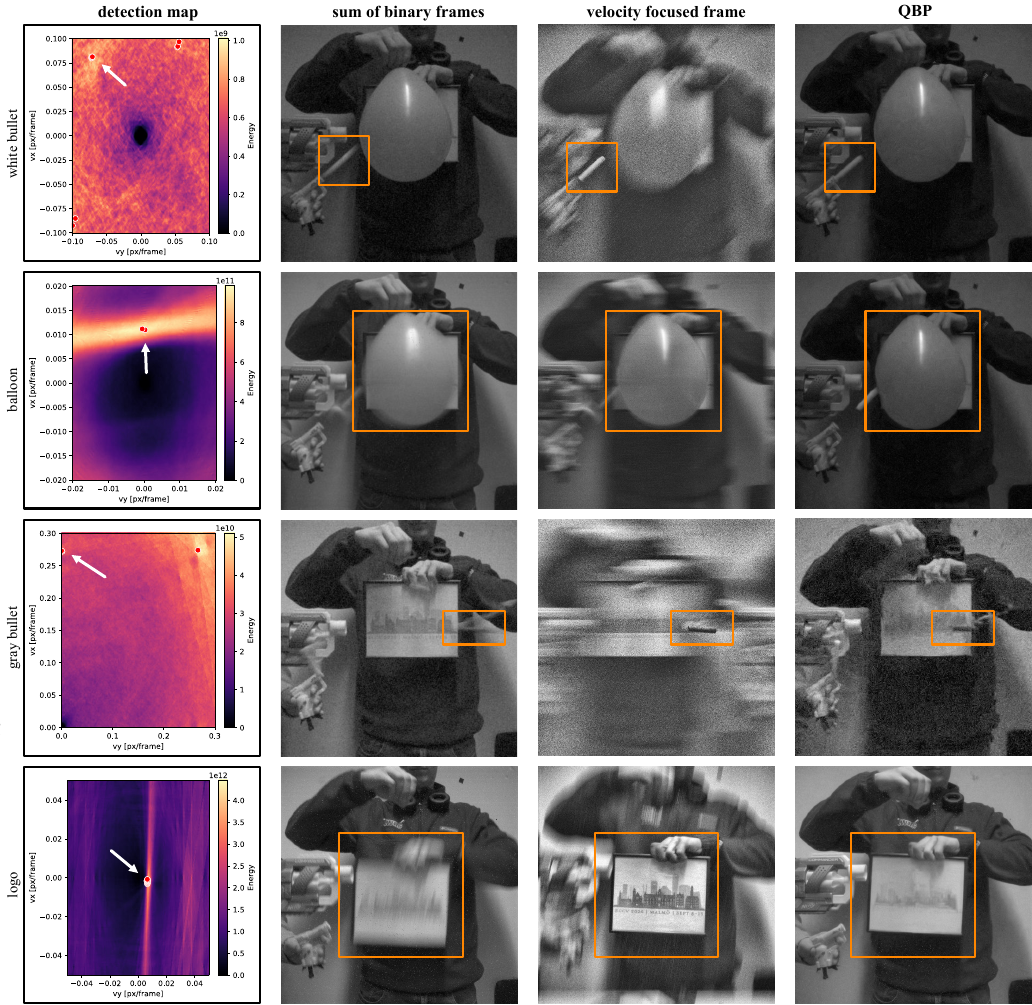}
    \caption{\textbf{ Velocity-selective videography on the Fig.~1 scene.} We apply velocity detection to four moving objects: the white bullet,
balloon, gray bullet, and ECCV logo (rows top to bottom). Each row
shows: the velocity-space energy map with detected velocities (red circles); the sum of binary frames over
the selected temporal interval; the velocity-selective reconstruction
focused on the detected velocity (annotated with the white arrow); and the QBP reconstruction over the
same interval. Orange boxes highlight the target object. The
velocity-selective reconstruction isolates each moving object from the
scene, whereas the summed frames and QBP mix all motions together.
Full video sequences are available on the supplemental webpage.}
    \label{fig:teaser_experiments}
\end{figure}

Figure~\ref{fig:teaser_experiments} shows four representative motions
from the same scene. From top to bottom, the targets are: the white
bullet, the balloon, the gray bullet, and the ECCV logo. Each row
contains four panels: (i) the velocity-space detection map with
detected velocities marked, (ii) the sum of binary frames over the
selected temporal interval, (iii) the velocity-selective reconstruction
for the detected motion, and (iv) the QBP reconstruction over the same
interval. Orange boxes highlight the region corresponding to the target
motion. Full video sequences are available on the supplemental webpage.

\paragraph{Velocity detection} We report the velocity detection results for each target motion below:
\begin{itemize}
    \item \textbf{White bullet:} 9 detections, 4 inliers.
    Detected velocity: $(0.076, -0.073)$\,px/frame.

    \item \textbf{Balloon:} 3 detections, 2 inliers.
    Detected velocity: $(0.011, -0.001)$\,px/frame.
    The balloon undergoes nonlinear motion (when hit by the white bullet),
    but at the frame rate of the SPAD512 ($\sim$100\,kfps) its
    trajectory is well approximated as locally linear. The successful
    detection suggests that our velocity detection method can extend
    beyond strictly constant-velocity scenes.

    \item \textbf{Gray bullet:} 3 detections, 2 inliers.
    Detected velocity: $(0.272,0)$\,px/frame.

    \item \textbf{ECCV logo:} 8 detections, all inliers.
    Detected velocity: $(-0.001,-0.007)$\,px/frame.
\end{itemize}

\paragraph{Observations}
Overall, the velocity detector recovers accurate motion hypotheses across a range of velocities and object sizes, and the resulting reconstructions successfully isolate the target motion from the rest of the scene. Interestingly, the
velocity-selective reconstruction isolates the target motion even when
the summed binary frames contain substantial overlap from other scene
content. This is most evident for the white (Fig.~\ref{fig:teaser_experiments}, row 1) and gray bullets (Fig.~\ref{fig:teaser_experiments}, row 3), where
the moving object is difficult to distinguish in the summed frames but
becomes clearly separated after velocity focusing. The balloon (Fig.~\ref{fig:teaser_experiments}, row 2)
demonstrates that the method also applies to larger, slower motions,
whose broader velocity-space signatures yield smoother reconstructions.
The ECCV logo (Fig.~\ref{fig:teaser_experiments}, row 4) produces a particularly coherent detection, with all
hypotheses classified as inliers, indicating a strong and well-isolated
motion signature. 
Compared with QBP~\cite{ma2020quanta}, our method selectively reconstructs the velocity of interest while suppressing other scene content, producing sharper images of the target object.

\subsection{Videography with other asynchronous sensors (Figure 5)}

We apply event probing to the datasets \texttt{shapes\_6dof} (top row, left) and \texttt{office\_spiral} (top row, right) from~\cite{mueggler_event-camera_2017}. We note that choosing to apply the Kalman filter in image spatial or Fourier domain affects the quality of the reconstruction depending on the scene. As a result, we apply Kalman filtering in the spatial domain for \texttt{shapes\_6dof}, and in the Fourier domain in \texttt{office\_spiral}.

We reconstruct the following spike camera videos from~\cite{zhu_retina-like_2020}: \texttt{balloon}, \texttt{rotation2x}, and \texttt{car-100kmh}. As we discuss in the main paper, pixels with constant flux produce nearly periodic spike trains, which
introduce spurious harmonic peaks at high frequencies. To suppress
these artifacts, we apply a temporal low-pass filter whose cutoff
depends on scene brightness: in darker regions, the inter-spike
interval is longer and the spurious harmonics appear at lower
frequencies, requiring a more aggressive cutoff.

\clearpage
\section{Simulations}

\subsection{Probing measurement normality}
\label{sec:probe-normality}

\subsubsection{Central limit theorem validity}

The probing measurements $\probemeas_\probe$ tend towards a normal 
distribution as the number of observed photons increases 
(\cref{prop:probe-dist}), with the rate of convergence quantified via 
the Berry--Esseen theorem~\cite{berry_accuracy_1941, 
esseen_liapounoff_1942}, which bounds the worst-case deviation between 
the CDF of a standardized sum of $n$ i.i.d.\ random variables and the 
standard normal CDF by $C\rho/(\sigma^3\sqrt{n})$, where $\sigma^2$ is 
the variance, $\rho = \mathbb{E}[|X - \mu|^3]$ is the third absolute central moment, 
and $C > 0$ is a universal 
constant. The ratio $\rho/\sigma^3$ 
characterizes how many samples are needed for a given accuracy, with 
heavier-tailed distributions requiring more.

For quanta image sensors in the Poisson regime, the CLT sample size is 
the number of photons contributing to a spatiotemporal probe. For a 
typical STFT window of $100\times100\times10{,}000$ space-time bins at 
$10^{-4}$~ppp, this gives ${\sim}10^4$ photons per probe, which as we 
show empirically in Section~\ref{sec:probe-normality-sim} is well above 
the threshold for normality to hold.

In the high-$q$ Bernoulli regime, each space-time bin produces a 
Bernoulli random variable with success probability $q$. The Berry--Esseen 
factor for a Bernoulli$(q)$ variable is
\begin{equation}
    \frac{\rho}{\sigma^3} 
    = \frac{(1-q)^3\cdot q+q^3\cdot(1-q)}{(q(1-q))^{3/2}} 
    = \frac{q^2 + (1-q)^2}{\sqrt{q(1-q)}},
\end{equation}
which grows rapidly as $q \to 1$: at $q = 0.999$, this factor is 
${\sim}32\times$ larger than at $q = 0.5$, meaning convergence to 
normality requires ${\sim}10^3\times$ more samples for comparable 
accuracy. However, for a probe with $n = 10^8$ space-time bins at $q = 0.999$, 
the expected number of nondetections is $(1-q)n \approx 10^5$, which 
are the primary drivers of variance in the probing sum. Despite the 
${\sim}10^3\times$ higher sample requirement relative to $q = 0.5$, 
the large probe support ensures the effective number of nondetections 
remains well into the regime where the CLT approximation holds.

\subsubsection{Empirical validation}
\label{sec:probe-normality-sim}

\begin{figure}[t]
    \centering
    \includegraphics[width=\linewidth]{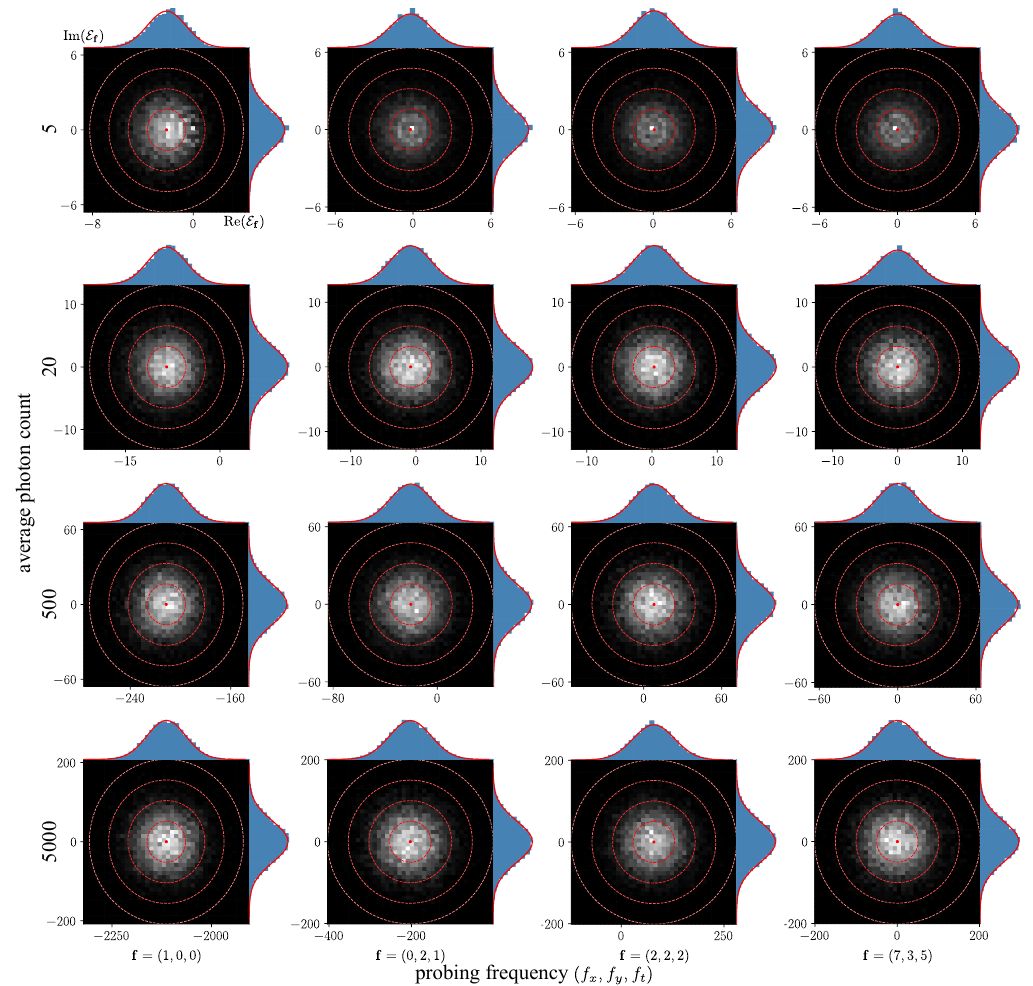}
    \caption{\textbf{Distribution of Fourier probing measurements 
at varying light levels.} Each cell shows the empirical joint 
distribution of the real and imaginary parts of $\probemeas_\freqvec$ 
over 10,000 trials, with marginal histograms (blue) and ground-truth 
normal curves (red) shown along the top and right axes. Ground-truth 
Gaussian contours (red), spaced one standard deviation apart, are 
overlaid on each 2D histogram. Columns correspond to selected probe 
frequencies assuming the domain is a unit spatiotemporal cube; rows correspond to expected photon counts of 5, 20, 500, 
and 5,000. The approximation is already accurate at 20 photons, well 
below the photon counts observed in our experiments.}
    \label{fig:probe-normality}
\end{figure}

We validate the normality approximation empirically through simulations 
in Fig.~\ref{fig:probe-normality}, where we probe photons realized from 
the same Gaussian blob flux function used in Fig.~\ref{fig:1duwb-fail} 
via the Fourier probing basis. The flux is scaled to different 
amplitudes to simulate different light levels, such that the expected 
number of photons for each simulation trial is 5, 20, 500, and 5,000 
respectively. For each light level, 10,000 trials are run. For each 
trial, we compute the probing measurements from the realized photons at 
selected frequencies and estimate the distribution of each 
$\probemeas_\freqvec$.

The empirical probing measurement histograms in 
Fig.~\ref{fig:probe-normality} are overlaid with ground-truth Gaussian 
contours computed from Equation~\ref{eq:probe-dist-fourier}, centered at the 
ground-truth mean, where each ring represents one standard deviation. 
For as few as 20 photons, both the joint and marginal distributions are 
already well approximated by a complex Gaussian. This is well within 
the photon counts observed in our experiments: even our darkest thinned 
datasets, with an STFT window of $100\times100\text{ pixels}\times 
10{,}000\text{ frames}$, operate at light levels significantly above 
this threshold. Furthermore, the contours are approximately circular 
across all selected frequencies, consistent with the diagonal covariance 
approximation of Equation~\ref{eq:probe-dist-fourier}; note however that 
this reflects the spectral properties of the Gaussian blob rather than 
natural video statistics, for which the diagonal approximation is 
instead justified by the power law decay of natural video power spectra 
(\cref{cor:cov-weaksig}).

\subsection{CFAR threshold validity}
\label{sec:cfar-robust}

\begin{figure}
    \centering
    \includegraphics[width=\linewidth]{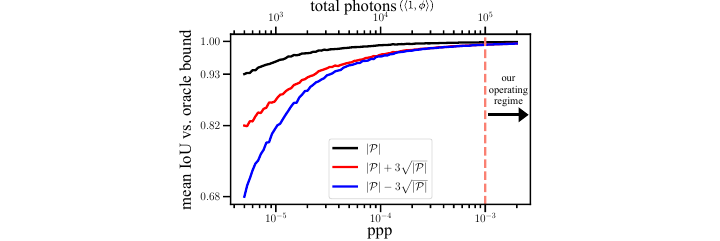}
    \caption{\textbf{CFAR threshold robustness.} Mean IoU of 
    recovered frequencies relative to the oracle CFAR threshold (derived 
    from the true flux integral $\langle 1, \flux \rangle$), evaluated 
    over 1,000 trials per light level. The observed-count threshold 
    $|\events|$ (black) and $\pm3\sigma$ bounds $|\events| \pm 
    3\sqrt{|\events|}$ (red/blue) are compared against the oracle. The 
    bottom axis shows photons-per-pixel (ppp) for an STFT window of 
    $100\times100\text{ pixels}\times10{,}000\text{ frames}$; the top 
    axis shows the corresponding total photon count. Our experimental 
    operating regime ($\geq 10^{-3}$ ppp, indicated by the arrow) lies 
    in the region where all three thresholds converge to near-perfect 
    agreement with the oracle.
    }
    \label{fig:cfar-stability}
\end{figure}

The CFAR detection threshold in Equation~\ref{eq:CFAR3d} uses the observed 
photon count $|\events|$ as a proxy for the unknown flux integral 
$\langle 1, \flux \rangle$. Since $|\events|$ is itself a Poisson 
random variable with mean $\langle 1, \flux \rangle$ and standard 
deviation $\sqrt{|\events|}$, the threshold is subject to shot noise. 
We evaluate the robustness of this approximation in 
Fig.~\ref{fig:cfar-stability} by comparing the set of frequencies 
recovered under the observed-count threshold against the oracle 
threshold derived from the true flux integral, measured via mean 
intersection-over-union (IoU) of the detected frequency masks. We also 
evaluate thresholds at $|\events| \pm 3\sqrt{|\events|}$, representing 
the extremes of a $3\sigma$ fluctuation in the observed count, to 
characterize the worst-case sensitivity of the detector to photon count 
noise.

At our experimental operating regime of $\geq 10^{-3}$ ppp 
($\geq 10^5$ photons), the observed-count threshold achieves a mean IoU 
close to 1 relative to the oracle bound, confirming that $|\events|$ is 
a reliable proxy for $\langle 1, \flux \rangle$ at practical flux 
levels. It also performs better on average compared to the extreme 
$\pm 3\sqrt{|\events|}$ bounds, suggesting that underestimating or overestimating the bound is not necessary.

\subsection{Quantitative reconstruction error}

\subsubsection{Simulating photon arrivals from high-speed videos}\label{sec:simDetails}

We simulate binary frames captured by a single photon camera from high-speed videos captured by conventional cameras using a procedure similar to \cite{sundar2024generalized}:

\begin{enumerate}
    \item We collect high-speed videos originally filmed with a Phantom camera \cite{World8KSlow}, at 1~kfps. The videos are downloaded at $1920\times1080$, and are downsampled to $512\times288$. The gamma-corrected videos are converted to linear intensity (exponent $2.2$).

    \item To increase temporal resolution, we first interpolate the videos by $16\times$ using RIFE \cite{huang2022RIFE}, yielding a nominal frame rate of 16~kfps. We then linearly interpolate 7 additional frames between consecutive frames, producing a final temporal resolution of approximately 128~kfps.

    \item Photon timestamps are generated at each pixel by thinning an inhomogeneous Poisson process whose rate is assumed proportional to the interpolated intensity signal. We treat the normalized intensity values in the range $[0,1]$ as representing an approximately constant photon arrival rate within each pixel. The resulting binary videos finally are quantized to 10-microsecond resolution.

\end{enumerate}

\subsubsection{Simulated evaluation}

Table~\ref{tab:quant-results} evaluates our video reconstruction method
across 6 simulated scenes at three photon levels, covering a broad
range of flux conditions. We compare against UWB~\cite{wei2023passive}, QNN~\cite{sundar2025quanta}
and bit2bit~\cite{liu2024bit2bit}, the current state-of-the-art in
photon-limited video reconstruction. QBP~\cite{ma2020quanta} is
computationally prohibitive for practical frame-by-frame video
synthesis\footnote{Using the implementation of~\cite{ma2020quanta},
processing the raw $130\text{k} \times 512 \times 512$ binary volume
and rendering $300$ output frames requires $2$--$3$ days on a single
CPU.}, so we omit it from the quantitative comparison. We do, however,
include QBP in the supplemental videos and in the velocity-selective
videography comparison (Section~\ref{sec:exp:velocity-sel}), given its
consistently strong reconstruction quality and widespread use as a
reference method~\cite{chennuri2024quanta}.

As expected, reconstruction quality degrades as the photon level
decreases for all methods. Our method achieves the highest PSNR on the
large majority of scene and light-level combinations. Perceptual metrics (LPIPS and SSIM) are more
competitive: our method leads on most scenes, but bit2bit achieves
stronger perceptual scores on scenes with smooth textures and liquid
motion (\eg, strawberrymilk, grasswaterfall), where its learned
denoiser can hallucinate plausible detail that improves perceptual
similarity without necessarily improving fidelity. Full video
reconstructions for all scenes and methods are available on the
supplemental webpage.

\begin{table*}[t]
\centering
\caption{Quantitative comparisons across six simulated scenes at three photons per pixel (ppp) levels. Higher PSNR, SSIM, and lower LPIPS indicate better image reconstruction quality. The best result among the four methods for each scene and ppp level is indicated in \textcolor{red}{red}.}
\label{tab:quant-results}
\resizebox{\textwidth}{!}{%
\begin{tabular}{lllcccccc}
\toprule
\multicolumn{3}{c}{} & \multicolumn{6}{c}{Scene} \\
\cmidrule(lr){4-9}
Metric & Method & ppp & coffeebeans & fireblast & grasswaterfall & shoewaterstep & strawberrymilk & wineglassfall \\
\midrule
\multirow{12}{*}{PSNR $\uparrow$} & \multirow{3}{*}{proposed} & 0.1 &  \textcolor{red}{29.18} & 33.21 &  \textcolor{red}{26.53} &  \textcolor{red}{31.26} &  \textcolor{red}{30.75} &  \textcolor{red}{33.14} \\
 &  & 0.01 &  \textcolor{red}{27.98} & 31.74 & 26.71 &  \textcolor{red}{27.97} &  \textcolor{red}{28.45} &  \textcolor{red}{28.74} \\
 &  & 0.002 &  \textcolor{red}{25.38} & 28.52 &  \textcolor{red}{24.77} &  \textcolor{red}{22.19} &  \textcolor{red}{20.08} &  \textcolor{red}{23.83} \\[2pt]
\cmidrule(lr){3-9}
 & \multirow{3}{*}{bit2bit} & 0.1 & 17.89 &  \textcolor{red}{44.77} & 24.11 & 17.73 & 14.28 & 17.06 \\
 &  & 0.01 & 20.89 & 32.67 &  \textcolor{red}{26.78} & 24.43 & 15.67 & 19.29 \\
 &  & 0.002 & 19.30 & 29.82 & 23.36 & 21.11 & 15.62 & 21.46 \\[2pt]
\cmidrule(lr){3-9}
 & \multirow{3}{*}{UWB} & 0.1 & 25.70 & 28.90 & 24.82 & 27.20 & 23.49 & 28.58 \\
 &  & 0.01 & 20.58 & 28.83 & 22.42 & 16.74 & 14.18 & 21.13 \\
 &  & 0.002 & 18.17 & 30.26 & 18.77 & 12.68 & 10.76 & 16.53 \\[2pt]
\cmidrule(lr){3-9}
 & \multirow{3}{*}{QNN} & 0.1 & 19.66 & 31.69 & 20.67 & 24.06 & 22.11 & 26.13 \\
 &  & 0.01 & 19.32 &  \textcolor{red}{37.14} & 24.95 & 13.09 & 9.88 & 15.01 \\
 &  & 0.002 & 16.81 &  \textcolor{red}{35.69} & 23.91 & 11.08 & 8.44 & 13.30 \\
\addlinespace
\midrule
\multirow{12}{*}{LPIPS $\downarrow$} & \multirow{3}{*}{proposed} & 0.1 &  \textcolor{red}{0.082} &  \textcolor{red}{0.063} &  \textcolor{red}{0.157} & 0.217 & 0.450 &  \textcolor{red}{0.083} \\
 &  & 0.01 &  \textcolor{red}{0.151} & 0.080 &  \textcolor{red}{0.263} & 0.412 & 0.629 &  \textcolor{red}{0.328} \\
 &  & 0.002 &  \textcolor{red}{0.344} & 0.153 & 0.456 & 0.635 & 0.883 &  \textcolor{red}{0.625} \\[2pt]
\cmidrule(lr){3-9}
 & \multirow{3}{*}{bit2bit} & 0.1 & 0.328 & 0.139 & 0.287 & 0.248 & 0.397 & 0.302 \\
 &  & 0.01 & 0.276 & 0.097 & 0.279 &  \textcolor{red}{0.387} &  \textcolor{red}{0.479} & 0.551 \\
 &  & 0.002 & 0.440 & 0.138 &  \textcolor{red}{0.434} &  \textcolor{red}{0.493} &  \textcolor{red}{0.624} & 0.674 \\[2pt]
\cmidrule(lr){3-9}
 & \multirow{3}{*}{UWB} & 0.1 & 0.315 & 0.173 & 0.490 & 0.365 & 0.750 & 0.257 \\
 &  & 0.01 & 0.570 & 0.278 & 0.855 & 0.708 & 1.088 & 0.533 \\
 &  & 0.002 & 0.696 & 0.338 & 1.059 & 0.812 & 1.221 & 0.818 \\[2pt]
\cmidrule(lr){3-9}
 & \multirow{3}{*}{QNN} & 0.1 & 0.192 & 0.073 & 0.195 &  \textcolor{red}{0.190} &  \textcolor{red}{0.340} & 0.176 \\
 &  & 0.01 & 0.271 &  \textcolor{red}{0.056} & 0.331 & 0.478 & 0.677 & 0.471 \\
 &  & 0.002 & 0.468 &  \textcolor{red}{0.095} & 0.519 & 0.629 & 0.827 & 0.727 \\
\addlinespace
\midrule
\multirow{12}{*}{SSIM $\uparrow$} & \multirow{3}{*}{proposed} & 0.1 &  \textcolor{red}{0.913} &  \textcolor{red}{0.909} &  \textcolor{red}{0.812} &  \textcolor{red}{0.887} & \textcolor{red}{0.782} &  \textcolor{red}{0.930} \\
 &  & 0.01 &  \textcolor{red}{0.843} & 0.841 &  \textcolor{red}{0.777} & 0.781 & 0.653 &  \textcolor{red}{0.746} \\
 &  & 0.002 &  \textcolor{red}{0.624} & 0.493 &  \textcolor{red}{0.625} & 0.556 & 0.362 &  \textcolor{red}{0.442} \\[2pt]
\cmidrule(lr){3-9}
 & \multirow{3}{*}{bit2bit} & 0.1 & 0.545 & 0.900 & 0.441 & 0.759 & 0.670 & 0.545 \\
 &  & 0.01 & 0.739 & 0.822 & 0.699 &  \textcolor{red}{0.801} &  \textcolor{red}{0.716} & 0.391 \\
 &  & 0.002 & 0.541 & 0.558 & 0.513 &  \textcolor{red}{0.688} &  \textcolor{red}{0.603} & 0.350 \\[2pt]
\cmidrule(lr){3-9}
 & \multirow{3}{*}{UWB} & 0.1 & 0.740 & 0.810 & 0.561 & 0.745 & 0.458 & 0.822 \\
 &  & 0.01 & 0.523 & 0.713 & 0.289 & 0.442 & 0.154 & 0.553 \\
 &  & 0.002 & 0.402 & 0.664 & 0.113 & 0.351 & 0.084 & 0.321 \\[2pt]
\cmidrule(lr){3-9}
 & \multirow{3}{*}{QNN} & 0.1 & 0.749 & 0.884 & 0.436 & 0.859 &  0.767 & 0.828 \\
 &  & 0.01 & 0.744 &  \textcolor{red}{0.911} & 0.633 & 0.510 & 0.352 & 0.356 \\
 &  & 0.002 & 0.573 &  \textcolor{red}{0.843} & 0.539 & 0.284 & 0.168 & 0.186 \\
\bottomrule
\end{tabular}
}
\end{table*}

\subsection{Velocity detection}

This section evaluates the accuracy and robustness of our velocity
detection method (Section~3.2 of the main paper and
Section~\ref{sec:vel-det}) on controlled synthetic benchmarks where
ground-truth motion is known exactly. We study four aspects of detector
behavior:
\begin{enumerate}
    \item \textbf{Best-detection accuracy:} how closely the top
    detected velocity matches the true velocity for a single moving
    Gaussian blob.
    \item \textbf{Detection-set quality:} the precision of the detector.
    \item \textbf{Velocity limits:} detector behavior as the true
    speed approaches the lower and upper bounds of the discretized
    search range.
    \item \textbf{Multiple objects:} a multi-object extension with
    several independently moving blobs, comparing global scene-level
    detection against local object-centered detection.
\end{enumerate}
Together, these experiments characterize how the detector responds to
changes in photon rate, motion speed, and scene complexity.

\subsubsection{Accuracy of the best detection}

\paragraph{Scene and motion model} We evaluate the accuracy of the best detected velocity on a synthetic benchmark consisting of a single translating Gaussian blob. The blob follows constant-velocity motion in the image plane, with velocity magnitude chosen from
$$
\{0.5,\;1.0,\;1.5,\;2.5\}\ \text{px/frame},
$$
and direction chosen from
$$
\{0^\circ,\;45^\circ,\;90^\circ,\;135^\circ,\;180^\circ,\;225^\circ,\;270^\circ,\;315^\circ\}.
$$
For each speed-direction pair, the blob trajectory is initialized so that the motion is approximately centered over the full sequence.

\paragraph{Photon simulation and spatiotemporal volume} For each motion configuration, we generate photon arrivals from the Gaussian flux model using the same synthetic event-generation pipeline as in the rest of the paper. The spatiotemporal volume has dimensions
$$
256 \times 128 \times 128,
$$
corresponding to time, height, and width, respectively. We repeat each configuration for \(10\) independent random seeds and for multiple photon levels by scaling the source amplitude. In the reported plots, the horizontal axis is the resulting photon rate, expressed in photons per second.

\paragraph{Velocity search and detection} For each simulated sequence, events are binned into a spatiotemporal tensor, transformed into the Fourier domain, and converted into a 2D velocity energy map using the plane-energy aggregation procedure described above. Candidate velocities are sampled uniformly over
$$
[-3,3]\times[-3,3]
$$
with \(150\) bins per axis. We then apply the local rank-based CFAR detector to this energy map with a window $\window=100$ and a guard of $g=3$. For each trial, we retain the detected velocity closest to the ground-truth motion and treat it as the best detection. Thus, this experiment isolates the intrinsic estimation accuracy of the detector once a plausible candidate has been produced, rather than emphasizing the total number of detections. In this sense, it measures the quality of the best-matched hypothesis, conditional on the presence of a reasonable candidate in the detection set.

\paragraph{Reported quantities} We report relative endpoint error as the
magnitude metric and angular error as the directional metric. The total
number of detections and inlier composition are analyzed separately in
the following subsection.

\begin{figure}[t]
    \centering
    \includegraphics[width=\linewidth]{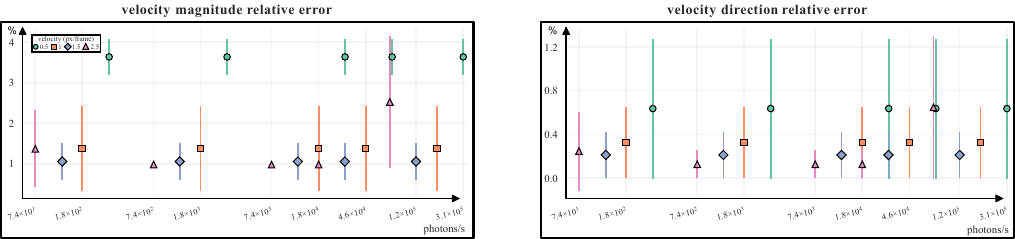}
    \caption{\textbf{Accuracy of the best detected velocity for a single translating Gaussian blob.} \textbf{Left:} Mean relative endpoint error in velocity magnitude,
expressed as a percentage of the ground-truth speed, versus light level.
\textbf{Right:} Mean directional error, expressed as a percentage of
$180^\circ$, versus light level. Markers correspond to different
ground-truth velocities and error bars indicate one standard deviation over
random seeds. The best detected velocity remains accurate and stable
across the tested light level range, with the smallest errors at
intermediate speeds ($1.0$--$1.5$\,px/frame) and slightly larger errors
at the slowest and fastest motions.}
    \label{fig:gaussian_bestdet}
\end{figure}

\paragraph{Velocity magnitude accuracy}
Figure~\ref{fig:gaussian_bestdet} shows that the best detected velocity
is accurate in magnitude over a broad range of light levels. For
intermediate velocities ($1.0$ and $1.5$\,px/frame), the mean relative
endpoint error remains low and nearly flat, indicating that once a
candidate is detected, its magnitude is recovered reliably. The
$0.5$\,px/frame case is similarly stable but at a consistently higher
error, while $2.5$\,px/frame produces the largest error across all
photon rates. Even in that case, however, the error remains modest.
Overall, the dominant source of variation is motion speed rather than
photon rate: across the tested range, the curves are relatively flat,
and the error level depends more strongly on the underlying velocity
magnitude.

\paragraph{Velocity directional accuracy}
The angular error in Figure~\ref{fig:gaussian_bestdet} follows the same
trend. Intermediate speeds again yield the most accurate direction
estimates, with small errors and weak photon-rate dependence. The
$0.5$\,px/frame case is slightly worse but stable, while
$2.5$\,px/frame produces the highest directional error, though the
degradation is gradual. These results indicate that the method recovers
both the magnitude and direction of the dominant motion with high
fidelity once a correct candidate is present in the detection set.

\paragraph{Interpretation} Taken together, these results show that the main limitation of the method in this single-object setting is not the precision of the best detection once a reasonable candidate has been produced. Rather, the best-detection accuracy is already strong over a wide range of photon rates, especially for intermediate motion magnitudes. The most noticeable degradation occurs at the smallest and largest tested speeds. This is consistent with the intuition that very slow motions lie closer to the near-static regime, while larger motions are more affected by discretization and support effects in the sampled velocity domain. Even so, the best detected velocity remains close to the ground truth throughout the tested range.

\subsubsection{Detection set quality for a single moving object}

\paragraph{Scene and motion model}
We extend the previous single-object Gaussian-blob experiment to study not only the accuracy of the best detected velocity, but also the quality of the entire detection set. As before, the scene consists of a single translating Gaussian blob undergoing constant-velocity motion in the image plane. The tested speeds are
\begin{equation}
\{0.5,\;1.0,\;1.5,\;2.5\}\ \text{px/frame},
\end{equation}
and the tested directions are
\begin{equation}
\{0^\circ,\;45^\circ,\;90^\circ,\;135^\circ,\;180^\circ,\;225^\circ,\;270^\circ,\;315^\circ\}.
\end{equation}
For each speed-direction pair, the trajectory is initialized so that the motion is approximately centered over the full sequence.

\paragraph{Photon simulation and spatiotemporal volume}
For each motion configuration, we simulate photon arrivals from the same Gaussian flux model used in the previous experiment. The spatiotemporal volume has dimensions
\begin{equation}
256 \times 128 \times 128,
\end{equation}
corresponding to time, height, and width, respectively. Each configuration is repeated for \(10\) independent random seeds and for multiple photon levels obtained by scaling the source amplitude. In the reported plots, the horizontal axis is the resulting photon rate, expressed in photons per second.

\paragraph{Velocity search and detection}
For each simulated sequence, events are binned into a spatiotemporal tensor, transformed into the Fourier domain, and converted into a 2D velocity energy map using the plane-energy aggregation procedure described above. Candidate velocities are sampled uniformly over
\begin{equation}
[-3,3]\times[-3,3]
\end{equation}
with \(400\) bins per axis. We then apply the local rank-based CFAR detector to this energy map with a window of \(300\) and a guard of \(3\). Unlike the previous subsection, where we retained only the closest detected velocity, here we keep the full set of detected velocity hypotheses for each trial.

\paragraph{Reported quantities}
Let \(\mathcal{D}\) denote the full detection set for a given trial, and let \(\velocityVec_{\mathrm{gt}}\) denote the ground-truth velocity. A detected hypothesis \(\hat{\velocityVec}\in\mathcal{D}\) is declared an inlier if it satisfies the hybrid criterion
\begin{equation}
\norm{\hat{\velocityVec}-\velocityVec_{\mathrm{gt}}}{2}
\leq
\max\!\bigl(\tau \norm{\velocityVec_{\mathrm{gt}}}{2},\; 3\,\Delta v\bigr),
\end{equation}
where \(\tau\) is the relative tolerance parameter and \(\Delta v\) is the spacing of the discrete velocity grid. In our implementation, we use \(\tau=0.15\). This hybrid tolerance is intended to avoid penalizing slow motions too harshly, since at small ground-truth speeds the grid quantization itself can represent a nonnegligible fraction of the motion magnitude.

Using this criterion, we compute three quantities for each trial:
\begin{enumerate}
    \item the total number of detections, \( |\mathcal{D}| \);
    \item the total number of inliers, \( |\mathcal{I}| \);
    \item the total number of outliers,
    \begin{equation}
    |\mathcal{O}| = |\mathcal{D}| - |\mathcal{I}|.
    \end{equation}
\end{enumerate}

\begin{figure}[t]
    \centering
    \includegraphics[width=0.6\linewidth]{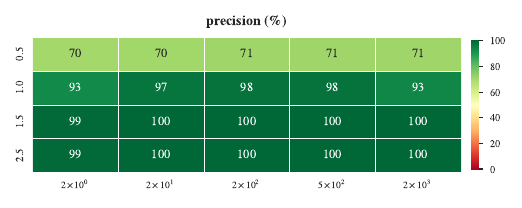}
    \caption{\textbf{Precision of the velocity detection set}, defined as the fraction of
detected hypotheses that are inliers, shown as a function of
ground-truth speed (vertical axis, px/frame) and photon rate (horizontal
axis, photons/s). Precision is near-perfect ($\geq 93\%$) for speeds of
$1.0$\,px/frame and above, and remains stable across photon rates. The
$0.5$\,px/frame case shows lower precision ($\sim$70\%), consistent
with the higher relative discretization error at slow speeds.}
    \label{fig:gaussian_detset}
\end{figure}

\paragraph{Detection precision}
Figure~\ref{fig:gaussian_detset} shows detection precision, defined as
the percentage of detected hypotheses satisfying the inlier criterion.
Precision depends primarily on speed rather than photon rate. At
$0.5$\,px/frame, precision plateaus around $70$--$71\%$ across all
photon levels, indicating that a substantial fraction of detections fall
outside the inlier neighborhood even though the total detection count is
stable. This is attributable to two factors: slower objects produce
broader Fourier-domain support that does not concentrate tightly on the
ideal motion plane, and the fixed absolute grid spacing represents a
larger fraction of the true speed. At $1.0$\,px/frame, precision rises
to $93$--$98\%$, and for $1.5$ and $2.5$\,px/frame it reaches
$99$--$100\%$. Once the motion is sufficiently fast relative to the grid
spacing, nearly all detections concentrate around the true velocity. 

\subsubsection{Detection accuracy across the velocity range}

\paragraph{Goal of the experiment} We next study the range of motion magnitudes over which the detector remains accurate under high illumination and single direction conditions. The goal is to isolate the effect of velocity itself, and in particular to probe failure modes near the lower and upper limits of the discretized search space.

\paragraph{Scene and motion model} We use the same synthetic single Gaussian blob benchmark as in the previous section, but fix the motion direction to \(0^\circ\), so that the ground-truth velocity is always horizontal:
$$
\velocityVec_{\mathrm{gt}} = (\velocity, 0),
\qquad \velocity > 0.
$$
This removes directional variation and allows the experiment to focus exclusively on the dependence of detection accuracy on speed.

\paragraph{Ground-truth speed sweep} We sample the ground-truth speed over the interval
$$
\velocity \in [0.5, 40]\ \text{px/frame},
$$
excluding \(\velocity=0\). The sampling is chosen to be denser near the lower and upper ends of the range, where breakdown is most likely to occur, and coarser in the middle where the method is expected to be more stable. Concretely, we use dense sampling from \(0.5\) to \(2\) px/frame, a coarser sweep over intermediate speeds, and dense sampling again in the high-speed regime. This design is intended to resolve both the low-speed regime, where motion approaches the minimum nonzero scale represented by the search grid, and the high-speed regime, where large displacements and truncation effects may degrade performance.

\paragraph{Photon simulation and spatiotemporal volume} For all trials, we keep the light level fixed at the standard high flux setting used throughout the synthetic experiments. The spatiotemporal volume has dimensions
$$
256 \times 128 \times 128,
$$
and we use the same Gaussian blob model and event-generation pipeline as in the previous benchmark.

\paragraph{Velocity search and detection} The detector is run on the full 2D velocity search space, but the search domain is restricted to nonnegative velocities:
$$
(\velocity_x,\velocity_y) \in [0,40]\times[0,40].
$$
The grid spacing is fixed to
$$
\Delta \velocity = 0.25\ \text{px/frame},
$$
which corresponds to uniform sampling of the first quadrant. Thus, while the ground-truth motion always lies on the horizontal axis, the detector is still free to return off-axis hypotheses, allowing the experiment to reveal both magnitude errors and deviations away from the true direction.

\paragraph{Evaluation protocol} Each speed is evaluated over \(10\) independent random seeds. For every trial, we run the standard plane-energy aggregation and rank-based CFAR detection pipeline, then identify the detected hypothesis nearest to the ground truth in the full 2D velocity plane.

\paragraph{Reported quantities} As the primary metric, we report the relative endpoint error
$$
100\cdot
\frac{\|\widehat{\velocityVec}-\velocityVec_{\mathrm{gt}}\|_2}
{\|\velocityVec_{\mathrm{gt}}\|_2},
$$
together with the directional error, expressed as a percentage of \(180^\circ\). We also visualize the recovered horizontal component \(\widehat{\velocity}_x\) and the recovered vertical component \(\widehat{\velocity}_y\). Since the true motion is purely horizontal, these two diagnostics make it possible to distinguish magnitude mismatch along the correct axis from off-axis leakage into the vertical component.

\begin{figure}[t]
    \centering
    \includegraphics[width=\linewidth]{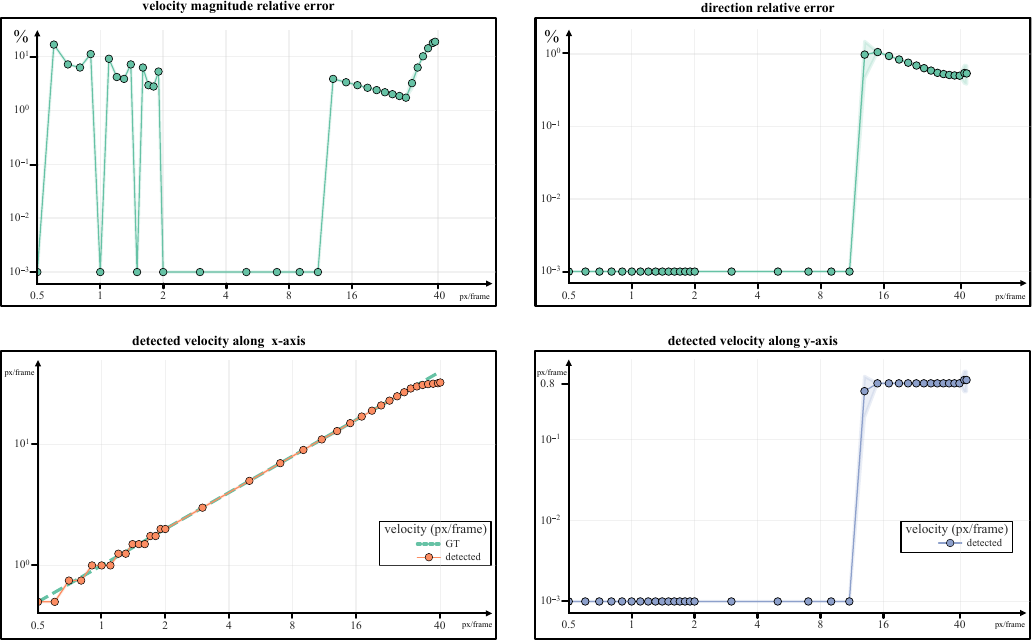}
    \caption{\textbf{Velocity-range benchmark for horizontal Gaussian-blob motion.} The ground-truth motion is purely horizontal with speed increasing from
$0.5$ to $40$\,px/frame. \textbf{Top left:} The relative magnitude error remains very small over an intermediate speed range, then rises at higher speeds as a spurious vertical component appears. \textbf{Top right:} The angular error follows the same transition and increases sharply once the estimate $\velocity_y > 0$. \textbf{Bottom left:} The recovered horizontal component tracks the ground truth closely over most of the sweep and only deviates noticeably at the largest speeds. \textbf{Bottom right:} The recovered vertical component stays near zero at low and intermediate speeds, then exhibits a clear jump, indicating that the breakdown begins as an off-axis error before substantially affecting the magnitude and angle estimates themselves.}
    \label{fig:velocity_limits}
\end{figure}

\paragraph{Velocity detection across speed regimes}
Figure~\ref{fig:velocity_limits} reveals three distinct regimes. In the
intermediate range ($\sim$2.5--10\,px/frame), the detector is
effectively exact: magnitude and directional errors fall to the
numerical floor, $\widehat{\velocity}_x$ follows the identity line, and
$\widehat{\velocity}_y$ remains zero. Below $\sim$2\,px/frame, the
relative error becomes irregular due to velocity-grid quantization---the
absolute error remains small, but the relative metric is sensitive when
$\|\velocityVec_{\mathrm{gt}}\|_2$ is comparable to the grid spacing.

\paragraph{High-speed breakdown}
Above $\sim$12\,px/frame, the detector undergoes a qualitative change.
The recovered horizontal component continues to track the ground truth
(Fig.~\ref{fig:velocity_limits}, bottom-left), but a persistent nonzero vertical component appears
(Fig.~\ref{fig:velocity_limits}, bottom-right), producing a sharp jump in directional error while
the magnitude error remains modest. This indicates that the dominant
failure mode at high speed is not a loss of detectability but a
systematic angular bias: the detector selects an off-axis hypothesis
with nearly correct $\widehat{\velocity}_x$ but spurious
$\widehat{\velocity}_y > 0$. At the highest tested speeds, the
magnitude error also increases, indicating that the breakdown extends
beyond direction to affect the recovered speed. This transition likely
reflects discretization or truncation effects in the Fourier-domain
velocity grid that become significant for large per-frame displacements.

\subsubsection{Detection accuracy with multiple moving objects}

In this section, we evaluate the velocity detector on scenes containing
multiple independently moving objects, testing whether it can recover
several distinct velocity hypotheses from a single observation. We vary
the number of objects from $K=2$ to $K=5$ and compare two detection
strategies: global detection over the full scene and local detection
within object-centered spatial crops.

\paragraph{Scene and motion model} We evaluate the detector in scenes containing multiple independently moving objects, rather than a single isolated blob. Each trial synthesizes a scene with
$$
K\in\{2,3,4,5\}
$$
moving Gaussian blobs undergoing constant-velocity motion in the image plane. For object \(k\in\{1,\dots,K\}\), the ground-truth image-plane velocity is denoted by \(\velocityVec_{\mathrm{gt}}^{(k)}=(v_x^{(k)},v_y^{(k)})\). We sample \(\velocityVec_{\mathrm{gt}}^{(k)}\) by drawing its speed uniformly from a prescribed interval \([v_{\min},v_{\max}]\) and its direction uniformly from \([0,360^\circ)\), and then converting to Cartesian components. To avoid degenerate cases in which two objects have nearly identical motions, we enforce a minimum pairwise separation in velocity space:
$$
\norm{\velocityVec_{\mathrm{gt}}^{(k)}-\velocityVec_{\mathrm{gt}}^{(\ell)}}{2} \ge d_{\min}
\qquad \text{for all } k\neq \ell.
$$
Objects are assigned unequal amplitudes using a fixed decreasing weight profile, so that each scene contains both stronger and weaker motion components. Spatially, each object is placed by sampling a midpoint location within the field of view and then back-projecting its start position from its velocity. This guarantees that each object passes through the scene during the observation window, while allowing objects to overlap spatially.

\paragraph{Photon simulation and spatiotemporal volume} The scene's spatiotemporal flux is generated as a sum of \(K\) translating Gaussian blobs, each with its own velocity and amplitude. Photon arrivals are simulated from this composite spatiotemporal flux using the same synthetic event-generation pipeline as in the single-object experiments, and are then binned into a spatiotemporal tensor of size
$$
T\times H\times W \;=\; 256\times 128\times 128,
$$
corresponding to time, height, and width, respectively. Each configuration \((K,\text{seed})\) is repeated over multiple random seeds, producing different realizations of velocities, layouts, and photon arrival randomness.

\paragraph{Velocity search and detection} For each simulated scene, events are binned into a spatiotemporal tensor, transformed into the Fourier domain, and converted into a 2D velocity energy map using the plane-energy aggregation procedure described above. Candidate velocities are sampled uniformly over
$$
[-3,3]\times[-3,3]
$$
with \(150\) bins per axis. We then apply the local rank-based CFAR detector to produce a binary detection set of velocity hypotheses.

We compare two motion-recovery strategies on the same simulated data.

\begin{enumerate}[label=(\roman*)]
    \item \textbf{Global detection:} In the global setting, the detector is applied once to the full spatiotemporal volume of the scene, producing a single detection set shared by all objects. For each object, we assign the closest detected velocity in this global set and compute the corresponding recovery metrics.
    \item \textbf{Local detection:} In the local setting, we assume known object locations and extract a
    spatial crop centered on the midpoint of each object's trajectory. The
    detection pipeline is then applied independently to each crop, yielding
    an object-specific detection set. For each object, we assign the nearest
    detected velocity from its local set and compute the same metrics as in
    the global case.
\end{enumerate}

\begin{figure}[t]
    \centering
    \includegraphics[width=\linewidth]{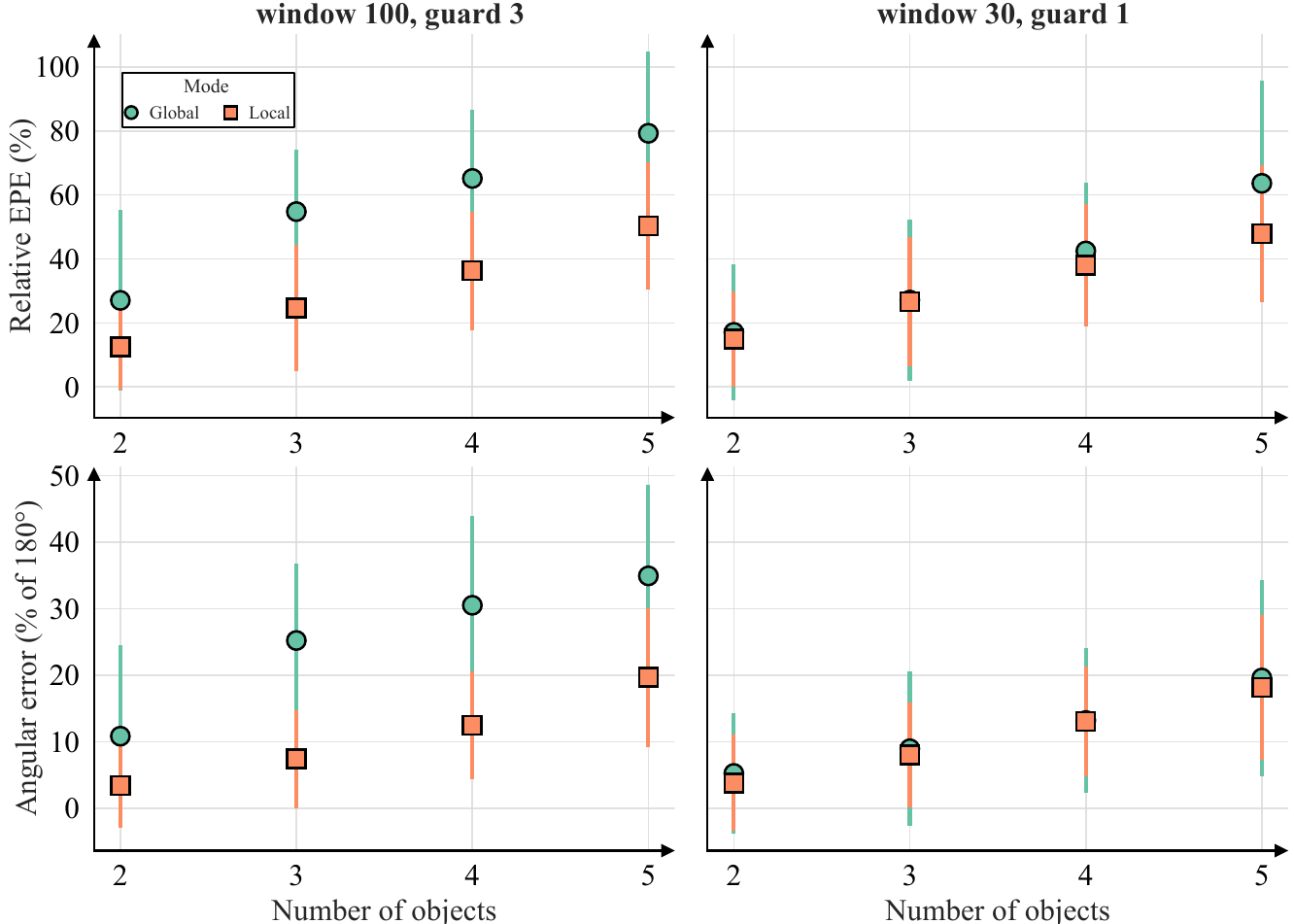}
    \caption{\textbf{Velocity detection with multiple moving objects.} Relative
endpoint error (top) and angular error (bottom) as a function of the
number of objects $K$, for global detection (green circles) and local
detection on object-centered crops (orange squares). Left and right
columns compare two CFAR neighborhood sizes (window $100$, guard $3$
vs.\ window $30$, guard $1$). Error bars indicate one standard
deviation over random seeds. Local detection consistently outperforms
global detection, with the gap widening as $K$ increases. A smaller
CFAR neighborhood partially closes the gap by reducing cross-object
interference in the global setting.}
    \label{fig:mixture_objects}
\end{figure}

\paragraph{Reported quantities} For each object and each mode (global/local), we record the relative endpoint error and the angular error between the estimated and ground-truth velocities. Denoting the matched estimate by \(\hat{\velocityVec}^{(k)}\), we report the relative endpoint error as a percentage of \(\norm{\velocityVec_{\mathrm{gt}}^{(k)}}{2}\), and the angular error as a percentage of \(180^\circ\). A recovery is declared successful when the matched estimate has relative endpoint error below \(10\%\). For the plots in Fig.~\ref{fig:mixture_objects}, results are first averaged over objects within each scene and then averaged over random seeds, yielding scene-level summaries as functions of \(K\).

\paragraph{Detector neighborhood sweep} To assess the effect of CFAR support size in multi-motion scenes, we repeat the experiment for two detector neighborhood settings: a larger neighborhood with window size \(100\) and guard size \(3\), and a smaller neighborhood with window size \(30\) and guard size \(1\). The two columns of Fig.~\ref{fig:mixture_objects} compare these settings directly.

\paragraph{Results}
Figure~\ref{fig:mixture_objects} shows that global detection degrades
steadily as scene complexity increases: both relative endpoint error and
angular error grow with $K$, reaching high values at $K=5$. This is
expected, since the global detector aggregates evidence from all objects
simultaneously, and competing motion signatures interfere in the
velocity domain.

Local detection consistently outperforms global detection across all
settings, confirming that spatial localization suppresses cross-object
interference. The benefit of localization is most pronounced with the
larger CFAR neighborhood (window $100$), where the global method
aggregates over a broader context and is therefore more susceptible to
contamination. With the smaller neighborhood (window $30$), the gap
narrows, indicating that a more localized CFAR support already
mitigates some of the interference even without explicit spatial
cropping.

These results motivate the practical pipeline demonstrated in our
real experiments: applying velocity detection to spatially
localized regions of the scene, where each region is dominated by a
single dominant motion.

\clearpage
{
    \small
    \bibliographystyle{ieeenat_fullname}
    \bibliography{main}
}